\documentclass[12pt]{article}

\usepackage{fontspec}
\usepackage{mathtools} 
\usepackage{amsmath, amssymb, amsthm}
\usepackage{tabularx, booktabs, ragged2e}
\usepackage{geometry}
\usepackage{enumitem}
\usepackage{array}
\usepackage{float}
\usepackage{graphicx} 
\usepackage{tikz} 
\usepackage{algorithm}
\usepackage{algpseudocode} 
\usetikzlibrary{positioning, arrows.meta, fit, shapes.geometric} 

\usepackage{hyperref}
\hypersetup{
    colorlinks=true,
    linkcolor=blue,      
    urlcolor=blue,
    bookmarks=true
}

\usepackage{polyglossia}
\setmainlanguage{english}
\setotherlanguage{arabic}

\newfontfamily\arabicfont[
    Script=Arabic,
    Scale=1.2
]{Amiri-Regular.ttf}
\newcommand{\dhammah}{\text{ḍammah}}

\newtheorem{definition}{Definition}
\newtheorem{theorem}{Theorem}
\newtheorem*{unnumberedtheorem}{Theorem}
\newtheorem{corollary}{Corollary}
\newtheorem{axiom}{Axiom}
\newtheorem{remark}{Remark}
\newtheorem{lemma}{Lemma}

\newtheorem{principle}{Principle} 

\newcommand{\ar}[1]{\textarabic{#1}}

\title{Theoretical Foundations of Computational Islamic Jurisprudence (Fiqh)}
\title{Theoretical Foundations of Computational Islamic Jurisprudence (Fiqh)}
\author{
Elnaserledinellah Mahmoud Abdelwahab\\
\texttt{elnaser@hotmail.com}
}
\date{July 18, 2026}
\date{July 18, 2026}

\begin{document}
\maketitle


\begin{abstract}
This paper presents the theoretical foundations and a comprehensive formal system for Computational Islamic Jurisprudence (Fiqh). We begin by establishing six foundational principles that distinguish computational Fiqh from conventional natural language processing and general theological formalization. These principles define Islamic jurisprudence as a normative deductive science operating on empirically grounded facts, necessitating a dual inductive-deductive methodology, and requiring a logical foundation specifically engineered for Fiqh rather than adopted from existing formalisms. Based on these principles, we construct a three-layer stratified computational language, $\mathcal{F} = D \circ A \circ G$, consisting of: (1) an Arabic Morphological/Grammatical Grounding Engine that transforms fully vocalized Classical Arabic into language-independent Grounded Action Structures (GAS); (2) the Arabic Reified Action Calculus (ARAC) that normalizes these structures into canonical logical representations; and (3) a function-free Horn-clause Datalog engine for executable legal deduction. Furthermore, we introduce a bridging methodology (Algorithm $\Omega$) that dynamically generates a Fiqh-Categorized Attribute Grammar from inductively extracted Horn clauses via an Ontological Meta-Map ($\mathcal{M}_{onto}$), seamlessly connecting the inductive extraction of legal rules to their deductive application. The paper rigorously proves the system's semantic correctness, representational completeness, termination, and decidability. Crucially, by leveraging morphosyntactic inflection markers ($\delta$ modes) as structural constraints, we prove the strict unambiguity of the proposed grammar, achieving deterministic $O(n)$ linear parsing complexity. This work provides a mathematically verifiable, epistemologically grounded, and computationally tractable foundation for automated Islamic jurisprudence.
\end{abstract}

\noindent\textbf{Keywords:}
Computational Fiqh; Uṣūl al-Fiqh; Islamic Jurisprudence;
Morphological Analysis; Syntactic Analysis; Taṣrīf; Iʿrāb;
Legal Causes; al-Ṣabr wa al-Taqsīm; Arabic Computational Linguistics;
Datalog; Knowledge Representation

\newpage
\tableofcontents
\newpage


\section*{Prior Work}
\addcontentsline{toc}{section}{Prior Work}

\renewcommand{\thesubsection}{P.\arabic{subsection}}
\renewcommand{\thesubsubsection}{P.\arabic{subsection}.\arabic{subsubsection}}

Before presenting the comprehensive formal system developed in this work, it is essential to situate our contribution within the broader intellectual and technical context from which it emerged. The present chapter reviews the prior investigations that directly informed and motivated the foundational principles and formal architecture established in this paper. These prior works—spanning historical analysis, epistemological classification, logical formalization, and practical prototype development—collectively constitute the conceptual seed from which the present formal system has grown.

\subsection{Historical and Methodological Foundations}

\subsubsection{The Algorithmic Roots of Islamic Jurisprudence}

A fundamental insight that underpins the present work is the recognition that algorithmic reasoning is not an external methodology imported into Islamic jurisprudence from modern computer science, but rather a native characteristic of the jurisprudential tradition itself. This insight was established in our earlier investigation (Abdelwahab et al., 2016), which traced the historical origins of the term ``algorithm'' to the work of Muhammad ibn Musa al-Khwarizmi (163–235 AH / 780–850 AD).

Contrary to the widespread belief that \textit{Kitab Al-Jabr wa-l-Muqabala} (The Compendious Book on Calculation by Completion and Balancing) is primarily a mathematical textbook, the 2016 paper demonstrated that it is first and foremost a textbook for Islamic jurisprudence [5, 17]. As al-Khwarizmi himself states in his introduction, the work is concerned with:

\begin{quote}
``[a] work on algebra, confining it to the fine and important parts of its calculations, such as people constantly require in cases of inheritance, legacies, partition, lawsuits, and trade, and in all their dealings with one another, or where surveying, the digging of canals, geometrical computation, and other objects of various sorts and kinds are concerned.” (Al-Khwarizmi, c. 830/1831 [13]; cited in Abdelwahab et al., 2016 [1])
\end{quote}

The remaining half of \textit{Kitab Al-Jabr} solves legal questions on trade, geometry, and testimonies, with the by far most important part dealing with Islamic inheritance law—which is complex with regard to both the number of variables and the number of Fiqh-rules (axioms). The introduction of a new Fiqh-method required, then as now, a standard set of persuasive arguments in terms of reproducible propositions and proofs. Therefore, every Fiqh-problem in \textit{Kitab Al-Jabr} is reducible to a linear equation, while the more fundamental method of quadratic polynomials is proven but not used in practice.

This historical fact carries a profound methodological implication: \textbf{computational procedures were not introduced into jurisprudence by modern computer science; rather, jurisprudential requirements themselves historically motivated the development of formal algorithmic methods.} The present work therefore views computational formalization not as the introduction of a new methodology into Fiqh, but as the continuation of an established algorithmic tradition using modern formal languages.

\medskip
\noindent\textbf{Integration into the Present Work}: This historical insight is formalized in \textbf{Foundational Principle II (Algorithmic Juristic Reasoning)} (Section 1.2), which establishes that Islamic jurisprudence is inherently algorithmic and that its computational formalization represents a continuation of an established scientific tradition rather than an external imposition.

\subsubsection{The Epistemological Tripartition of Jurisprudential Questions}

A second major contribution of the 2016 paper was the systematic classification of jurisprudential questions into three distinct epistemological categories, a framework that directly informs the layered architecture of the present formal system.

The 2016 paper distinguished three types of questions:

\begin{enumerate}
\item \textbf{A Logical Question (Qgic)}: A syntactic problem that cannot be solved except through a mental link between the realities of the outside world and the natural human concepts, independent of agreed terminology and belief. This corresponds to the formal deductive processes of logical reasoning.

\item \textbf{A Language Question (QLn)}: A semantic problem solved by reference to the meanings of linguistic patterns, whether these are words, expressions, or sentences. These meanings are either essentially or traditionally known in Arabic.

\item \textbf{A Source Question (SF)}: A search and authentication problem that can only be solved by reference to a finite set of rulings approved on the basis of Quran and Sunnah.
\end{enumerate}

The 2016 paper further demonstrated that these three categories interact systematically within jurisprudential reasoning. For example, the deductive process of analogy (qiyas) relies on Qgic for its logical structure, QLn for the linguistic interpretation of the terms involved, and SF for the authentication of the primary sources \cite{hallaq1997}. The paper concluded that ``logic is a subset of Islamic Jurisprudence'' and that ``any derivation, no matter how long or complex, can be characterized syntactically,'' implying that the deductive part of Shariah could be mechanized completely.

\medskip
\noindent\textbf{Integration into the Present Work}: This tripartition maps directly to the three computational layers of the present formal system:

\begin{itemize}
\item \textbf{QLn} corresponds to the \textbf{Epistemic Grounding Layer} (Sections 3–4), which performs morphological recognition, syntactic recognition, and semantic grounding—all of which depend on the linguistic structure of Classical Arabic.
\item \textbf{Qgic} corresponds to the \textbf{Executable Deductive Layer} (Section 5), which performs formal logical deduction using function-free Horn-clause Datalog.
\item \textbf{SF} corresponds to the \textbf{Bridging Methodology} (Section 12), which connects the inductive extraction of legal rules from primary sources to their deductive application.
\end{itemize}

This direct correspondence validates the epistemological architecture of the present work and demonstrates its continuity with earlier methodological investigations.

\subsection{Logical and Computational Formalization}

\subsubsection{The Entscheidungsproblem for Fiqh-Systems}

A central contribution of the 2016 paper was the explicit formulation
of an \textit{Entscheidungsproblem} for Islamic jurisprudence
\cite{turing1936}. The underlying question was whether the properties
of Complete and Consistent Legislation could themselves be determined
by means of a suitable formal logical system.

The 2016 paper introduced three sets for this purpose:

\begin{equation}
SF=\text{the set of Fiqh-questions},
\end{equation}

\begin{equation}
QR=\text{the set of legal rulings},
\end{equation}

and

\begin{equation}
QE=\text{the set of reasons}.
\end{equation}

It then formulated two propositions.

\paragraph{Proposition 1: Consistent Legislation.}

The original formulation states that every religious ruling has a
reason. In the notation of the 2016 paper, this was expressed by:

\begin{equation}
\forall M\in SF;\forall R\in QR:
\left[
(M\rightarrow R)
\rightarrow
\exists E\in QE;(E\rightarrow R)
\right].
\end{equation}

The intended meaning is that whenever a legal question $M$ leads to a
legal ruling $R$, there exists a corresponding reason $E$ for that
ruling.

The present work retains this original formulation as the conceptual
starting point, but gives it a more explicit and operational
interpretation. Instead of treating the existence of a reason as an
abstract relation between elements of $QE$ and $QR$, the present formal
system represents the reason for a ruling through an explicit
Fiqh-rule and its formal derivation. Thus, for the formalized system,
the corresponding operational requirement can be expressed as:

\begin{equation}
\forall R\in QR,\qquad
\exists F\in FR:
F\vdash R.
\end{equation}

Here $FR$ denotes the set of Fiqh-rules represented in the formal
system, and $F\vdash R$ means that the ruling $R$ is derivable through
the rule $F$.

This formulation is not only nearer to the informal statement of Proposition 1 in the 2016 paper, but also more concrete than the original abstract formulation because the present system explicitly represents the rules, their premises, their conclusions, and the deductive process. The ``reason'' of a ruling is therefore not merely asserted to exist: it is represented by an identifiable rule together with a formal
derivation.

The two formulations should not be regarded as contradictory. Rather,
the second gives a constructive realization of the first:

\begin{equation}
\boxed{
\text{Existence of a reason}
\quad\Longrightarrow\quad
\text{Explicit rule and formal derivation}.
}
\end{equation}

In this sense, the present work turns the abstract requirement of
Consistent Legislation formulated in 2016 into an operational
property of the implemented Fiqh system.

\paragraph{Proposition 2: Complete Legislation.}

The second proposition of the 2016 paper states that every religious
question has a legal ruling. Formally:

\begin{equation}
\forall M\in SF,\qquad
\exists R\in QR:
M\rightarrow R.
\end{equation}

This proposition concerns the coverage of the legal question space.
It requires that no admissible Fiqh-question belonging to $SF$ be
left without a corresponding legal ruling.

The two propositions therefore express complementary properties.
Complete Legislation concerns the existence of a ruling for every
admissible question, whereas Consistent Legislation concerns the
existence of a formal reason or derivation for the ruling.

They may consequently be represented schematically as:

\begin{equation}
\text{Question}
\longrightarrow
\text{Ruling}
\longleftarrow
\text{Rule and Reason}.
\end{equation}

The significance of this reformulation for the present work is that
the two requirements can now be connected directly to the actual
formal machinery of the system. The question $M$ is represented in
the formal query space; the ruling $R$ is represented as a legal
conclusion; and the reason for $R$ is represented by the Horn clause
whose premises are satisfied by $M$ and whose conclusion produces
$R$.

Thus, the abstract structure proposed in 2016 acquires the following
constructive interpretation in the present system:

\begin{equation}
M
\longrightarrow
\text{instantiated premises}
\longrightarrow
F
\longrightarrow
R.
\end{equation}

This is the essential transition from the original conceptual
formulation to the present executable formalization.

\textbf{Lemma 1} of the 2016 paper stated that if Complete and
Consistent Legislation hold, then Islamic Jurisprudence is complete
with respect to a given set of Fiqh-axioms and an intrinsic formal
system of deduction. The present work makes this statement
computationally explicit by specifying the ontology, Fiqh Grammar,
Horn-clause axioms, and deductive mechanism required for such a
system.

\textbf{Proposition 3} of the 2016 paper then established the
connection with the Entscheidungsproblem: if the formal system

\begin{equation}
(QL_n,Q_c,Q_v,G_{mr},Q_g)
\end{equation}

is both logically complete and decidable, then it is decidable whether
Complete and Consistent Legislation hold, and hence whether Lemma 1
holds.

The present work realizes this program within a specified closed
Fiqh chapter through Theorems 26--28. The resulting system does not
merely assert that a reason for a ruling exists. It provides a
constructive mechanism by which the relevant premises are matched to
a Horn rule and the corresponding legal conclusion is derived.

Accordingly, the relation between the 2016 formulation and the
present formalization may be summarized as:

\begin{equation}
\begin{aligned}
\text{2016 Consistent Legislation}
&\longrightarrow
\text{existence of a reason},\ \\
\text{Present formalization}
&\longrightarrow
\text{explicit rule and derivation},\ \\
\text{2016 Complete Legislation} 
&\longrightarrow
\text{every admissible question has a ruling},\ \\
\text{Present formalization}
&\longrightarrow
\text{constructive computation of that ruling}.
\end{aligned}
\end{equation}

The resulting formulation preserves the conceptual foundations of the
2016 paper while replacing its abstract description of reasons and
deduction by the explicit formal objects and executable procedures
developed in the present work.

\subsubsection{The Epistemic Weight of Completeness in Islamic Jurisprudence}

The distinction between completeness and incompleteness is fundamental
to the logical analysis of formal systems. A sufficiently expressive
theory of arithmetic, such as first-order Peano Arithmetic
($\mathrm{PA}$), cannot in general decide every sentence expressible in
its language. Gödel's incompleteness theorems show that, once a formal
system is sufficiently expressive to encode the relevant self-referential
arithmetic constructions, there are true arithmetic statements that are
not derivable from the axioms.

The situation is fundamentally different for the restricted algebraic
domain used by Al-Khwārizmī in inheritance calculation. The relevant
constraints can be represented as finite systems of linear equations
and inequalities over the rational numbers. This restriction removes
the source of incompleteness associated with unrestricted arithmetic:
the system does not attempt to represent arbitrary natural-number
computation, recursive self-reference, or unrestricted multiplication
of variables. Instead, it operates inside the decidable theory of
linear rational arithmetic.

The completeness of this linear theory has a precise meaning. For a
fixed finite-dimensional rational space, every first-order sentence
expressible in the corresponding language is either true in the
intended structure $\mathbb{Q}^{d}$ and derivable from the theory, or
false in $\mathbb{Q}^{d}$ and its negation is derivable. In other words,
for every sentence $\varphi$,

\begin{equation}
T_d \vdash \varphi
\qquad\text{or}\qquad
T_d \vdash \neg\varphi.
\end{equation}

The reason this completeness is valid is structural rather than
accidental (detailed in Section 10, Part II of this work). The theory is restricted to a fixed rational vector-space
structure whose coordinates are completely determined over the field
$\mathbb{Q}$. Linear constraints can be transformed by effective
elimination procedures, such as Fourier--Motzkin elimination, into
equivalent constraint systems with progressively fewer variables.
Because every elimination step is finite and preserves satisfiability,
quantified linear statements can ultimately be reduced to quantifier-free
linear conditions. Thus there is no residual class of undecidable
linear propositions hidden beyond the computational procedure.

This should be distinguished carefully from Gödel completeness for
first-order logic. Gödel's completeness theorem states that semantic
validity in first-order logic coincides with formal provability:
every first-order sentence that is true in \emph{all} models of a
first-order theory is provable from that theory. By contrast, the
completeness relevant here is the completeness of a particular
restricted mathematical theory: the theory of the specified rational
linear structure is sufficiently constrained that its sentences are
effectively decidable and its truth can be captured by formal
derivation. The result therefore concerns both the logical structure
and the restricted computational domain.

This distinction is especially important in the jurisprudential
context. The purpose of an algebraic formalization is not merely to
produce correct answers for a collection of examples. A complete
formalization of a delimited class of inheritance problems should
provide a general mechanism under which every admissible instance can
be represented, tested, and resolved without leaving an unhandled
logical residue. Al-Khwārizmī's algebra achieves this by replacing
case-by-case numerical guessing with a uniform system of equations
whose solutions are determined by rational arithmetic.

The historical importance of this transformation is therefore
epistemic as well as computational. The legal question is first
translated into a finite algebraic structure; the resulting constraints
are then manipulated by operations whose correctness is mathematically
established; and the resulting allocation can be independently
verified by substitution into the original constraints. Within the
scope of the represented inheritance problems, the computation is
therefore transparent, terminating, and mechanically checkable.

The present framework pursues the same methodological objective for a
different class of jurisprudential domains. Under the closed-chapter
assumption, the relevant case space is exhausted relative to the
specified ontology, and the \textit{al-sabr wa-al-taqsīm} procedure
constructs the corresponding Horn Fiqh axiom set \cite{abdelwahab2026sabr}. If the resulting
axiomatization is complete for that closed case space and the Horn
deductive engine is complete for the specified rule class, then every
admissible legal question in that domain is represented by the formal
system and has a derivable ruling.

This gives the following correspondence:

\begin{equation}
\begin{aligned}
\text{Al-Khwārizmī:}\quad
&\text{complete inheritance representation}
\;\longrightarrow\;
\text{complete linear deduction},\\
\text{Present framework:}\quad
&\text{complete closed-chapter axiomatization}
\;\longrightarrow\;
\text{complete Horn deduction}.
\end{aligned}
\end{equation}

The connection with the two propositions established in the 2016
paper is direct. Proposition~1, \emph{Consistent Legislation}, asserts
that every ruling has a reason and therefore excludes arbitrary
unmotivated rulings. Proposition~2, \emph{Complete Legislation},
asserts that every religious question in the specified question space
has a ruling:

\begin{equation}
\forall M\in SF,\qquad
\exists R\in QR:\;M\longmapsto R.
\end{equation}

For a closed chapter $CC$, the exhaustive construction of
$A_{CC}$ supplies the formal representation of this complete case
space, while completeness of the Horn engine supplies the corresponding
derivability. Hence the resulting claim is not universal completeness
of Fiqh, nor completeness of unrestricted mathematics. It is
\emph{domain-relative computational completeness}: completeness of a
specified jurisprudential subchapter relative to its ontology, its
closed question space, its generated axioms, and its deductive engine.

In this precise sense, the achievement sought by the present framework
is analogous to the methodological achievement of Al-Khwārizmī.
Al-Khwārizmī transformed a delimited class of inheritance problems into
a complete and decidable algebraic procedure. The present framework
extends the same principle from numerical inheritance calculation to a
broader formal representation of closed Fiqh subchapters, using
exhaustive case generation and complete Horn deduction rather than
linear equations.

\subsubsection{Complexity of Deduction: Detailed vs. General Queries}

The 2016 paper made an important distinction between two types of deduction, each with different computational complexity properties:

\begin{enumerate}
\item \textbf{Detailed Deduction (Proposition 4)}: The deduction of detailed sentences (concrete queries) by substitution into general ones is efficient, with worst-case polynomial complexity with respect to the length of sentences. Proof: straightforward substitution of values into logic formulae using simple procedures of linear complexity.

\item \textbf{General Deduction (Proposition 5)}: The deduction of general sentences from a set of general and detailed ones is NP-Complete, because it is an instance of the SAT problem in Propositional Logic.
\end{enumerate}

This distinction has a direct practical consequence: while the amount of computation needed for answering concrete queries is relatively small if the Fiqh-sub-system is complete (i.e., if there are Fiqh-rules to be applied to each query), deducing new Fiqh-rules from others may be computationally challenging.

\medskip
\noindent\textbf{Integration into the Present Work}: The present formal system addresses this distinction by:
\begin{itemize}
\item Focusing on \textbf{detailed deduction} (the application of established rules to concrete queries), achieving \textbf{\( O(n) \) deterministic linear complexity} for the specialized parser (\textbf{Theorem 20}).
\item Acknowledging the computational challenge of \textbf{general deduction} (inductive rule extraction) in \textbf{Section 13 (Challenges and Future Work)}, noting that the construction of the Ontological Meta-Map \( \mathcal{M}_{\text{onto}} \) and the Horn-clause program \( P \) for a closed chapter remains a significant knowledge engineering task.
\item The bridging methodology (\textbf{Algorithm \( \Omega \), Theorem 28}) addresses this challenge by providing a formal, bijective connection between inductively extracted rules and their deductive application, thereby modularizing the general deduction problem as a separate research task.
\end{itemize}

\subsection{Practical Prototype and Heuristic Methodologies}

\subsubsection{The Four Essential Elements of a Fiqh Question}

The 2016 paper introduced a structural heuristic for decomposing compound Fiqh questions into four essential elements: \textbf{subject, tool, reason, and method}.

The subject refers to the person or entity to whom the ruling applies; the tool refers to the means or action involved; the reason refers to the legal cause ('illah) that triggers the ruling; and the method refers to the procedural or evidentiary framework within which the ruling is determined. These four elements interact to produce numerous possibilities and complex interactivity, with main qualities containing secondary qualities that may also interact within or outside the main qualities.

The 2016 paper constructed trees of defined terms for each question element, devised to contain all aspects of the addressable Fiqh-questions. This was followed by the generation of practical questions via a straightforward combinatory algorithm, which was then reduced by deriving negative and positive rules to exclude null-results.

\medskip
\noindent\textbf{Integration into the Present Work}: This structural heuristic maps directly to the non-terminals of the Fiqh-Categorized Attribute Grammar \( G_{\mathrm{Fiqh}}^{\delta} \) (Section 10.3). For example:

\begin{itemize}
\item \textbf{Subject} corresponds to non-terminals like \texttt{Seller\_Raf'}, \texttt{Buyer\_Jarr}, \texttt{Owner\_Raf'}.
\item \textbf{Tool} corresponds to non-terminals like \texttt{Commodity\_Nasb}, \texttt{Price\_Jarr}, \texttt{Property\_Jarr}.
\item \textbf{Reason} corresponds to the logical predicates in the body of Horn clauses (e.g., \texttt{health\_state}, \texttt{purification\_state}).
\item \textbf{Method} corresponds to the procedural grammar rules (e.g., \texttt{LegalCondition}, \texttt{LegalQuery}).
\end{itemize}

The \( \delta \)-typed non-terminals enforce the morphosyntactic constraints that distinguish these roles, ensuring that the parser assigns each element to its correct legal slot.

\subsubsection{Negative and Positive Rules}

To address the combinatory explosion that results from generating all possible questions from a tree of defined terms, the 2016 paper introduced the concept of \textbf{negative and positive rules}.

The process involved:
\begin{enumerate}
\item Generating a complete truth table containing all possible Fiqh-questions in a sub-chapter.
\item Inspecting the truth table entries for the possibility of abstracting rule patterns.
\item Deriving two sets of rules—negative and positive—that substitute the entire truth table.
\end{enumerate}

For example, in the purification (Taharah) chapter according to the Shafi'i school, the 2016 paper derived a set of negative rules (e.g., ``Tayammum becomes invalid when water becomes available,'' ``It is not permissible to use something of value to remove impurity'') and positive rules (e.g., ``When the lesser impurity and ceremonial impurity apply together, the grand ablution is enough to remove both,'' ``The discharge of semen does not make wudu necessary'').

This process of deducing negative and positive rules was identified as an efficient practical (heuristic) alternative to classical deduction of general formulas, which may be computationally intensive (Proposition 5).

\medskip
\noindent\textbf{Integration into the Present Work}: The concept of negative and positive rules is formalized in the present system through:
\begin{itemize}
\item The \textbf{LegalNegation} non-terminal, which explicitly flags logical structures as negated (e.g., mapping predicate \texttt{unavailable(water)} to a non-terminal structure that outputs negation facts).
\item The bijective Ontological Meta-Map \( \mathcal{M}_{\text{onto}} \) (Definition 24), which ensures that a positive predicate like \texttt{available(water)} cannot be mapped to a negated non-terminal, and vice versa.
\item The bridging methodology (\textbf{Algorithm \( \Omega \)} ), which generates grammars that explicitly distinguish between positive and negated premises.
\end{itemize}

This formalization provides a principled, mathematically grounded treatment of the heuristic insight introduced in the 2016 prototype.

\subsubsection{Practical Prototype Implementation}

The 2016 paper implemented conceptual prototypes for two sub-chapters of purification (Taharah): Wudu and Tayammum. For Tayammum, both simple and compound question-answering procedures were realized; for Wudu, only the compound question part was implemented.

The method relied on generating all possible questions under a Fiqh-chapter or sub-chapter by means of an algorithmic machine. The truth table was reduced (manually) to two sets of rules which formed the backbone of a logical engine. A restricted logic compiler, understanding the format of Fiqh-rules (propositional calculus formulas), was written to enable matching of user queries with rules and providing intelligent results with explanations.

For compound questions (involving sequences of actions), the algorithm was realized as a finite-state automaton, with the order of actions being represented as state transitions. For example, a Hanafi Wudu-Automaton is not necessarily deterministic since preserving the order of Wudu-actions is not obligatory according to the Hanafi school.

\medskip
\noindent\textbf{Integration into the Present Work}: While the present work focuses on formalization rather than implementation, the practical prototype demonstrates the feasibility of the approach and provides a concrete reference for future implementation efforts. The finite-state automaton concept is generalized in the present work through the \( \delta \)-driven deterministic parser, which achieves \( O(n) \) linear complexity by leveraging the unambiguous grammar (Lemma 1).

The challenges identified in the 2016 prototype—particularly the combinatory explosion and the need for manual reduction—are addressed in the present work through:
\begin{itemize}
\item The unambiguity proof (Lemma 1), which eliminates combinatory explosion at the syntactic level.
\item The \( \delta \)-driven parser (Theorem 20), which achieves deterministic linear complexity.
\item The Ontological Meta-Map (Definition 24), which provides a principled formal connection between logical predicates and syntactic non-terminals.
\end{itemize}

\subsection{Synthesis and Contribution of the Present Work}

The prior investigations reviewed above collectively established the following foundational insights:

\begin{enumerate}
\item \textbf{Historical-Algorithmic Continuity}: Algorithmic reasoning is native to Islamic jurisprudence, as demonstrated by the historical connection between al-Khwarizmi's algebra and Fiqh methodology.

\item \textbf{Epistemological Tripartition}: Jurisprudential questions can be systematically classified into linguistic (QLn), logical (Qgic), and source-authentication (SF) categories, each requiring distinct computational treatments.

\item \textbf{Entscheidungsproblem}: The decidability of Fiqh-systems depends on the consistency and completeness of the underlying formal system, and a positive solution requires a formal system that is both complete and decidable.

\item \textbf{Complexity Distinction}: Detailed deduction (applying rules to concrete queries) is polynomial-time, while general deduction (inferring new rules) is NP-Complete.

\item \textbf{Practical Heuristics}: The construction of negative and positive rules, the decomposition of questions into four essential elements, and finite-state automaton representations provide practical pathways for implementation.
\end{enumerate}

\textbf{The present work builds upon these insights by transforming them into a comprehensive, mathematically rigorous formal system.} Specifically:

\begin{itemize}
\item The tripartition (QLn, Qgic, SF) is realized in the three-layer computational architecture \( F = D \circ A \circ G \).
\item The Entscheidungsproblem is resolved through rigorous proofs of decidability (Theorems 15, 26, 28).
\item The complexity distinction is addressed by achieving \( O(n) \) deterministic linear complexity for detailed deduction (Theorem 20) while acknowledging the NP-Complete nature of general deduction in Section 13.
\item The practical heuristics (negative/positive rules, four essential elements, finite-state automata) are generalized and formalized through the \( \delta \)-driven grammar, the LegalNegation non-terminal, and the Ontological Meta-Map.
\end{itemize}

The 2016 paper provided the \textbf{seed}—the conceptual insight that Fiqh can be formalized algorithmically. The present work provides the \textbf{full tree}—a complete, rigorous, and verifiable formal system grounded in that insight, satisfying the six foundational principles established in Chapter 1.

\subsection*{Integration Note}

Throughout the present work, the foundational insights from these prior investigations are explicitly referenced and formalized. The reader will find direct correspondence between the tripartition of questions and the three computational layers, between the Entscheidungsproblem and the decidability theorems, between the complexity distinction and the \( O(n) \) linear parsing bound, and between the practical heuristics and the \( \delta \)-driven grammar. This intellectual continuity ensures that the present work is not merely a formalization of Islamic jurisprudence in isolation, but the culmination of a coherent research program that began with the earliest systematic treatments of Fiqh methodology and continues through the present comprehensive formal system.


\renewcommand{\thesubsection}{\arabic{subsection}}
\renewcommand{\thesubsubsection}{\arabic{subsection}.\arabic{subsubsection}}


\section*{The Computational Fiqh Research Program: A Unified Overview}
\addcontentsline{toc}{section}{The Computational Fiqh Research Program}

This research program establishes a comprehensive, mathematically verified framework for Computational Islamic Jurisprudence (Fiqh) by sequentially bridging classical legal theory with modern computer science. It begins by formalizing the classical method of al-Ṣabr wa al-Taqsim to computationally extract minimal legal rules, conditions, and causes from the complete truth tables of "closed" Fiqh chapters, which are then proven to collapse into deterministic Finite State Automata (FSA) for constant-time O(1) execution with mathematically guaranteed LLM safety. To justify using Arabic morphology as a reliable semantic foundation, the project formally refutes the Saussurean axiom of the "arbitrary sign" for Classical Arabic, proving that its root-pattern system is structurally motivated and computationally compressible. This linguistic foundation supports a rigorous, three-layer formal system—comprising a morphological grounding engine, a Reified Action Calculus, and a function-free Datalog deductive core—bridged by an algorithm that dynamically generates unambiguous, linear-time parsing grammars directly from inductively extracted legal rules. Finally, the theoretical framework is actualized in a pragmatic neuro-symbolic prototype that processes unvocalized Arabic natural language through symbolic wildcard matching and human-in-the-loop LLM verification, augmented by Avicennian modal logic and dynamic morphological variant generation, ultimately yielding an end-to-end pipeline that extracts normative legal rulings from primary sources and applies them with absolute epistemological soundness, decidability, and computational tractability.

\renewcommand{\thesubsection}{P.\arabic{subsection}}
\renewcommand{\thesubsubsection}{P.\arabic{subsection}.\arabic{subsubsection}}

\setcounter{subsection}{4} 

\subsection{The Five Stages of the Research Program}

The present work is not an isolated theoretical exercise but the culmination of a coherent, multi-layered research program spanning five interconnected investigations. Each paper addresses a distinct component of the computational problem while building upon the preceding ones. The program as a whole may be understood as progressing through five sequential stages, each corresponding to one of the papers in this series.

\subsubsection{Stage I: Inductive Knowledge Acquisition (Sabr wa al-Taqsim)}

The first paper formalizes the classical usuli method of al-Sabr wa al-Taqsim (Examination and Division) as a set-theoretic algorithm operating on truth tables of closed Fiqh chapters [2]. It extracts minimal positive and negative rules, computes Shurut (conditions), Candidate ’Ilal (causes), and Candidate Mawani’ (preventatives) [12].The paper proves the Structural Decomposition Theorem:

\[
h \iff S \wedge I^*
\]

where \( h \) is the ruling function, \( S \) is the set of Shurut (conditions), and \( I^* \) is the disjunction of all candidate 'Ilal. The paper establishes bounded complexity \( O(3^k) \) for fixed closed chapters, where \( k \) is the number of variables in the Ontological Concepts Tree (OCT).

\medskip
\noindent\textbf{Significance}: This paper solves the \textbf{inductive knowledge acquisition} problem: it provides a mechanical method for deriving the operational rules of a Fiqh chapter from a complete truth table of its legal states.

\medskip
\noindent\textbf{Link to Subsequent Papers}: The extracted Horn clauses \( P \) become the input to the FSA framework (Stage IV) and the bridging methodology (Stage III, Algorithm \( \Omega \)).

\subsubsection{Stage II: Linguistic and Epistemological Foundation (No Meaning)}

The second paper addresses a foundational assumption underlying much of modern linguistics and computational semantics: the arbitrariness of the linguistic sign. Through a formal mathematical model of Arabic non-concatenative morphology [3, 7], the paper proves that Classical Arabic contains no absolutely arbitrary primitive signifier.

The paper establishes the following key results:

\begin{itemize}
\item \textbf{Morphological Correspondence Theorem}: Every surface word is uniquely generated by a root-pattern pair \( (R, P) \).
\item \textbf{Semantic Localization Theorem}: Lexical meaning is determined at the derivational level before surface realization.
\item \textbf{Theorem (Structural Impossibility of the Absolutely Arbitrary Sign)}: Given the root-pattern architecture and the phonosemantic motivation documented by Ibn Jinni, Arabic contains no absolutely arbitrary signifier.
\item \textbf{Three-Level Framework}: Arabic is motivated at Levels W (word) and M (morpheme), with arbitrariness confined to Level P (sub-morpheme). This places Arabic arbitrariness two levels deeper than Indo-European languages.
\item \textbf{Epistemological Asymmetry}: The universal claim "all signs are arbitrary" is undecidable, while the paper's claims about Arabic are decidable.
\end{itemize}

\medskip
\noindent\textbf{Significance}: This paper provides the \textbf{linguistic and epistemological foundation} for the entire research program. It establishes that Classical Arabic is a motivated, structurally locked system—not a flat, arbitrary sign system. This justifies the morphological grounding layer \( G \) in the formal system.

\medskip
\noindent\textbf{Link to Subsequent Papers}: The foundational paper (Stage III) explicitly references this linguistic framework in Principle I and builds its morphological grounding engine on the non-concatenative root-pattern architecture formalized here. The prototype (Stage V) implements the morphological repository derived from this theory.

\subsubsection{Stage III: A Comprehensive Formal System (The Foundational Paper)}

The present paper synthesizes the insights from the preceding investigations into a complete formal system for computational Islamic jurisprudence. It establishes six foundational principles and constructs a three-layer stratified computational language.

The six foundational principles are:

\begin{enumerate}
\item \textbf{Epistemological Foundation}: Fiqh is a normative deductive science operating on empirically grounded facts, distinct from theology and conventional NLP.
\item \textbf{Algorithmic Juristic Reasoning}: Islamic jurisprudence is inherently algorithmic, comprising both inductive rule acquisition and deductive rule application.
\item \textbf{Engineering the Logical Foundation}: A formal system for Fiqh must be explicitly engineered for Fiqh's requirements, not adopted from existing formalisms.
\item \textbf{Epistemic Grounding}: Legal deduction must be preceded by a formal grounding process that transforms natural-language descriptions into grounded legal facts.
\item \textbf{Minimal Computational Expressiveness}: The computational realization shall employ the least expressive formalism capable of faithfully executing jurisprudential reasoning.
\item \textbf{Layered Architecture}: The system shall be organized into distinct architectural layers, each responsible for a single well-defined scientific function.
\end{enumerate}

The formal language is defined as the composition:

\[
F = D \circ A \circ G
\]

where:

\begin{itemize}
\item \( G \) is the Arabic Morphological/Grammatical Grounding Engine
\item \( A \) is the Arabic Reified Action Calculus (ARAC)
\item \( D \) is the function-free Horn-clause Datalog engine
\end{itemize}

The paper proves the system's correctness, completeness, termination, and decidability. Crucially, it introduces the \( \delta \)-driven grammar, proving strict unambiguity (Lemma 1) and achieving \( O(n) \) deterministic linear parsing (Theorem 20). It also introduces Algorithm \( \Omega \), a bridging methodology that dynamically generates a Fiqh grammar from Horn-clause programs via a bijective Ontological Meta-Map \( \mathcal{M}_{\text{onto}} \), connecting inductive rule extraction to deductive application.

\medskip
\noindent\textbf{Significance}: This paper provides the \textbf{complete mathematical foundation} for computational Islamic jurisprudence. It synthesizes the inductive extraction (Stage I), the linguistic grounding (Stage II), and the execution logic (Stage IV) into a unified formal system.

\medskip
\noindent\textbf{Link to Subsequent Papers}: The prototype (Stage V) implements a pragmatic subset of this formal system, acknowledging deviations and extensions while maintaining epistemological soundness.

\subsubsection{Stage IV: Backend Execution via Finite-State Automata (Al-Ghasali)}

Given the formal system and its Horn-clause representation, the fourth paper addresses the problem of efficient, safe execution. It proves that closed Horn chapters collapse to deterministic Finite State Automata (FSA), enabling \( O(1) \) verdict extraction after offline compilation.

The key results are:

\begin{itemize}
\item \textbf{Lemma 1 (Finite-State Collapse of Closed Horn Chapters)}: The operational semantics of any strict Horn-clause program derived from a closed Fiqh chapter collapse to those of a deterministic finite-state automaton.
\item \textbf{Lemma 2 (Finite-State Preservation under Hierarchical Composition)}: Hierarchical composition of sub-chapter FSAs preserves \( O(1) \) complexity.
\item \textbf{Formalization of Tartib (Juristic Order)}: Temporal sequences are flattened into static ontological variables via FSA transitions, preserving the memoryless property of the automaton.
\item \textbf{LLM Safety via Grammar-Guided Constrained Decoding}: The system dynamically constructs right-linear grammars from the FSA execution history, augmented by a Bounded Neutral Lexicon and Valence Locking, mathematically guaranteeing the absence of Decisional Hallucination, Juristic Factual Hallucination, and Semantic Inversion.
\end{itemize}

\medskip
\noindent\textbf{Significance}: This paper solves the \textbf{backend execution} problem: it provides the most efficient, mathematically safe runtime engine for Fiqh simulation.

\medskip
\noindent\textbf{Link to Subsequent Papers}: The FSA framework complements the prototype (Stage V). The prototype handles frontend linguistic grounding; the FSA handles backend execution. The prototype's "Previous Practical Work" section explicitly acknowledges this complementarity.

\subsubsection{Stage V: Frontend Pragmatic Implementation (Neuro-Symbolic Prototype)}

The final paper translates the dense mathematical formalism into executable software that handles real-world, unvocalized Arabic input. It presents a dual-application prototype sharing a common non-concatenative morphological repository.

The two applications are:

\begin{enumerate}
\item \textbf{Fiqh Question-Answering (QA) Engine}: Maps unvocalized Arabic natural language to hard-wired Horn-clause rule systems via symbolic wildcard matching and a human-in-the-loop neuro-symbolic verification process.
\item \textbf{Deductive Avicenna Engine}: Applies Avicennian modal logic [6] augmented by dynamic morphological variant generation (Epicene Bridges) to perform complex syllogistic reasoning.

\end{enumerate}

The novel components introduced are:

\begin{itemize}
\item \textbf{Epicene Bridges}: Dynamic morphological variant generation from root-pattern implications, enabling seamless syllogistic chaining across different morphological forms.
\item \textbf{Symbolic Wildcard Pattern Matching}: A robust matcher for unvocalized Arabic with scoped negation handling and strict structural constraints.
\item \textbf{Human-in-the-Loop Verification}: LLM proposals \( S' \) are verified by the user before entering the deductive core, preventing hallucination.
\item \textbf{Symbolic Provenance and Backtracking}: Every derived conclusion is tracked back to its originating premises and rules, providing full auditability.
\item \textbf{Constrained Neuro-Symbolic Integration}: The LLM is confined to grounding (hypothesis generation) and post-deductive translation; normative deduction remains strictly symbolic, preserving the epistemological separation mandated by Foundational Principle I.
\end{itemize}

The prototype acknowledges deliberate deviations from the foundational paper (unvocalized input, Python forward-chaining, hard-wired rules, flat symbolic states) while extending it with novel components (Epicene Bridges, wildcard matching, human verification).

\medskip
\noindent\textbf{Significance}: This paper demonstrates the \textbf{practical feasibility} of the theoretical framework and provides a blueprint for future, fully compliant implementations.

\medskip
\noindent\textbf{Link to Previous Papers}: The prototype's morphological repository implements the grounding layer \( G \) from the foundational paper, while its Avicennian logic engine applies the modal syllogistic reasoning discussed in Principle III.




\subsection{Summary of the Five Stages}

Table \ref{tab:five_stages} summarizes the five stages of the research program, identifying the corresponding paper, its core contribution, and the key theorem established.

\begin{table}[H]
\centering
\begin{tabular}{|p{2cm}|p{2.5cm}|p{5.5cm}|p{3.5cm}|}
\hline
\textbf{Stage} & \textbf{Paper} & \textbf{Core Contribution} & \textbf{Key Theorem} \\
\hline
I & Sabr wa al-Taqsim & Extracts minimal rules from truth tables: \( S, I, W \) & \( h = S \wedge I^* \) \\
II & No Meaning & Refutes arbitrary sign; Arabic motivated at Levels W and M & Structural Impossibility of Arbitrary Sign \\
III & Foundational & Six principles; \( F = D \circ A \circ G \); Algorithm \( \Omega \) & Foundational Adequacy \\
IV & Al-Ghasali (FSA) & FSA compilation; \( O(1) \) verdict extraction; LLM safety & Finite-State Collapse (Lemma 1) \\
V & Avicenna Prototype & Neuro-symbolic prototype; wildcard matching; Epicene Bridges & Epistemological Soundness of LLM Integration \\
\hline
\end{tabular}
\caption{The five stages of the Computational Fiqh research program, their core contributions, and their key theorems.}
\label{tab:five_stages}
\end{table}

\subsection{The Intellectual Evolution Across the Papers}

Table \ref{tab:evolution} traces the intellectual evolution across the five papers, showing how each investigation addresses a distinct primary problem using a specific methodology and yielding a key theorem.

\begin{table}[H]
\centering
\begin{tabular}{|p{2.2cm}|p{2.5cm}|p{2.5cm}|p{3.5cm}|}
\hline
\textbf{Paper} & \textbf{Primary Problem} & \textbf{Methodology} & \textbf{Key Theorem} \\
\hline
Sabr wa al-Taqsim & Inductive rule extraction & Set-theoretic algorithm & \( h = S \wedge I^* \) \\
No Meaning & Linguistic foundation & Formal morphology + Kolmogorov complexity & Structural Impossibility of Arbitrary Sign \\
Foundational & Complete formalization & Six principles + stratified logic & Foundational Adequacy \\
Al-Ghasali (FSA) & Backend execution efficiency & Automata theory + constrained decoding & Finite-State Collapse (Lemma 1) \\
Avicenna Prototype & Pragmatic implementation & Neuro-symbolic engineering & Epistemological Soundness of LLM Integration \\
\hline
\end{tabular}
\caption{The intellectual evolution across the five papers, showing the primary problem, methodology, and key theorem of each investigation.}
\label{tab:evolution}
\end{table}

\subsection{The Unified Architecture}

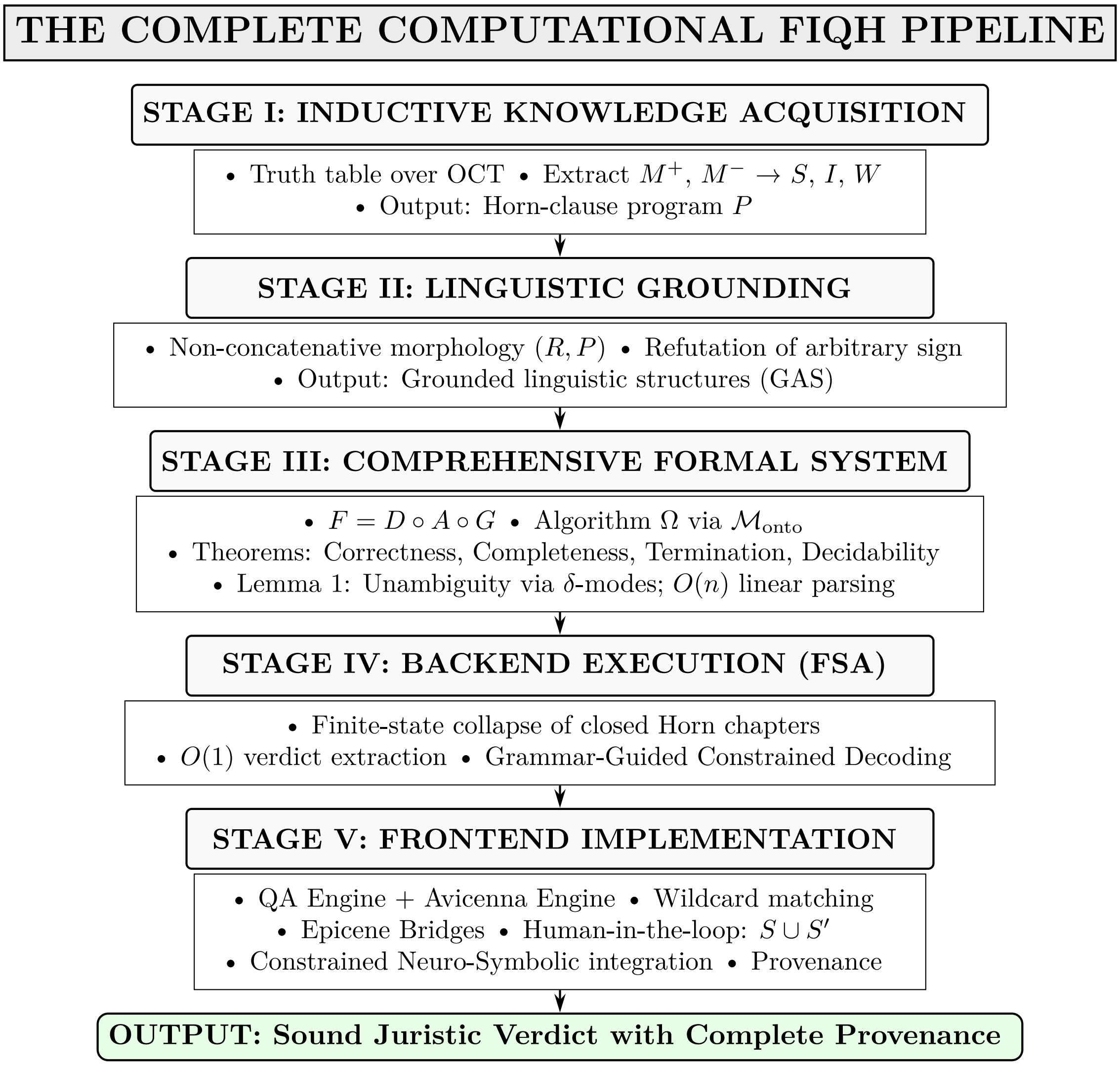
\begin{figure}[htbp]
\centering
\begin{tikzpicture}[
    node distance=0.3cm and 0.4cm,
    box/.style={
        rectangle,
        draw=black,
        thick,
        minimum width=10cm,
        minimum height=0.8cm,
        align=center,
        font=\small\bfseries,
        rounded corners=2pt,
        fill=gray!5
    },
    subbox/.style={
        rectangle,
        draw=black,
        thin,
        minimum width=8.5cm,
        minimum height=0.6cm,
        align=center,
        font=\footnotesize,
        fill=white
    },
    arrow/.style={
        -{Stealth[length=2.5mm, width=1.5mm]},
        thick
    },
    titlebox/.style={
        rectangle,
        draw=black,
        thick,
        minimum width=10cm,
        minimum height=0.6cm,
        align=center,
        font=\large\bfseries,
        fill=gray!15
    },
    outputbox/.style={
        rectangle,
        draw=black,
        thick,
        minimum width=7cm,
        minimum height=0.6cm,
        align=center,
        font=\small\bfseries,
        fill=green!10,
        rounded corners=4pt
    }
]

\node[titlebox] (title) {THE COMPLETE COMPUTATIONAL FIQH PIPELINE};

\node[box, below=0.3cm of title] (stage1) {
    \textbf{STAGE I: INDUCTIVE KNOWLEDGE ACQUISITION}
};
\node[subbox, below=0.05cm of stage1] (stage1details) {
    \begin{tabular}{c}
    • Truth table over OCT • Extract $M^+$, $M^-$ → $S$, $I$, $W$ \\
    • Output: Horn-clause program $P$
    \end{tabular}
};

\draw[arrow] (stage1details.south) -- ++(0,-0.2cm) -- ++(0,-0.1cm);

\node[box, below=0.3cm of stage1details] (stage2) {
    \textbf{STAGE II: LINGUISTIC GROUNDING}
};
\node[subbox, below=0.05cm of stage2] (stage2details) {
    \begin{tabular}{c}
    • Non-concatenative morphology $(R, P)$ • Refutation of arbitrary sign \\
    • Output: Grounded linguistic structures (GAS)
    \end{tabular}
};

\draw[arrow] (stage2details.south) -- ++(0,-0.2cm) -- ++(0,-0.1cm);

\node[box, below=0.3cm of stage2details] (stage3) {
    \textbf{STAGE III: COMPREHENSIVE FORMAL SYSTEM}
};
\node[subbox, below=0.05cm of stage3] (stage3details) {
    \begin{tabular}{c}
    • $F = D \circ A \circ G$ • Algorithm $\Omega$ via $\mathcal{M}_{\text{onto}}$ \\
    • Theorems: Correctness, Completeness, Termination, Decidability \\
    • Lemma 1: Unambiguity via $\delta$-modes; $O(n)$ linear parsing
    \end{tabular}
};

\draw[arrow] (stage3details.south) -- ++(0,-0.2cm) -- ++(0,-0.1cm);

\node[box, below=0.3cm of stage3details] (stage4) {
    \textbf{STAGE IV: BACKEND EXECUTION (FSA)}
};
\node[subbox, below=0.05cm of stage4] (stage4details) {
    \begin{tabular}{c}
    • Finite-state collapse of closed Horn chapters \\
    • $O(1)$ verdict extraction • Grammar-Guided Constrained Decoding
    \end{tabular}
};

\draw[arrow] (stage4details.south) -- ++(0,-0.2cm) -- ++(0,-0.1cm);

\node[box, below=0.3cm of stage4details] (stage5) {
    \textbf{STAGE V: FRONTEND IMPLEMENTATION}
};
\node[subbox, below=0.05cm of stage5] (stage5details) {
    \begin{tabular}{c}
    • QA Engine + Avicenna Engine • Wildcard matching \\
    • Epicene Bridges • Human-in-the-loop: $S \cup S'$ \\
    • Constrained Neuro-Symbolic integration • Provenance
    \end{tabular}
};

\draw[arrow] (stage5details.south) -- ++(0,-0.2cm) -- ++(0,-0.1cm);

\node[outputbox, below=0.3cm of stage5details] (output) {
    \textbf{OUTPUT: Sound Juristic Verdict with Complete Provenance}
};

\end{tikzpicture}
\caption[\parbox{\textwidth}{\footnotesize The Complete Computational Fiqh Pipeline: A unified view of the five-stage research program.}]
{\parbox{\textwidth}{\footnotesize The Complete Computational Fiqh Pipeline: A unified view of the five-stage research program. Stage I: Inductive knowledge acquisition (Sabr wa al-Taqsim). Stage II: Linguistic grounding (No Meaning). Stage III: Comprehensive formal system (Foundational). Stage IV: Backend execution via FSA (Al-Ghasali). Stage V: Frontend implementation (Avicenna Prototype). Each stage feeds into the next, culminating in a sound juristic verdict with complete provenance chain.}}
\label{fig:pipeline}
\end{figure}

When viewed collectively, the five papers constitute a complete, end-to-end computational pipeline for Islamic jurisprudence. Figure \ref{fig:pipeline} illustrates the complete pipeline, showing the progression from inductive knowledge acquisition through linguistic grounding, formalization, backend execution, and frontend implementation to the final juristic verdict.
The pipeline progresses through five distinct stages. Stage I (Sabr wa al-Taqsim) performs inductive knowledge acquisition by extracting minimal rules from truth tables. Stage II (No Meaning) provides the linguistic and epistemological foundation by formalizing Arabic morphology and refuting the arbitrary sign assumption. Stage III (Foundational) synthesizes these insights into a comprehensive formal system with rigorous mathematical guarantees. Stage IV (Al-Ghasali / FSA) solves the backend execution problem by compiling rules into deterministic finite-state automata with \( O(1) \) verdict extraction and mathematically guaranteed LLM safety. Stage V (Avicenna Prototype) delivers a pragmatic frontend implementation that handles real-world, unvocalized Arabic input through neuro-symbolic integration. Each stage feeds into the next, culminating in a sound juristic verdict with complete provenance chain.

\renewcommand{\thesubsection}{\arabic{subsection}}
\renewcommand{\thesubsubsection}{\arabic{subsection}.\arabic{subsubsection}}

\part*{Part I: Principles}

\setcounter{section}{0}

\section{Informal Foundations}

\subsection{Foundational Principle I: The Epistemological Nature of Computational Fiqh}

\subsubsection{Motivation}
Every formal computational system necessarily rests upon explicit epistemological assumptions concerning the nature of the discipline it seeks to formalize. Mathematics, physics, linguistics, law, and theology each investigate fundamentally different classes of objects and therefore require different computational paradigms. Before constructing a formal system for Islamic jurisprudence, it is therefore necessary to identify the precise scientific nature of Fiqh and to distinguish it from neighboring disciplines with which it is frequently conflated.

Within contemporary academic literature, computational treatments of Islamic texts generally inherit one of two epistemological paradigms. The first places Islamic jurisprudence under the broader category of theology or religious studies. The second treats Islamic legal texts as ordinary natural-language corpora interpreted according to conventional linguistic assumptions regarding the arbitrariness of the linguistic sign. As we argue below, neither paradigm adequately captures the epistemological structure of classical Fiqh, and both lead to computational assumptions fundamentally different from those adopted in the present work.

\subsubsection{Fiqh and Theology}
The first distinction concerns the relationship between Islamic jurisprudence and theology. Theology (Kalām) and jurisprudence (Fiqh) constitute two distinct sciences within the classical Islamic intellectual tradition. Although both ultimately derive their authority from the same revealed sources, they investigate fundamentally different questions. Theology is primarily concerned with the epistemological foundations of religion itself, including questions concerning revelation, prophethood, divine attributes, metaphysical doctrines, and the rational justification of religious belief. Islamic theological reasoning is therefore not merely speculative or dogmatic. Classical scholars developed sophisticated rational methodologies intended to establish these foundational principles through systematic argumentation. Among these methodologies is the principle of incapability (burhān al-’ajz), whereby the persistent inability of all admissible natural explanations to account for a phenomenon is taken as rational evidence that the phenomenon belongs to a fundamentally different explanatory category. Within Islamic thought this principle has historically been employed in discussions concerning the divine origin of the Qur’an, while analogous forms of reasoning based upon principled impossibility or explanatory incapability also appear in several areas of modern philosophy of science and theoretical physics. A formal treatment of these epistemological methodologies, however, belongs to Kalām and lies outside the scope of the present work.

The present paper therefore assumes, rather than attempts to establish, the epistemic authority of the accepted sources of Islamic law. Our objective begins only after these sources have been accepted and concerns the computational problem of deriving legal rulings from them.

\subsubsection{The Conventional Linguistic Paradigm}
A second computational assumption inherited by many contemporary AI systems originates in influential traditions of modern structural linguistics. Beginning with Saussure’s principle of the arbitrariness of the linguistic sign, numerous linguistic and computational frameworks assume that the relationship between linguistic form (the signifier) and meaning (the signified) is fundamentally conventional rather than structurally determined. Under this assumption, lexical semantics is recovered primarily through contextual interpretation, statistical association, and language usage, while morphology contributes comparatively little formally recoverable semantic information.

Modern Large Language Models naturally inherit this assumption. Their semantic representations are learned statistically from large corpora, treating morphology primarily as an aid to tokenization and distributional prediction rather than as an independent source of deductively recoverable semantic structure [9].

The present work adopts a different assumption for Classical Arabic. Building upon the constructive linguistic framework established in our previous work [3], we assume that substantial portions of lexical and grammatical meaning are systematically encoded within the morphological organization of the language itself. Roots, patterns, particles, and grammatical connectors therefore constitute algorithmically exploitable semantic structures rather than merely historical linguistic artifacts. Consequently, morphological analysis is not merely a preprocessing stage but an essential component of logical grounding.

\subsubsection{The Computational-Normative Paradigm}
Having distinguished the present work from both theological formalization and conventional statistical language processing, we now identify the epistemological model adopted throughout this paper.

The objective of Fiqh is neither the establishment of theological doctrine nor the statistical interpretation of language. Its objective is the derivation of normative legal judgments governing concrete human actions under specified circumstances. Consequently, every juristic deduction depends simultaneously upon two fundamentally different categories of knowledge. The first consists of empirically established facts describing the actual state of the world. Depending upon the legal question under consideration, these facts may originate from linguistics, medicine, astronomy, economics, commerce, finance, geography, forensic sciences, or other empirical disciplines. Typical examples include determining whether water is available, whether a medical condition legally constitutes hardship, whether the prayer time has entered, whether a commercial transaction satisfies the legal definition of usury, or whether the conditions of ritual purity have been fulfilled.

The second consists of formal deductive reasoning, whereby these empirically established premises are transformed into normative legal conclusions according to the accepted principles of Islamic jurisprudence.

This distinction is not merely methodological; it reflects the epistemological structure of Fiqh itself. Logic does not determine empirical reality. Rather, it determines the legal consequences of empirically established reality.

Accordingly, any computational architecture for Islamic jurisprudence must explicitly distinguish between these two epistemologically distinct processes.

\begin{principle}[Epistemological Foundation of Computational Fiqh]
Islamic jurisprudence is a normative deductive science whose application presupposes empirically established facts and formally accepted legal sources. It is therefore distinct both from Kalām, whose objective is the rational establishment of the epistemological foundations of religion, and from computational linguistic models that treat semantic grounding primarily as a statistical inference problem.
\end{principle}

Consequently, every computational formalization of Fiqh must explicitly separate (i) empirical grounding of legally relevant facts—including linguistic and domain-specific scientific knowledge—from (ii) formal deduction of normative legal consequences. This epistemological separation constitutes the first foundational assumption upon which the remainder of the present formal system is constructed.

\subsection{Foundational Principle II: The Legitimacy of Algorithmic Reasoning in Islamic Jurisprudence}

\subsubsection{Motivation}
Having established that Islamic jurisprudence is a normative deductive science operating upon empirically established facts, the next fundamental question naturally arises: Is it methodologically legitimate to formalize juristic reasoning as an algorithmic process? This question concerns neither artificial intelligence nor computer science alone. It concerns the methodological foundations of Islamic jurisprudence itself. If the inferential procedures employed by jurists are inherently algorithmic, then their computational formalization represents a continuation of an established scientific tradition. Conversely, if juristic reasoning is fundamentally intuitive or non-constructive, then any attempt at formal computational implementation would necessarily fail.

The present work argues that algorithmic reasoning is not an external methodology imported into Islamic jurisprudence from modern computer science. Rather, it has constituted one of the defining characteristics of juristic methodology since the formative period of the Islamic sciences.

\subsubsection{Historical Foundations of Algorithmic Jurisprudence}
The historical relationship between Islamic jurisprudence and algorithmic reasoning is far older than the emergence of modern computation. Among the earliest systematic computational methods developed within Islamic civilization are the constructive procedures introduced by Muḥammad ibn Mūsā al-Khwārizmī. His development of algebra (al-jabr) and systematic computational procedures was motivated in significant part by practical legal problems, including inheritance distribution (farā’iḍ), financial transactions, commercial accounting, and other jurisprudential calculations requiring deterministic and reproducible procedures [5, 17].

The modern term algorithm itself derives from the Latinized form of al-Khwārizmī’s name, reflecting the historical origin of constructive computation within this scientific tradition.

This historical fact carries an important methodological implication. Computational procedures were not introduced into jurisprudence by modern computer science; rather, jurisprudential requirements themselves historically motivated the development of formal algorithmic methods. The present work therefore views computational formalization not as the introduction of a new methodology into Fiqh, but as the continuation of an established algorithmic tradition using modern formal languages.

\subsubsection{The Methodological Status of Logic}
The legitimacy of formal reasoning within Islamic jurisprudence has likewise received extensive treatment throughout the classical Islamic intellectual tradition. Early scholarly discussions primarily concerned the conditions under which logical reasoning (manṭiq) could legitimately be employed within juristic methodology. Over time, a broad methodological consensus emerged that logic constitutes an instrument (ālah) for preserving the correctness of reasoning rather than an independent source of legislation.

The significance of this distinction cannot be overstated. Logic neither establishes revelation nor creates legal rulings. It merely guarantees that accepted premises are transformed into conclusions according to valid rules of inference. Accordingly, logic occupies within Islamic jurisprudence a role analogous to that of mathematics within the physical sciences. It is not itself a source of empirical knowledge; rather, it provides the formal framework through which established knowledge may be manipulated consistently and without contradiction.

The computational formalization proposed in the present work should therefore be understood as a formalization of juristic methodology rather than a replacement for juristic authority.

\subsubsection{Algorithmic Juristic Reasoning: Induction and Deduction}
A common misunderstanding in computational treatments of legal reasoning is to identify jurisprudence exclusively with deductive inference. Classical Islamic jurisprudence, however, employs two fundamentally different classes of constructive inference, each serving a distinct epistemological function.

The first is inductive reasoning, whose objective is the discovery, validation, and generalization of legal principles from the primary sources. 
Methodologies such as Qiyās, al-Ṣabr wa al-Taqsīm, Takhrīj al-Manāṭ, and Tanqīḥ al-Manāṭ belong to this category [10, 12]. Their objective is not the application of existing rules but the identification of the effective legal cause (‘illah) and the construction of new general legal rules applicable beyond the original textual cases. Computationally, this stage constitutes knowledge acquisition. It transforms revealed texts into formally representable legal principles.

The second is deductive reasoning, whose objective is the application of these previously established legal principles to concrete factual situations. Once legal rules have been identified and formally represented, deductive inference derives their normative consequences for particular cases. This is the role performed by formal logical systems, including the Avicennian modal Horn framework developed in the present work. Computationally, this stage constitutes knowledge application.

The distinction between these two forms of reasoning is fundamental. Induction generates legal rules. Deduction applies legal rules. Neither can replace the other. Deduction cannot derive legal rules that are absent from the knowledge base, while induction alone cannot determine the legal consequences of newly observed factual situations.

Consequently, a complete computational theory of Islamic jurisprudence necessarily requires both inductive and deductive formalization. The present paper addresses the deductive component of this program, while our previous work on al-Ṣabr wa al-Taqsīm provides a formal computational treatment of the inductive component responsible for extracting legal causes and constructing the corresponding rule base [2]. Together, these two complementary systems constitute a unified computational methodology for Islamic jurisprudence.

\subsubsection{Computational Consequences}
Recognizing the dual inductive-deductive structure of Fiqh fundamentally changes the objectives of computational jurisprudence. The goal is not merely to automate legal deduction. Rather, the complete computational problem consists of two sequential stages:
\begin{itemize}
    \item Inductive knowledge acquisition, whereby legal rules are systematically extracted and generalized from the primary sources through established methodologies of Uṣūl al-Fiqh.
    \item Deductive knowledge application, whereby these formalized rules are applied to empirically grounded legal facts to derive normative rulings.
\end{itemize}

The correctness of a computational jurisprudential system therefore depends upon the correctness of both stages. An implementation addressing only deductive reasoning formalizes only one half of the juristic methodology developed within the classical Islamic tradition.

\begin{principle}[Algorithmic Juristic Reasoning]
Islamic jurisprudence is inherently algorithmic, but its methodology is not purely deductive. Classical juristic reasoning comprises two complementary classes of constructive inference: induction, responsible for discovering, validating, and generalizing legal principles from the primary sources through methodologies such as Qiyās and al-Ṣabr wa al-Taqsīm, and deduction, responsible for applying these established principles to concrete factual situations. Consequently, a complete computational formalization of Fiqh must formalize both inductive rule acquisition and deductive rule application.
\end{principle}

The present work addresses the deductive component of this framework, while building upon our previous formalization of the inductive component, thereby completing the two principal computational stages of juristic reasoning.

\subsection{Foundational Principle III: Engineering the Logical Foundation of Computational Islamic Jurisprudence}

\subsubsection{Motivation}
The preceding foundational principles established both the epistemological nature of Islamic jurisprudence and the legitimacy of its computational formalization. Principle I identified Fiqh as a normative science operating upon empirically established facts, while Principle II demonstrated that its methodology consists of complementary inductive and deductive computational procedures. The remaining question concerns the logical foundation upon which the deductive component should be constructed.

At first sight, this question appears to require selecting one of the existing logical systems developed throughout the history of philosophy, mathematics, or computer science. Classical Aristotelian logic, Avicennian logic, first-order predicate logic, modal logic, Horn logic, Datalog, and numerous other formalisms each provide mature theories of formal reasoning and have all been successfully applied within their respective domains.

The present work, however, adopts a different perspective. The objective is not to determine which existing logical system is “the best.” Formal systems are not natural phenomena waiting to be discovered; they are scientific artifacts deliberately designed to solve particular classes of reasoning problems. Throughout the history of logic, new formal systems have repeatedly been introduced whenever existing systems proved inadequate for newly emerging scientific requirements. Predicate logic extended syllogistic logic, modal logics extended classical logic, Horn logic restricted first-order logic to obtain efficient computation, and Datalog further specialized Horn logic for database reasoning. Each of these developments illustrates the same fundamental principle: formal systems are engineered to satisfy the requirements of a particular application domain.

Computational Islamic jurisprudence presents its own distinctive requirements. It combines normative reasoning, conditional legal rules, inductive rule acquisition, deductive rule application, Arabic linguistic semantics, and the practical necessity of efficient automated inference. There is no reason to expect that a formal system originally designed for philosophical argumentation, mathematical proof, or database processing should satisfy all of these requirements simultaneously.

Accordingly, the objective of the present section is not merely to compare existing logical systems. Rather, it is to identify the essential requirements imposed by computational Islamic jurisprudence, evaluate how existing logical traditions address those requirements, extract their useful characteristics, identify their limitations, and finally derive the design principles of a new formal system specifically engineered for computational Fiqh.

\subsubsection{Scientific Problem}
The engineering problem addressed by the present work may now be formulated precisely. What properties must a formal logical system possess in order to serve as an adequate foundation for computational Islamic jurisprudence?

This formulation deliberately differs from the more common question: Which existing logical system should be selected? The distinction is fundamental. The latter assumes that one of the existing logical systems already provides an adequate solution. The former recognizes that existing logical systems were designed for different scientific purposes and therefore serve primarily as sources of ideas rather than as predetermined solutions.

Consequently, the objective of the present investigation is constructive rather than comparative. Existing logical systems are analyzed in order to determine:
\begin{itemize}
    \item which desirable properties they contribute,
    \item which requirements they fail to satisfy,
    \item and which additional mechanisms must be introduced to obtain a formal system capable of supporting computational Islamic jurisprudence.
\end{itemize}

The resulting logical architecture is therefore expected to inherit useful properties from several existing traditions while simultaneously introducing novel components motivated by the specific requirements of Fiqh.

\subsubsection{Engineering Requirements}
Before examining existing logical systems, it is necessary to establish the engineering requirements that any satisfactory formalization of Islamic jurisprudence must satisfy. These requirements naturally divide into two categories.

\textbf{A. Semantic Requirements} The formal system must faithfully represent the conceptual structure of Islamic jurisprudence itself. In particular, it should provide:
\begin{itemize}
    \item sufficient expressive power to represent legal entities, relations, conditions, contracts, obligations, permissions, prohibitions, ownership, witnesses, inheritance relations, and other jurisprudential concepts;
    \item native support for conditional reasoning, since legal rules are predominantly formulated as conditional propositions rather than categorical assertions;
    \item explicit representation of normative modalities, including obligation, recommendation, permissibility, prohibition, validity, invalidity, exemption, and related legal statuses;
    \item representation of legal causes (‘ilal), conditions (shurūṭ), impediments (mawāni‘), exceptions, and contextual qualifications central to Uṣūl al-Fiqh [10, 12];
    \item compatibility with the Arabic semantic meta-level introduced in the present work, allowing morphological and grammatical information to contribute directly to semantic interpretation;
    \item natural correspondence with the conceptual organization employed by jurists, thereby minimizing artificial encodings and facilitating scholarly verification.
\end{itemize}

\textbf{B. Computational Requirements} The formal system must also support practical automated reasoning. Accordingly, it should provide:
\begin{itemize}
    \item deterministic executable inference procedures;
    \item guaranteed termination whenever possible;
    \item computational complexity compatible with practical legal expert systems;
    \item scalability to large jurisprudential knowledge bases;
    \item direct compatibility with the inductive rule extraction methodologies developed in computational Uṣūl al-Fiqh;
    \item modularity, allowing future extensions without modifying the underlying semantic architecture.
\end{itemize}

\subsubsection{Evaluation of Existing Logical Frameworks}
The objective of this section is not to determine whether one logical system is universally superior to another. Every formal system represents an engineering compromise between expressive power, computational complexity, semantic richness, and ease of implementation.

\begin{table}[htbp]
\centering
\caption{Summary of Design Requirements}
\renewcommand{\arraystretch}{1.2} 
\begin{tabular}{p{1.2in} p{2.3in} p{2.8in}}
\toprule
Category & Requirement & Purpose \\
\midrule
Semantic & Expressive Power & Represent jurisprudential concepts naturally \\
Semantic & Conditional Reasoning & Represent legal implication directly \\
Semantic & Normative Modalities & Represent legal status explicitly \\
Semantic & Legal Causes and Exceptions & Capture the structure of Uṣūl al-Fiqh \\
Semantic & Arabic Semantic Compatibility & Integrate linguistic semantic information \\
Semantic & Natural Juristic Representation & Preserve juristic conceptual organization \\
Computational & Computational Executability & Support constructive automated reasoning \\
Computational & Computational Termination and Decidability & Ensure reliable inference \\
Computational & Computational Efficiency & Permit practical deployment \\
Computational & Scalability & Handle large knowledge bases \\
Computational & Compatibility with Inductive Rule Extraction & Integrate with computational Uṣūl al-Fiqh \\
Computational & Modularity & Facilitate future extensions \\
\bottomrule
\end{tabular}
\end{table}

\paragraph*{Propositional Logic}
Propositional logic represents the simplest formal deductive system in widespread use. Its atomic objects are complete propositions connected by truth-functional operators such as conjunction, disjunction, implication and negation. Because of its simplicity, propositional logic has become one of the principal foundations of automated reasoning. Modern SAT solvers routinely solve propositional reasoning problems containing millions of variables.

Jurisprudential Example: Suppose a jurist formulates the following rule: “Every traveller who cannot use water is permitted to perform Tayammum.” A jurist immediately understands this as a universal legal principle. Within propositional logic, however, variables do not exist. Consequently one must instead write:
\begin{verbatim}
Traveller Ahmed AND NoWater Ahmed -> Tayammum Ahmed
Traveller Fatima AND NoWater Fatima -> Tayammum Fatima
Traveller Khalid AND NoWater Khalid -> Tayammum Khalid
...
\end{verbatim}
The legal principle disappears. Only individual instances remain.

Limitations: These computational advantages are obtained only by sacrificing expressive power. Variables standing for domain objects cannot be represented. General legal rules cannot be expressed naturally. Relations between individuals cannot be represented. Quantification is impossible. Legal concepts such as ownership, kinship, guardianship, inheritance relationships, contractual obligations and witnesses require artificial encoding. Consequently, the resulting knowledge base grows according to the number of individuals rather than according to the number of legal principles. This is fundamentally incompatible with the methodology of Islamic jurisprudence, whose objective is to formulate general rules applicable to arbitrarily many concrete situations.

\paragraph*{First-Order Predicate Logic}
First-order predicate logic was developed to overcome the principal expressive limitations of propositional logic. By introducing variables, predicates, relations and quantification, predicate logic transformed formal reasoning from isolated propositions into reasoning about general structures.

Jurisprudential Example: The previous legal rule now becomes $\forall x Traveller(x) \wedge UnableToUseWater(x) \rightarrow Permitted(Tayammum, x)$. Unlike propositional logic, this formulation directly represents the general legal principle rather than individual instances.

Limitations: The increase in expressive power is accompanied by increased computational complexity. General first-order theorem proving is semi-decidable. Proof search may continue indefinitely whenever no proof exists. Unrestricted quantification dramatically enlarges the search space explored by automated reasoning systems. Function symbols permit the construction of infinitely many terms. Consequently, although first-order logic represents legal knowledge naturally, it does not provide the computational guarantees expected of practical legal expert systems. Furthermore, normative concepts such as obligation, prohibition, validity, exemption and legal effect continue to be represented indirectly through ordinary predicates rather than through native jurisprudential constructs.

\paragraph*{Higher-Order Logic}
Higher-order logic extends first-order predicate logic by permitting quantification not only over individuals but also over predicates, relations, functions, and sets.

Jurisprudential Example: Consider the statement: “Every legal cause whose conditions are satisfied produces its corresponding legal ruling.” Within higher-order logic one may quantify directly over legal predicates: $\forall C (LegalCause(C) \rightarrow Applicable(C))$ where $C$ itself denotes a legal relation rather than an individual object.

Limitations: The expressive power of higher-order logic comes at a considerable computational price. Automated reasoning becomes substantially more difficult than in first-order logic. General higher-order theorem proving lacks the computational guarantees expected of practical expert systems. Proof search spaces become extremely large, and complete automated reasoning procedures are unavailable in the general case.

\paragraph*{Modern Modal Logics}
One of the principal limitations of classical mathematical logic is its treatment of propositions solely in terms of truth and falsity. Statements involving necessity, possibility, obligation, belief, knowledge, time, or uncertainty require richer semantic structures. To address these limitations, modern logic introduced modal systems.

Jurisprudential Example: Consider the legal proposition “Prayer is obligatory.” A modal formalization might express this as $O(Prayer)$ where $O$ denotes obligation. Similarly, “It is permissible to perform Tayammum” may be represented as $P(Tayammum)$.

Limitations: Despite these advantages, an important distinction must be recognized. The modalities developed within modern modal logic were designed primarily for general philosophical reasoning. Their semantic interpretation is based upon accessibility relations between possible worlds. Islamic jurisprudence, however, employs modalities of a different nature. Legal obligation is not merely metaphysical necessity. Legal permission is not merely logical possibility. Legal validity is not simply truth in another accessible world. Instead, these modalities derive from normative relationships established through revelation, legal causes, contextual conditions, evidentiary standards, and methodological principles developed within Uṣūl al-Fiqh. Consequently, the semantic interpretation provided by general modal logic does not directly correspond to the conceptual organization of Islamic jurisprudence.

\paragraph*{Horn Logic and Datalog}
While first-order predicate logic provides a powerful language for representing knowledge, its unrestricted expressive power leads to considerable computational difficulties. Horn logic emerged as a restricted fragment of first-order predicate logic specifically designed to overcome these computational limitations. By restricting formulas to Horn clauses, efficient inference procedures such as forward chaining, backward chaining, and SLD-resolution become possible. Building upon these ideas, Datalog introduced additional syntactic restrictions—most notably the elimination of function symbols—in order to guarantee finite domains and terminating inference procedures [8, 15].

Jurisprudential Example: The legal rule “Every traveller who cannot use water is permitted to perform Tayammum” may naturally be represented as:
\begin{verbatim}
Permitted(Tayammum,X) :- Traveller(X), UnableToUseWater(X).
\end{verbatim}
Given the facts \texttt{Traveller(Ahmad).} and \texttt{UnableToUseWater(Ahmad).}, the inference engine automatically derives \texttt{Permitted(Tayammum,Ahmad).}

Limitations: Despite these substantial advantages, Horn logic was never intended to serve as a complete semantic theory. Its principal objective is efficient computation rather than conceptual representation. Several important jurisprudential notions therefore remain external to the logical language itself. For example, legal obligation, legal validity, effective legal causes, conditions, exceptions, conflicting evidences, hierarchies of proofs, and linguistic semantic relations are not primitive concepts of Horn logic. Instead, they must be encoded indirectly through additional predicates. Similarly, Horn implication is purely syntactic. The rule \texttt{ObligatoryPrayer(X) :- Adult(X), Sane(X).} contains no intrinsic distinction between a causal relationship, a definitional relationship, an evidential relationship, or a jurisprudential obligation. All implications possess exactly the same logical status. From the perspective of Islamic jurisprudence, however, these distinctions are fundamental. Jurists distinguish carefully between legal causes (‘illah), conditions (sharṭ), impediments (māni‘), evidences (dalīl), legal objectives (maqāṣid), and numerous other semantic relationships. Horn logic itself does not distinguish among these categories.

\paragraph*{Aristotelian Logic}
Long before the emergence of modern mathematical logic, Aristotle developed the first systematic theory of formal reasoning. His syllogistic logic dominated philosophical thought for nearly two thousand years.

Limitations: Despite its historical significance, Aristotelian logic was not designed as a general computational language. Its primary inferential mechanism is the categorical syllogism rather than the conditional rule. Islamic jurisprudence, however, is overwhelmingly conditional. Legal reasoning typically proceeds through propositions of the form “If certain conditions hold, then a legal consequence follows.” Furthermore, Aristotelian logic provides relatively limited mechanisms for representing complex relational knowledge, multiple interacting conditions, exceptions, procedural rules, or executable inference. Its expressive resources therefore become insufficient for large-scale computational jurisprudence.

\paragraph*{Avicennian Logic}
Among the numerous logical traditions developed throughout history, Avicennian logic occupies a unique position with respect to Islamic jurisprudence. While deeply rooted in the Aristotelian tradition, Avicenna did not merely preserve Aristotle’s logical system. Instead, he fundamentally reorganized and extended it, producing a substantially richer theory of reasoning that gradually replaced Aristotelian logic throughout much of the Islamic intellectual world [6, 21].

Conditional Reasoning: One of the most significant innovations introduced by Avicenna concerns the status of conditional propositions. Within Aristotelian syllogistic logic, categorical propositions occupy the central position. Avicenna, by contrast, recognized that scientific reasoning frequently proceeds through conditional statements expressing relationships between circumstances and their consequences. This observation possesses immediate relevance for Islamic jurisprudence. Legal rules are rarely formulated as isolated categorical propositions. Instead, they almost invariably possess the following general structure: “If certain legally relevant conditions are satisfied, then a particular legal consequence follows.” For example, “If water cannot be used, then Tayammum becomes permissible.” Likewise, “If the prescribed amount of wealth reaches the legal threshold, then Zakat becomes obligatory.”

Causal Structure: A second characteristic shared by Avicennian logic and Islamic jurisprudence concerns the importance of causal explanation. Classical jurists rarely ask merely whether a legal ruling is true. Instead, they seek to identify the legal cause (‘illah), the relevant conditions (shurūṭ), possible impediments (mawāni‘), and the circumstances under which the ruling applies. The objective is therefore explanatory rather than merely deductive.

Limitations: Despite these substantial advantages, Avicennian logic cannot simply be adopted unchanged as the logical foundation of computational jurisprudence. Its primary objective remained philosophical and scientific reasoning rather than executable computation. Consequently, it does not define formal inference procedures suitable for automated implementation. Its semantics, although conceptually rich, were not developed using the precise mathematical methods characteristic of modern formal logic. Likewise, it provides neither explicit computational complexity analyses nor algorithmic execution models comparable to those available for Horn logic or Datalog. Most importantly, Avicennian logic predates modern computer science by nearly a millennium. Questions concerning deterministic execution, algorithmic termination, indexing strategies, rule compilation, complexity analysis, and scalable automated inference naturally lie outside its historical scope.

\subsubsection{Synthesis of Design Principles from Existing Logical Traditions}
The preceding sections examined several of the principal logical traditions developed throughout the history of philosophy, mathematics, and computer science. Each system was originally engineered to solve a particular class of reasoning problems and therefore reflects different design priorities. This historical diversity leads to an important observation: No existing logical system simultaneously satisfies all of the semantic and computational requirements established in Section 3.3.

The objective is therefore not to select one historical logic and apply it unchanged. Instead, the engineering task consists of identifying those principles developed within the existing traditions that remain valuable, determining the limitations that prevent their direct application, and synthesizing these insights into a new formal system specifically designed for computational jurisprudence.

Design Principles Inherited:
\begin{itemize}
    \item Propositional Logic: Efficient executable inference should remain a primary engineering objective.
    \item Predicate Logic: Relational representation, variables, general legal rules, quantified knowledge.
    \item Higher-Order Logic: Semantic architecture capable of representing multiple conceptual levels.
    \item Modern Modal Logic: Normative modalities should be represented explicitly, but their semantics must be jurisprudential rather than purely philosophical.
    \item Horn Logic and Datalog: Executable rule-based inference, constructive proof procedures, deterministic computation, scalable rule execution, efficient forward and backward chaining.
    \item Aristotelian Logic: Commitment to rigorous deductive validity while replacing purely categorical reasoning by richer conditional structures.
    \item Avicennian Logic: Conditional propositions as the primary semantic unit, explicit distinction between legal causes, conditions, impediments, and consequences, explanation-oriented reasoning, methodology-centered semantics.
\end{itemize}

\subsubsection{Foundational Principle III: Engineering the Logical Foundation of Computational Islamic Jurisprudence}
The preceding analysis established that the design of a formal system for computational Islamic jurisprudence cannot be reduced to the selection of an existing logical formalism. Throughout the history of logic, every major formal system was developed to satisfy the requirements of a particular scientific domain. Aristotelian logic was engineered primarily for categorical deduction, first-order predicate logic for mathematical formalization, modal logics for reasoning about necessity and possibility, and Horn logic for efficient executable inference. Each of these traditions contributes valuable principles, yet none simultaneously satisfies all the semantic, linguistic, methodological, and computational requirements imposed by Islamic jurisprudence.

This observation leads to the following foundational principle.

\begin{principle}[Engineering Principle of the Logical Foundation]
Let $\mathcal{J}$ denote the body of Islamic jurisprudence (Fiqh) together with its methodological foundations (Uṣūl al-Fiqh). A formal system $\mathcal{F}$ constitutes an adequate logical foundation for computational Islamic jurisprudence if and only if its architecture is explicitly engineered to satisfy the semantic, linguistic, methodological, and computational requirements imposed by $\mathcal{J}$, rather than being derived solely from the structure of any pre-existing logical tradition.
\end{principle}

Accordingly, every adequate formal system for computational Islamic jurisprudence shall satisfy the following principles:
\begin{enumerate}
    \item Semantic Adequacy: The formal system shall faithfully represent the conceptual organization of Islamic jurisprudence, including legal rules, legal causes (‘ilal), conditions (shurūṭ), impediments (mawāni‘), exceptions, contextual qualifications, normative modalities, and the methodological concepts developed within Uṣūl al-Fiqh.
    \item Linguistic Grounding: The semantic interpretation of legal propositions shall remain compatible with the linguistic structure of Classical Arabic. Whenever formally recoverable semantic information is provided by morphology, syntax, lexical structure, or other linguistic regularities, such information shall participate directly in the semantic interpretation of legal propositions.
    \item Jurisprudential Naturalness: The representation of legal knowledge shall correspond as closely as possible to the conceptual organization employed by jurists. Artificial representations introduced solely for computational convenience should be avoided whenever they obscure the semantic relationships recognized within the classical jurisprudential tradition.
    \item Methodological Compatibility: The formal system shall support both principal components of jurisprudential reasoning. It shall provide a deductive mechanism for applying established legal rules while simultaneously remaining compatible with the inductive methodologies of Uṣūl al-Fiqh responsible for discovering, validating, and organizing those rules.
    \item Constructive Executability: The formal system shall admit explicit constructive inference procedures capable of computing legal conclusions through deterministic algorithmic execution.
    \item Computational Tractability: The computational mechanisms employed by the formal system shall remain sufficiently efficient to permit practical implementation over large jurisprudential knowledge bases.
    \item Independence of Realization: The validity of the formal system shall depend exclusively upon the correctness of its semantic foundations and formal inference principles, not upon any particular computational realization.
\end{enumerate}

Concluding Remark: Although the preceding principle establishes the logical requirements that any formalization of Islamic jurisprudence must satisfy, the resulting formal language must also faithfully represent the primary objects of jurisprudential reasoning themselves. Islamic jurisprudence is concerned not only with static legal facts but, more fundamentally, with the normative evaluation of human actions. Acts such as praying, fasting, buying, selling, marrying, divorcing, testifying, travelling, and countless other legally relevant actions constitute the central objects to which legal judgments are attached. Consequently, the formal language developed in the present work must provide explicit mechanisms for representing both static legal assertions and dynamic juridical actions.

\subsection{Foundational Principle IV: Epistemic Grounding of Computational Islamic Jurisprudence}

\subsubsection{Motivation}
The preceding foundational principles established the epistemological nature of Islamic jurisprudence, the methodological interaction between induction and deduction, and the logical requirements imposed upon any formal system intended to support computational jurisprudence. An equally fundamental question nevertheless remains unanswered. Logical deduction does not operate directly upon reality. Instead, it operates upon formally represented facts. Consequently, before any legal inference may be performed, the factual circumstances of the case under consideration must first be transformed into a formal representation suitable for logical reasoning.

This transformation is frequently overlooked within existing work on computational law and logic programming, where factual assertions are commonly assumed to be available a priori. Rules are written over predicates such as \texttt{Traveller(Ahmad)}, \texttt{Adult(Fatimah)}, or \texttt{UnableToUseWater(Ahmad)} without addressing the preceding scientific problem: How were these facts established in the first place?

For computational Islamic jurisprudence, this question is indispensable. Unlike mathematical theorem proving, jurisprudential reasoning begins with concrete descriptions of real situations expressed by human users in natural language. These descriptions must first be interpreted, analyzed, verified, and transformed into legally relevant facts before deductive reasoning may begin. The present work therefore recognizes epistemic grounding as an independent computational process preceding formal deduction.

\subsubsection{Scientific Problem}
The central question addressed by the present principle may therefore be formulated as follows: How are formally represented legal facts obtained from empirical reality before deductive reasoning begins? This question is fundamentally different from logical inference itself. Deduction computes new conclusions from existing facts. Epistemic grounding computes the facts upon which deduction subsequently operates. The two computational processes therefore perform distinct scientific functions and should not be conflated.

\subsubsection{The Empirical Nature of Jurisprudential Facts}
As established in Foundational Principle I, Islamic jurisprudence is a normative science whose application depends upon empirically established facts. Legal rulings do not arise solely from textual interpretation. Rather, they arise from the interaction between revealed legal principles and accurately established descriptions of reality. Consequently, every legal consultation implicitly involves two complementary scientific activities. The first determines what is the factual situation. The second determines what legal ruling follows from that situation. The former belongs to empirical investigation. The latter belongs to jurisprudential reasoning. Accordingly, computational jurisprudence necessarily begins with the acquisition of empirically grounded facts rather than with deduction itself.

\subsubsection{Sources of Epistemic Grounding}
An important consequence immediately follows. The factual description supplied to the deductive engine does not originate from a single discipline. Instead, it may require the integration of knowledge obtained from numerous empirical sciences, depending upon the legal problem under consideration. These sources include, among others: Classical Arabic morphology, Classical Arabic syntax, lexical semantics, rhetorical analysis, authenticated textual sources, medicine, astronomy, economics, biology, geography, commercial practice, social custom (‘urf), and any additional empirical discipline relevant to the legal question under consideration. Computational Islamic jurisprudence should therefore be regarded as an integration framework within which multiple empirical sciences contribute formally represented facts to a common deductive methodology.

\subsubsection{The Foundational Role of Classical Arabic}
Although numerous empirical sciences contribute to jurisprudential reasoning, Classical Arabic occupies a unique epistemological position. Medicine contributes medical facts. Astronomy contributes astronomical facts. Economics contributes economic facts. Arabic, however, contributes something fundamentally different. It is the language in which the primary legal sources themselves are expressed. Consequently, Arabic is not merely an auxiliary discipline supporting jurisprudence. It constitutes the primary medium through which revelation is interpreted. Any computational misunderstanding of Arabic therefore directly produces a jurisprudential misunderstanding.

Furthermore, as demonstrated in our previous investigations concerning Arabic morphology and semantic computation [3], Arabic possesses systematic linguistic regularities through which semantic information may be recovered formally. Morphological derivation, root-pattern organization, grammatical structure, lexical relations, and other linguistic phenomena frequently encode semantic information that is directly relevant to legal interpretation. Accordingly, formally recoverable linguistic information should participate directly in the grounding process rather than being discarded before logical deduction begins. Classical Arabic therefore forms the primary epistemic grounding layer of the proposed computational architecture.

\subsubsection{Integration of Domain-Specific Empirical Sciences}
The same principle extends naturally to every empirical discipline involved in jurisprudential reasoning. Consider a legal consultation concerning the permissibility of postponing fasting because of illness. The legal ruling cannot be determined solely by logical deduction. Before deduction begins, medical science must establish whether the patient’s condition satisfies the medically relevant criteria recognized by the jurists. Only after these medical facts have been established may jurisprudential reasoning determine their legal consequences. Similarly, inheritance jurisprudence depends upon accurately established family relationships, commercial jurisprudence depends upon economic and contractual information, astronomical rulings depend upon astronomical observation, and numerous additional legal domains depend upon corresponding empirical sciences. In every case, the empirical discipline establishes the factual description of reality, while jurisprudence establishes the corresponding normative judgment.

\subsubsection{Separation Between Grounding and Deduction}
The preceding analysis leads to an important architectural consequence. The deductive engine should never communicate directly with raw user input. Instead, every legal consultation shall proceed through two logically distinct computational stages. The first stage transforms natural-language descriptions and empirical observations into formally represented legal facts. The second stage applies jurisprudential inference to those grounded facts. This separation significantly clarifies both the semantics and the correctness of computational jurisprudence. Errors arising during empirical fact acquisition remain distinguishable from errors arising during legal deduction. Likewise, improvements in empirical grounding methods do not require modification of the deductive core, provided that the formally represented facts preserve the required semantics.

\begin{principle}[Epistemic Grounding Principle]
Let $Q$ denote a jurisprudential query expressed in natural language. Before any deductive inference may be performed, the information contained in $Q$ shall first be transformed into a formally grounded set of legal facts $G(Q) = \{f_1, f_2, \ldots, f_n\}$, where every fact is obtained through one or more formally defined empirical grounding processes appropriate to the domain under consideration.
\end{principle}

Accordingly, an adequate formal system for computational Islamic jurisprudence shall satisfy the following requirements:
\begin{enumerate}
    \item Empirical Origin: Every fact supplied to the deductive engine shall originate from an identifiable empirical source whose contribution can be formally specified.
    \item Linguistic Grounding: Whenever legal interpretation depends upon linguistic analysis, formally recoverable semantic information contained in Classical Arabic shall participate directly in the grounding process.
    \item Multi-Domain Integration: The grounding architecture shall permit arbitrary empirical disciplines to contribute formally represented facts without modifying the deductive methodology.
    \item Separation of Computational Responsibilities: Epistemic grounding and jurisprudential deduction shall constitute distinct computational processes whose correctness may be analyzed independently.
    \item Extensibility: Additional empirical sciences may be incorporated into the grounding architecture without altering the semantic foundations or logical correctness of the deductive system.
\end{enumerate}

\subsection{Foundational Principle V: Computational Expressiveness of the Formal System}

\subsubsection{Motivation}
The preceding foundational principles established the epistemological nature of Islamic jurisprudence, its methodological structure, the logical requirements of its formalization, and the epistemic grounding process through which formally represented legal facts are obtained. A further foundational question nevertheless remains. Having determined the semantic and logical properties required of the formal system, what degree of computational expressiveness should its execution mechanism possess?

The answer to this question is not obvious. Modern theoretical computer science offers a wide spectrum of computational formalisms ranging from unrestricted Turing-complete systems to highly restricted executable rule systems possessing strong mathematical guarantees concerning termination, decidability, and computational complexity. Choosing among these alternatives constitutes a fundamental architectural decision whose consequences extend throughout the entire computational framework. Accordingly, the expressive power of the proposed formal system should not be selected arbitrarily, but derived from the scientific requirements of Islamic jurisprudence itself.

\subsubsection{The Engineering Principle of Minimal Sufficient Expressiveness}
A common misconception in computational system design is that greater computational expressiveness necessarily produces superior systems. In reality, increased expressive power is accompanied by significant computational costs. Unrestricted computational models generally sacrifice guaranteed termination, decidability, complexity bounds, formal verification, and predictable execution. These properties become increasingly important for systems intended to support scientific reasoning. Consequently, computational expressiveness should be regarded not as an objective to maximize but as an engineering resource to be employed only when scientifically necessary.

This observation leads to the following engineering principle: A formal system should possess exactly the computational expressive power required by its scientific domain, and no more. Additional expressive power is scientifically justified only when it enables forms of reasoning that cannot otherwise be represented.

\subsubsection{Evaluation Criteria}
The computational expressiveness of a formal system should therefore be evaluated according to several complementary criteria. An appropriate computational foundation should provide: sufficient expressive power to represent jurisprudential reasoning faithfully; guaranteed termination whenever legal conclusions are computed; decidable inference procedures; computational tractability for large jurisprudential knowledge bases; deterministic and reproducible execution; transparent explanation of derived legal conclusions; compatibility with formal verification; scalability under continued expansion of legal knowledge.

\subsubsection{Computational Requirements of Islamic Jurisprudence}
The methodology of Islamic jurisprudence imposes several important observations concerning computational expressiveness. First, deductive reasoning within jurisprudence primarily consists of applying previously established legal rules to formally represented facts. Second, these rules are conditional in nature and therefore naturally correspond to executable rule systems. Third, jurisprudential computation requires transparent chains of justification rather than opaque computational procedures. Fourth, legal consultation demands predictable and terminating computation. Finally, the inductive methodologies responsible for discovering legal rules constitute a separate scientific activity from the deductive process responsible for applying those rules. Consequently, the computational engine responsible for legal consultation is not required to perform arbitrary computation. Its purpose is to execute an already established body of jurisprudential knowledge correctly, efficiently, and explainably.

\subsubsection{Separation Between Semantic and Computational Expressiveness}
The preceding observation suggests an important architectural distinction. The semantic richness of a formal system should not be confused with the computational expressiveness of its execution mechanism. A formal system may possess an extremely rich semantic structure while employing a comparatively restricted computational engine. For computational Islamic jurisprudence, this distinction is particularly significant. The semantic layer must faithfully represent legal causes, conditions, impediments, contextual qualifications, normative modalities, linguistic information, and methodological concepts developed within Uṣūl al-Fiqh. The computational layer, however, need only execute formally represented rules over grounded legal facts. Accordingly, semantic richness should reside primarily within the representation of legal knowledge, whereas computational simplicity should characterize the execution mechanism itself. This separation permits faithful jurisprudential representation while preserving desirable computational guarantees.

\subsubsection{Scientific Consequences}
The preceding analysis leads to an important conclusion. There exists no scientific justification for adopting unrestricted computational expressiveness merely because it is available. Whenever the reasoning tasks required by Islamic jurisprudence can be faithfully represented within a more restricted executable formalism, that restricted formalism should be preferred. Such a choice improves predictability, facilitates formal verification, guarantees termination under the adopted formalism, enhances explainability, and supports efficient implementation without sacrificing jurisprudential correctness.

Consequently, computational restraint should be regarded as an engineering virtue rather than a computational limitation.

\begin{principle}[Principle of Minimal Sufficient Computational Expressiveness]
Let $\mathcal{F}$ denote a formal system satisfying the preceding foundational principles. The computational realization of $\mathcal{F}$ shall employ the least expressive computational formalism capable of faithfully executing the semantic and logical requirements imposed by Islamic jurisprudence.
\end{principle}

Accordingly:
\begin{enumerate}
    \item computational expressiveness shall never exceed what is scientifically required for jurisprudential reasoning;
    \item semantic richness shall remain independent of computational complexity;
    \item the execution mechanism shall preserve deterministic, terminating, decidable, and explainable inference whenever these properties are compatible with the expressive requirements of the formal system;
    \item unrestricted computational mechanisms shall be adopted only when their additional expressive power is demonstrated to be scientifically indispensable.
\end{enumerate}

\subsection{Foundational Principle VI: Layered Architecture of Computational Islamic Jurisprudence}

\subsubsection{Motivation}
The preceding foundational principles established the essential scientific requirements governing computational Islamic jurisprudence. Collectively, these principles determine not merely the individual components of the formal system but also the manner in which those components must interact. The present principle therefore addresses the following engineering question: What system architecture simultaneously satisfies the preceding five foundational principles?

\subsubsection{Assessment of Existing Computational Architectures}
The preceding investigation demonstrated that executable Horn-style rule systems provide the closest existing computational realization of the requirements established by the previous principles. Compared with unrestricted computational models, executable Horn systems naturally provide deterministic execution, transparent chains of inference, efficient rule evaluation, decidable inference under appropriate restrictions, guaranteed termination within the adopted computational fragment, and practical scalability for large rule bases. These properties correspond closely to the computational objectives established by Foundational Principle V.

Nevertheless, Horn logic alone cannot constitute a complete computational foundation for Islamic jurisprudence. Although its execution mechanism is highly suitable, it lacks explicit support for several essential components established by the previous principles. Among these are epistemic grounding, Arabic linguistic computation, jurisprudential semantic structures, normative modalities, integration of empirical sciences, provenance of derived facts, explanation-oriented reasoning, and interaction between inductive and deductive methodologies. Consequently, existing executable rule systems should be regarded as computational substrates rather than complete jurisprudential architectures.

\subsubsection{Separation of Computational Responsibilities}
The preceding observations suggest an important architectural principle. Different scientific tasks require different computational mechanisms. Attempting to perform all jurisprudential activities within a single monolithic computational component unnecessarily couples independent scientific problems and complicates both formal analysis and implementation. Instead, computational Islamic jurisprudence should explicitly separate the principal scientific responsibilities into independent but interacting architectural layers. Each layer addresses a distinct scientific problem while communicating with adjacent layers through formally specified interfaces. This separation improves modularity, verification, extensibility, explainability, and long-term maintainability without altering the underlying jurisprudential methodology.

\subsubsection{The Layered Architecture}
Accordingly, the proposed formal system is organized into the following conceptual layers:
\begin{itemize}
    \item \textbf{Layer 1 — Epistemic Grounding Layer:} This layer transforms natural-language descriptions and empirical observations into formally represented legal facts. It integrates Classical Arabic linguistic analysis together with every empirical science relevant to the legal question under consideration. Its output consists exclusively of grounded formal facts.
    \item \textbf{Layer 2 — Jurisprudential Semantic Layer:} This layer represents the conceptual structures of Islamic jurisprudence. Legal causes (‘ilal), conditions (shurūṭ), impediments (mawāni‘), exceptions, normative modalities, contextual qualifications, provenance information, and methodological concepts are represented at this level. This layer determines the semantics of legal reasoning independently of its computational execution.
    \item \textbf{Layer 3 — Executable Deductive Layer:} This layer performs the deterministic execution of jurisprudential rules over the grounded facts supplied by the previous layers. Its computational realization employs a restricted executable rule-based formalism satisfying the requirements established in Foundational Principle V. This layer is responsible solely for legal deduction and does not perform empirical grounding or semantic interpretation.
    \item \textbf{Layer 4 — Explanation and Justification Layer:} Every legal conclusion produced by the deductive engine is accompanied by a formally reconstructible chain of reasoning identifying the grounded facts, legal rules, intermediate inferences, and methodological principles responsible for the resulting legal judgment. This layer provides transparency, auditability, and scientific reproducibility.
    \item \textbf{Layer 5 — Knowledge Acquisition Layer:} The acquisition, verification, refinement, and extension of jurisprudential knowledge constitute a distinct scientific activity from legal consultation itself. This layer supports the inductive methodologies of Uṣūl al-Fiqh, including the discovery, validation, organization, and refinement of jurisprudential rules. Its output extends the knowledge base subsequently employed by the deductive layer.
\end{itemize}

\subsubsection{Interaction Between the Layers}
The layered organization reflects the natural scientific workflow of jurisprudential reasoning. Empirical reality is first analyzed by the epistemic grounding layer. The resulting grounded facts are interpreted within the jurisprudential semantic layer. These formally represented concepts are then processed by the executable deductive layer to derive legal conclusions. The resulting conclusions are accompanied by explicit explanations generated by the justification layer. Meanwhile, the knowledge acquisition layer continuously refines and extends the jurisprudential knowledge base through the inductive methodologies developed within Uṣūl al-Fiqh. Thus, induction, grounding, deduction, and explanation become coordinated but logically independent computational processes.

\begin{principle}[Architectural Principle]
Let $\mathcal{F}$ denote a formal system satisfying Foundational Principles I–V. Then an adequate realization of $\mathcal{F}$ shall organize its computational processes into distinct architectural layers, each responsible for a single well-defined scientific function.
\end{principle}

Accordingly:
\begin{enumerate}
    \item epistemic grounding shall remain independent of logical deduction;
    \item semantic representation shall remain independent of computational execution;
    \item deductive reasoning shall operate exclusively upon grounded formal facts;
    \item explanation shall constitute an explicit computational product rather than an informal by-product;
    \item inductive knowledge acquisition shall remain computationally distinct from deductive legal consultation.
\end{enumerate}

The correctness of the complete formal system therefore results not from a single computational mechanism but from the coordinated interaction of multiple formally specified computational layers.

\subsection{Consequences of the Foundational Principles}

\subsubsection{Theoretical Consequences}
The preceding six foundational principles collectively establish the theoretical foundations upon which computational Islamic jurisprudence may be developed as a coherent scientific discipline. Rather than presenting isolated observations concerning logic, linguistics, computation, or software architecture, the present chapter has derived a unified theoretical framework whose components follow naturally from one another.

The resulting framework establishes the following conclusions:
\begin{enumerate}
    \item First, computational Islamic jurisprudence is founded upon an empirical normative science whose practical application depends simultaneously upon revealed legal sources and empirically established descriptions of reality.
    \item Second, jurisprudential reasoning cannot be reduced to deductive logic alone but necessarily combines inductive methodologies responsible for the discovery and organization of legal knowledge with deductive methodologies responsible for its application to individual legal consultations.
    \item Third, no existing historical logical formalism completely satisfies the semantic, methodological, linguistic, and computational requirements imposed by Islamic jurisprudence. Existing logical systems should therefore be regarded as valuable components from which a new formal system may be systematically engineered rather than adopted without modification.
    \item Fourth, legal deduction cannot operate directly upon empirical reality. Before deduction begins, natural-language descriptions and empirical observations must first undergo a formally defined epistemic grounding process capable of producing grounded legal facts suitable for jurisprudential reasoning.
    \item Fifth, computational expressiveness should not be regarded as an objective to maximize. Instead, the computational realization of jurisprudential reasoning should employ the least expressive executable formalism capable of faithfully representing the semantic and methodological requirements of Islamic jurisprudence while preserving desirable computational guarantees.
    \item Finally, the interaction between epistemic grounding, semantic representation, deductive reasoning, explanation, and knowledge acquisition naturally leads to a layered computational architecture whose organization follows directly from the preceding principles rather than from arbitrary software engineering preferences.
\end{enumerate}

\subsubsection{Scientific Evaluation Criteria}
The six foundational principles also establish objective criteria according to which any proposed formal system for computational Islamic jurisprudence may be scientifically evaluated. Rather than assessing implementations solely according to computational performance or implementation convenience, an adequate formal system should satisfy all of the theoretical requirements established throughout the present chapter.

\begin{table}[htbp]
\centering
\caption{Scientific Evaluation Criteria}
\begin{tabular}{ll}
\toprule
Evaluation Criterion & Established by \\
\midrule
Epistemological adequacy & Foundational Principle I \\
Methodological adequacy & Foundational Principle II \\
Logical adequacy & Foundational Principle III \\
Correctness of epistemic grounding & Foundational Principle IV \\
Computational adequacy & Foundational Principle V \\
Architectural adequacy & Foundational Principle VI \\
\bottomrule
\end{tabular}
\end{table}

\subsubsection{Consequences for the Present Work}
The remaining chapters of this work should therefore be understood as the construction of one concrete realization satisfying the six foundational principles established above. Accordingly, the proposed architecture is not presented as the unique possible realization of computational Islamic jurisprudence. Rather, it represents one implementation deliberately engineered to satisfy the epistemological, methodological, logical, epistemic, computational, and architectural requirements established throughout this chapter. Future research may propose alternative realizations while preserving the same foundational principles. Likewise, future advances in logic, computational linguistics, knowledge representation, or automated reasoning may motivate improved implementations without requiring revision of the underlying theoretical framework established herein.

\subsubsection{Foundational Principle of Computational Islamic Jurisprudence}
The preceding discussion may be summarized by the following foundational theorem.

\begin{unnumberedtheorem}
\textbf{The Principle of Foundational Adequacy}\par
Let $\mathcal{F}$ denote a formal system intended to support computational Islamic jurisprudence. Then $\mathcal{F}$ constitutes a theoretically adequate formal system if and only if it simultaneously satisfies:
\begin{enumerate}
    \item the epistemological requirements established in Foundational Principle I;
    \item the methodological requirements established in Foundational Principle II;
    \item the logical requirements established in Foundational Principle III;
    \item the epistemic grounding requirements established in Foundational Principle IV;
    \item the computational requirements established in Foundational Principle V; and
    \item the architectural requirements established in Foundational Principle VI.
\end{enumerate}
No subset of these principles is sufficient to establish a complete computational theory of Islamic jurisprudence. Conversely, any formal system satisfying all six principles provides a coherent theoretical foundation upon which computational jurisprudential reasoning may be systematically developed.
\end{unnumberedtheorem}

\subsubsection{Closing Remarks}
The principal contribution of the present chapter is therefore not the specification of a particular software architecture, programming language, or logical formalism. Rather, it establishes a coherent theoretical framework from which such architectures may be systematically derived, scientifically analyzed, and objectively evaluated. The layered architecture presented in the subsequent chapters should consequently be understood as one realization of these foundations rather than the foundations themselves. This distinction is fundamental. Theoretical principles should remain stable under technological evolution, whereas computational realizations may continue to evolve as improved logical formalisms, linguistic technologies, knowledge representation methods, and implementation techniques become available. Accordingly, the remainder of this work is devoted not to establishing new theoretical assumptions, but to demonstrating how the proposed computational architecture realizes the theoretical framework developed throughout the present chapter.

\newpage
\part*{Part II: A Formal System of Fiqh}
\setcounter{theorem}{0}
\addcontentsline{toc}{part}{Part II: A Formal System of Fiqh}
\setcounter{section}{0}

\section{Introduction}
\label{sec:intro}

The preceding chapters established the epistemological, methodological, logical, computational, and architectural principles underlying the formalization of Islamic jurisprudence. The objective of the present chapter is to transform these principles into a mathematically precise computational language suitable for automated jurisprudential reasoning.

The principal design objective is to reconcile two requirements that are often regarded as incompatible. On the one hand, the formal language must faithfully represent the linguistic and semantic richness of Classical Arabic together with the methodological requirements of U\d{s}\=ul al-Fiqh. On the other hand, it must remain computationally tractable and executable using well-understood logical inference techniques.

Rather than extending an existing logical formalism directly, we propose a layered computational language in which linguistic processing, semantic normalization, and logical deduction are explicitly separated. This separation allows each component to be developed according to its own theoretical principles while preserving a clean and mathematically rigorous interface between successive stages.

The resulting formal language consists of three interacting components:
\begin{enumerate}
    \item An Arabic Morphological/Grammatical Grounding Engine, responsible for transforming Arabic linguistic expressions into canonical semantic representations.
    \item An Arabic Reified Action Calculus (ARAC), responsible for constructing explicit logical representations of juridically relevant actions and their semantic participants.
    \item A Datalog deductive engine, responsible for deriving legal conclusions by means of executable Horn-clause inference over the normalized semantic representation produced by the preceding layers.
\end{enumerate}

Unlike conventional natural-language processing systems, the proposed architecture does not perform logical inference directly over Arabic sentences. Instead, Arabic serves as the source language from which semantic structures are extracted. Logical deduction operates only upon the normalized semantic representation generated by the grounding layer.

This design follows directly from the foundational principles established in Chapter~2. Principle~III requires a logical formalism capable of representing both static legal facts and legally relevant actions. Principle~IV requires that such representations be obtained through an explicit epistemic grounding process. Principle~V requires that the deductive core remains computationally tractable. Finally, Principle~VI requires that the overall architecture preserves a strict separation between linguistic analysis, semantic representation, and logical inference.

Several assumptions underlying the formalization should be stated explicitly before the formal development.

\begin{enumerate}
    \item
    This work does not claim to capture the entirety of Islamic jurisprudence. It formalizes a well-defined projection of Fiqh and proves mathematical properties of that formalization. Completeness is therefore always understood relative to the formally specified domain and representation. Requiring a formal system to be complete relative to an unformalized phenomenon such as the entirety of Islamic jurisprudence would confuse the formal model with the phenomenon it represents. The completeness theorems are consequently substantive guarantees about the deductive power of the formal system over its specified domain; they are not tautologies and do not assert completeness of the underlying jurisprudential tradition as a whole.

    \item
    Every Fiqh school operates with a systematic ontology of concepts, categories, and their interrelations. Classical and modern Fiqh texts are organized into chapters that introduce core concepts---such as sale, marriage, ownership, purity, and prayer---before discussing specific rulings and legal cases. This ontological infrastructure is therefore internal to Fiqh rather than an external addition imposed by the formalization. The Ontological Meta-Map used in the subsequent formalization is the formal projection of this existing conceptual apparatus. Differences between schools may therefore be represented by parameterizing the formal system with the appropriate Meta-Map for the jurisprudential school under consideration.

    \item
    The formal language is restricted to fully vocalized Classical Arabic sentences. This is a deliberate design choice analogous to the syntactic well-formedness requirements of programming languages. Full vocalization substantially reduces morphological ambiguity and provides the morphological information required by the subsequent syntactic analysis. The restriction therefore establishes a well-defined input language in which deterministic parsing can be formally studied. It also provides a natural interface for classical Fiqh texts and for domain experts familiar with fully vocalized Arabic.
\end{enumerate}

The requirement of full vocalization should not, however, be interpreted as asserting that every fully vocalized surface form has exactly one morphological analysis. Rare lexical forms may remain morphologically ambiguous even after vocalization. For example, the fully vocalized form \ar{أَحْمَدُ} may be analyzed either as a proper noun in the nominative case, as in \ar{جَاءَ أَحْمَدُ}, or as a first-person singular present-tense verb in the nominative case, as in \ar{أَحْمَدُ اللَّهَ عَلَى كُلِّ شَيْءٍ}. The formal architecture accommodates such cases explicitly.

At the morphological stage, the Ta\c{s}r\={i}f procedure allows the function $\mu(w)$ to return all morphologically admissible analyses of a word. Thus, for \ar{أَحْمَدُ}, the morphological stage may preserve both analyses together with their respective morphological features and $\delta$ descriptors. Morphological ambiguity is therefore not silently discarded at the grounding stage.

The subsequent I\text{'}r\={a}b procedure resolves such ambiguity through syntactic context. Guided by the Fiqh-Categorized Grammar $G_{\mathrm{Fiqh}}^\delta$, the parser tests the admissible morphological candidates against the syntactic roles permitted by the relevant production rules. In \ar{جَاءَ أَحْمَدُ}, the proper-noun analysis is compatible with the required subject role, whereas in \ar{أَحْمَدُ اللَّهَ عَلَى كُلِّ شَيْءٍ}, the verbal analysis is compatible with the syntactic structure governing the object. Consequently, the grounding architecture preserves morphological alternatives and permits the syntactic layer to eliminate candidates that are incompatible with context.

This distinction is important for the formal guarantees established later. The unambiguity theorem concerns the grammar and its admissible syntactic derivations; it does not require the morphological recognition procedure to suppress every possible lexical analysis in advance. Instead, the grounding layer may expose a finite set of candidates, after which the syntactic constraints determine the admissible structure.

The purpose of the formal development is therefore threefold. First, it formally defines the proposed computational language. Second, it establishes its mathematical properties, including correctness, relative completeness, termination, and decidability. Third, it demonstrates the operation of the language through jurisprudential examples showing the interaction between Arabic morphology, semantic normalization, and executable legal reasoning.

The formal system is developed through three successive strata. The first establishes the abstract algebraic architecture; the second provides the general linguistic and semantic grounding machinery; and the third instantiates these abstractions as the concrete Fiqh formal system. A final bridging methodology connects the deductive representation of legal rules with their inductive extraction from closed Fiqh chapters.

At the abstract architectural level (Section~2), the computational language is defined as the functional composition
\[
F = D \circ A \circ G,
\]
where $G$ denotes the grounding engine, $A$ the Arabic Reified Action Calculus, and $D$ the Datalog deductive system. The components are initially treated as abstract functions between formally specified domains. This permits the fundamental algebraic properties of the architecture to be established independently of any particular implementation of the individual components. Theorem~1 establishes the functional separation of the three layers, while Theorems~12 and~13 establish abstract correctness and completeness. Theorems~14 and~15 establish termination and decidability, respectively. Theorem~16 characterizes the possible exponential worst-case behavior arising from ambiguity at this abstract level, and Theorem~17 consolidates the principal abstract guarantees.

The conditional nature of these early theorems is deliberate. The reader should not interpret the hypotheses appearing in the correctness, completeness, termination, and decidability results of the first two strata as unresolved weaknesses in the framework. At these levels, the internal behavior of the grounding and semantic components is intentionally abstracted away. The corresponding theorems therefore establish conditional guarantees: whenever the abstract components satisfy the stated conditions, the composed system inherits the corresponding properties.

The later formalization supplies precisely the missing instantiations. In particular, Section~10 provides detailed definitions and constructions for the linguistic, grammatical, semantic, and jurisprudential components used by the practical system. The subsequent theorems establish that these concrete components satisfy the conditions required by the earlier abstract results. Thus, the correctness, completeness, and decidability results proved for the concrete Fiqh system in Section~10 and the following sections do not stand independently of the earlier strata; rather, they discharge the hypotheses of the corresponding abstract theorems. The conditional guarantees of the first strata consequently become unconditional guarantees for the concrete formal system with respect to the assumptions explicitly stated for that system.

The general grammatical and semantic grounding layer (Sections~3--4) then operationalizes the abstract functions $G$ and $A$. The Ta\c{s}r\={i}f procedure for morphological recognition is formalized together with its correctness and termination properties. The general syntactic recognition function $\iota$ computes grammatically admissible I\text{'}r\={a}b structures, while the semantic grounding function $\sigma$ maps these structures to language-independent Grounded Action Structures (GAS) through lexical semantic signatures. The corresponding theorems establish the correctness, completeness, and termination of this general grounding architecture under its stated conditions. At the ARAC level, the canonical representation of actions is established through lexical roots, and the translation of ARAC structures into Datalog facts is shown to preserve the relevant semantics.

The concrete practical formal system of Fiqh (Sections~10--12) instantiates the preceding abstractions with the Fiqh-Categorized Attribute Grammar $G_{\mathrm{Fiqh}}^\delta$. Its non-terminals represent jurisprudentially meaningful syntactic roles, while its productions are constrained by the morphosyntactic $\delta$ descriptors recovered during morphological recognition. The resulting grammar is proved unambiguous under the established construction conditions. This permits the Specialized I\text{'}r\={a}b Procedure to operate deterministically in $O(n)$ time, thereby satisfying the tractability requirement of Principle~V.

The concrete system also discharges the hypotheses required by the preceding abstract results. Its correctness and completeness results establish that the concrete grounding and semantic components satisfy the corresponding conditions of the earlier composition theorems, while its termination and decidability results establish the corresponding computational properties of the instantiated system. Lemma~2 establishes the representational completeness of the ARAC transformation over finite Grounded Action Structures, and the subsequent theorems establish the correctness and decidability of the complete Fiqh pipeline.

Finally, Section~12 introduces the bridging methodology through Algorithm~1 ($\Omega$). Given a Horn-clause program $P$ representing the legal rulings of a closed Fiqh chapter $C$, together with an Ontological Meta-Map $\mathcal{M}_{\mathrm{onto}}$, the algorithm constructs a specialized grammar $G_C$. The Meta-Map is defined as a total bijection between the logical predicates of $P$ and the corresponding $\delta$-typed non-terminals of the specialized grammar. This provides a formal bridge between the inductive extraction of legal rules, including the methodology of al-\d{S}abr wa al-Taq\c{s}\={i}m, and their deductive application. Theorem~27 establishes the correspondence between grammatical acceptance and instantiation of the premises of the relevant Horn clauses, while Theorem~28 extends the decidability and polynomial-time guarantees to the generated grammar.

The resulting stratification is therefore not merely an organizational convenience. It separates three logically different questions: whether an abstract composition has the desired properties under specified component assumptions; whether the linguistic and semantic components actually satisfy those assumptions; and whether the resulting Fiqh-specific system consequently inherits the desired guarantees. This structure makes explicit where each mathematical property is established and how the concrete construction discharges the hypotheses of the preceding abstract results.

The formal development thus proceeds from generic algebraic guarantees, through general linguistic and semantic constructions, to a concrete Fiqh-specific instantiation and its rule-generation methodology. Under the assumptions explicitly stated throughout the formalization, this development yields a formally verifiable, computationally tractable, epistemologically grounded, and decidable framework for automated Islamic jurisprudence.

\section{The Computational Language} \label{sec:comp-lang}
The computational framework proposed in this work is not a direct extension of an existing logical formalism, but a new layered formal language specifically designed for executable Islamic jurisprudence. The principal objective is to separate linguistic analysis, semantic normalization, and logical deduction into independent mathematical components while preserving a formally defined interface between successive stages.

Unlike conventional logic programming systems, which operate directly upon symbolic predicates supplied by the programmer, the present framework accepts Arabic linguistic expressions as its input. These expressions are first transformed into canonical semantic representations before any deductive reasoning is performed.

\begin{definition}[Computational Language] \label{def:comp-lang}
The formal computational language developed in this work is the triple
\[
F = \langle G, A, D \rangle
\]
where
\begin{itemize}
\item $G$ denotes the Arabic Morphological/Grammatical Grounding Engine;
\item $A$ denotes the Arabic Reified Action Calculus (ARAC);
\item $D$ denotes the Datalog Deductive System.
\end{itemize}
Each component performs a distinct mathematical transformation whose output constitutes the input of the subsequent component. Consequently, the computational language is not identified with any individual formalism, but with the composition of these three formally specified transformations.
\end{definition}

\begin{definition}[Component Functions] \label{def:comp-func}
Each component of the computational language is modeled as a total function between formally defined domains.

The grounding component is the function $G: L_A \rightarrow S$, where
\begin{itemize}
\item $L_A$ denotes the set of well-formed Arabic linguistic expressions;
\item $S$ denotes the space of grounded semantic structures.
\end{itemize}
The function $G$ performs linguistic analysis, morphological/grammatical decomposition, semantic role identification, lexical normalization, and contextual grounding. No logical inference is performed at this stage.

The second component is the function $A: S \rightarrow K$, where $K$ denotes the normalized jurisprudential knowledge base. The purpose of ARAC is to transform grounded semantic structures into a declarative logical representation consisting of static legal facts together with explicit representations of juridically relevant actions.

Finally, $D: K \rightarrow C$, where $C$ denotes the set of deductively derivable legal conclusions. The deductive component consists of a standard function-free Horn-clause Datalog program evaluated according to its least fixed-point semantics.
\end{definition}

\begin{definition}[Composition] \label{def:comp}
The complete computational language is defined by the functional composition
\[
F = D \circ A \circ G
\]
Thus, for every Arabic linguistic input $l \in L_A$, the resulting legal conclusions are given by $F(l) = (D \circ A \circ G)(l)$. This composition defines the formal semantics of the computational language.
\end{definition}

\begin{remark}[Layered Separation] \label{rem:layer-sep}
The three transformations perform fundamentally different computational tasks.
\begin{itemize}
\item The grounding layer analyzes Arabic linguistic structure and constructs semantic representations.
\item ARAC converts these semantic representations into normalized logical structures suitable for automated reasoning.
\item The Datalog layer derives legal conclusions exclusively from these normalized logical structures.
\end{itemize}
No component performs the responsibilities assigned to another component. Consequently, linguistic analysis, semantic representation, and deductive reasoning remain mathematically independent.
\end{remark}

\begin{theorem}[Layer Separation] \label{thm:layer-sep}
Within the computational language $F = \langle G, A, D \rangle$, the three computational layers are functionally independent. Specifically:
\begin{enumerate}
\item the grounding layer performs no deductive inference;
\item ARAC performs no logical deduction and introduces no new legal conclusions;
\item the Datalog layer performs no linguistic or morphological analysis.
\end{enumerate}
\textit{Proof.} By Definition~\ref{def:comp-func}, each component is defined as a function whose codomain coincides with the domain of the subsequent transformation. The grounding function $G: L_A \rightarrow S$ produces grounded semantic structures exclusively from Arabic linguistic expressions. Since its codomain contains semantic representations rather than logical consequences, no deductive reasoning is performed. Similarly, $A: S \rightarrow K$ constructs normalized logical representations from grounded semantic structures. Its purpose is representational rather than inferential. Therefore ARAC neither derives new legal conclusions nor evaluates logical consequences. Finally, $D: K \rightarrow C$ accepts only normalized logical representations as input. Since Arabic linguistic expressions do not belong to the domain of $D$, no linguistic or morphological processing occurs during deduction. Therefore each computational responsibility is confined exclusively to its designated transformation, establishing the functional independence of the three computational layers. $\blacksquare$
\end{theorem}

\paragraph{Discussion}
The preceding definitions establish the computational language as the composition of three mathematically distinct transformations rather than as a monolithic logical system. This separation has important theoretical consequences. First, it localizes linguistic complexity within the grounding layer, allowing the deductive engine to remain a standard decidable Horn-clause evaluator. Second, it permits the semantic representation of Arabic linguistic structures to evolve independently of the deductive mechanism, thereby facilitating future extensions without affecting logical correctness. Finally, the modular structure allows the correctness, completeness, and decidability of the overall system to be established compositionally by proving the corresponding properties for each transformation separately.

\section{The Arabic Morphological/Grammatical Grounding Engine} \label{sec:grounding}
\subsection{Formal Language of Classical Arabic and the Arabic Recognition Engine}
The proposed computational framework operates over a formally defined fragment of Classical Arabic rather than unrestricted natural language. As with every formal language, admissible inputs are required to satisfy precise lexical, morphological, and grammatical constraints before semantic interpretation and deductive reasoning may begin.

The objective of these constraints is not to reduce the expressive richness of Arabic, but to establish a mathematically precise linguistic foundation upon which sound, deterministic, and decidable legal reasoning can be constructed.

Unlike ordinary written Arabic, where omitted vocalization (\ar{التشكيل}) may introduce lexical and grammatical ambiguity, the language accepted by the present formal system consists exclusively of fully vocalized Classical Arabic sentences satisfying the grammatical rules of the classical Arabic linguistic tradition [16, 18].

This design principle is directly analogous to the lexical and syntactic requirements imposed by programming languages [4]. A compiler assumes that a source program satisfies the grammar of the language before semantic analysis is attempted. Likewise, the proposed formal system assumes that every input sentence satisfies the formal grammatical conventions of Classical Arabic before computational grounding begins.

Accordingly, the system does not attempt to reconstruct omitted vowel markings or infer missing grammatical information. Complete vocalization constitutes an explicit requirement of the formal language itself.

\begin{definition}[Formal Language of Admissible Classical Arabic] \label{def:adm-ca}
Let $\mathcal{L}_A$ denote the formal language of admissible Classical Arabic sentences. A sentence $l \in \mathcal{L}_A$ is \textbf{admissible} if and only if it satisfies the following conditions:
\begin{enumerate}
\item \textbf{Complete Vocalization.} Every lexical item is fully vocalized (\ar{مشكول}), including all vowel markings, case endings, and orthographic information required for grammatical interpretation.
\item \textbf{Morphological Well-Formedness.} Every lexical item admits at least one valid morphological analysis according to the rules of Classical Arabic morphology (\ar{التصرف}).
\item \textbf{Grammatical Well-Formedness.} The sentence admits at least one valid grammatical analysis (\ar{إعراب}) according to the rules of Classical Arabic grammar (\ar{النحو}).
\item \textbf{Semantic Admissibility.} Every governing predicate occurring in the sentence possesses a semantic interpretation within the lexical knowledge base of the formal system.
\end{enumerate}
Sentences violating any of these conditions do not belong to the language $\mathcal{L}_A$ and are rejected before the grounding process begins.
\end{definition}

\begin{remark}[Classical Arabic as a Formal Language] \label{rem:ca-formal}
The requirement of complete vocalization is a design assumption of the computational framework rather than a linguistic claim concerning ordinary Arabic writing. The objective of the present work is to define a formal language suitable for automated legal reasoning. As in every formal language, lexical and grammatical well-formedness are prerequisites for deterministic computation. Requiring fully vocalized input therefore plays the same role as requiring syntactically correct programs in conventional programming languages. Consequently, ambiguity arising solely from omitted vowel markings lies outside the scope of the present formalization.
\end{remark}

\begin{definition}[Arabic Recognition Engine] \label{def:are}
The linguistic component of the proposed framework is called the \textbf{Arabic Recognition Engine (ARE)}. The Arabic Recognition Engine is a layered recognition system founded directly upon the classical Arabic linguistic sciences. It consists of three successive computational stages:
\begin{enumerate}
\item \textbf{Morphological Recognition (\ar{التصرف}).} This stage identifies the internal morphological structure of every lexical item, including its root, morphological pattern, grammatical category, tense, voice, gender, number, and other lexical attributes.
\item \textbf{Syntactic Recognition (\ar{الإعراب}).} This stage determines the grammatical function of every sentence constituent according to the rules of Classical Arabic grammar, identifying relations such as \ar{فاعل، مفعول به، مبتدأ، خبر، حال، تمييز، مضاف إليه، جار ومجرور}.
\item \textbf{Semantic Grounding.} This stage interprets the recognized grammatical structure using the lexical semantics associated with the governing Arabic roots, producing canonical semantic representations suitable for formal reasoning.
\end{enumerate}
Collectively, these three stages constitute the grounding engine of the proposed formal system.
\end{definition}

\begin{definition}[Grounding Function] \label{def:ground-func}
Let $\mathcal{A}_G$ denote the set of Grounded Action Structures. The grounding function is the transformation
\[
\mathcal{G} : \mathcal{L}_A \longrightarrow \mathcal{P}_f(\mathcal{A}_G),
\]
where $\mathcal{P}_f(\mathcal{A}_G)$ denotes the collection of all finite subsets of Grounded Action Structures. For every admissible sentence $l \in \mathcal{L}_A$, the grounding function returns the complete finite set $\mathcal{G}(l) = \{G_1, \ldots, G_n\}$, containing every Grounded Action Structure corresponding to a grammatically admissible interpretation of the sentence. When the sentence admits a unique grammatical analysis, $n = 1$, and the grounding process is deterministic. When multiple grammatically admissible analyses exist, each analysis is preserved independently throughout the grounding process. The framework therefore represents linguistic ambiguity symbolically rather than resolving it prematurely or assigning probabilistic weights.
\end{definition}

\paragraph{Linguistic Grounding Principle}
The proposed framework deliberately separates \textbf{linguistic recognition} from \textbf{deductive reasoning}. The Arabic Recognition Engine is responsible exclusively for recognizing the linguistic structure of admissible Classical Arabic through the successive application of \ar{التصرف}, \ar{الإعراب}, and semantic grounding. Only after linguistic recognition has terminated are the resulting Grounded Action Structures translated into the Arabic Reified Action Calculus (ARAC) and subsequently encoded as Datalog facts for deductive reasoning. This layered architecture establishes a clean interface between language understanding and formal legal inference. It preserves the modularity of the framework, allows each component to be verified independently, and provides the mathematical foundation upon which the correctness, completeness, and decidability results established in the subsequent sections are proved.

\subsection{Formal Specification of the Ta\c{s}r\={\i}f Procedure (Morphological Recognition)} \label{sec:tasrif}
To operationalize the abstract morphological function
\[
\mu:L_A\rightarrow\mathcal{M},
\]
we formally define the computational mechanics of the Ta\c{s}r\={\i}f procedure.
Unlike concatenative languages, Classical Arabic employs a non-concatenative morphological system in which lexical roots and morphological patterns are interleaved [7]. Consequently, morphological recognition cannot be performed by simple string decomposition, but requires a specialized recognition procedure capable of recovering the canonical lexical root together with its associated morphological pattern and morphosyntactic inflection [20].

The Ta\c{s}r\={\i}f procedure is defined as a deterministic sequence of finite computational transformations.

\subsubsection*{1. Morphological Knowledge Base}
Let $\Sigma$ denote the alphabet of fully vocalized Classical Arabic.
The Ta\c{s}r\={\i}f procedure operates with respect to the finite morphological knowledge base
\[
K_{\mathrm{morph}}
=
\langle
R,
P,
A,
\tau,
H_{\mathrm{repo}}
\rangle,
\]
where
\begin{itemize}
\item $R$ is the finite set of admissible triliteral and quadriliteral Arabic roots.
\item $P$ is the finite set of admissible Arabic morphological patterns.
\item $A=A_{pre}\cup A_{suf}$ is the finite set of prefixes, suffixes, proclitics, enclitics and grammatical particles.
\item $\tau:\Sigma^*\rightarrow\Sigma^*$ is the orthographic normalization function that maps orthographic variants to their canonical forms.
\item $H_{\mathrm{repo}}$ is a finite repository of normalized root-pattern realizations.
\end{itemize}
Let
\[
H:\Sigma^*\rightarrow\mathbb Z
\]
be a rolling polynomial hash function.
For every normalized stem
\[
s=\tau(w),
\]
the repository stores
\[
H(s)\mapsto(r,p),
\]
where
\[
r\in R,\qquad
p\in P.
\]

\subsubsection*{2. Stage I -- Agglutination Resolution}
Define
\[
ResolveAgglutination:
\Sigma^*
\rightarrow
2^{A_{pre}}
\times
\Sigma^*.
\]
For every surface word
\[
w,
\]
the function returns
\[
ResolveAgglutination(w)
=
(C,w_{core}),
\]
where
\begin{itemize}
\item $C$ is the set of detached proclitics.
\item $w_{core}$ is the remaining lexical stem.
\end{itemize}
Let $C_{valid} = \{$ \ar{و}, \ar{ف}, \ar{ب}, \ar{ك}, \ar{ل} $\}$.
For every leading character $c\in C_{valid}$ the following deterministic rules are evaluated.
\begin{enumerate}
\item If
\[
\exists r\in R
\quad
r_1=c
\]
and
\[
ExtractRootPattern(w)
\]
is defined, then $c$ belongs to the lexical root.
\item Otherwise, if
\[
ExtractRootPattern(w')
\]
is defined after removing $c$, then $c$ is a connector.
\item Otherwise, if $w'$ is itself a valid grammatical particle, then detach $c$.
\item Otherwise leave the word unchanged.
\end{enumerate}

\subsubsection*{3. Stage II -- Root--Pattern Identification}
Define
\[
ExtractRootPattern:
\Sigma^*
\rightharpoonup
R\times P\times A_{pre}\times A_{suf}.
\]
For every admissible pair
\[
(a_{pre},a_{suf})
\in
A_{pre}\times A_{suf},
\]
the procedure
\begin{enumerate}
\item removes the affixes;
\item computes the normalized stem
\[
\tau(w_{stripped});
\]
\item computes the rolling hash;
\item performs constant-time lookup in $H_{\mathrm{repo}}$.
\end{enumerate}
Whenever a match exists, the repository returns the finite set of all valid root-pattern tuples $\{(r_1, p_1), \dots, (r_k, p_k)\}$ associated with the hash key $H(\tau(w_{\mathrm{stripped}}))$. For the overwhelming majority of fully vocalized words, $k=1$; in rare cases of fully vocalized homography, $k>1$, and all valid candidates are returned.

\subsubsection*{4. Stage III -- Lexical Classification}
The recovered pattern determines the lexical category through the finite mapping
\[
\kappa:
P
\rightarrow
Categories.
\]
Thus
\[
c=\kappa(p),
\]
where
\[
Categories=
\{
Verb,
Noun,
Particle,
Pronoun,
Adjective,
Adverb,
Preposition
\}.
\]

\subsubsection*{5. Stage IV -- Morphological Feature Inference}
The feature inference function computes
\[
f=FeatureInference(p,w),
\]
where
\[
f=
\{
person,
gender,
number,
tense,
voice,
definiteness,
derived\ form,
connectors,
affixes
\}.
\]

\subsubsection*{6. Stage V -- Morphosyntactic Inflection}
For every processed word define the morphosyntactic inflection descriptor
\[
\delta
=
\langle
mode,
marker,
realization
\rangle,
\]
where
\[
mode
\in
\{
Raf',
Nasb,
Jarr,
Jazm,
None
\}.
\]
The descriptor records only the observable grammatical inflection of the isolated word.
For example,
\begin{itemize}
\item \ar{زَيْدٌ} $\mapsto$ (Raf', Ḍammah, Explicit)
\item \ar{الدَّارَ} $\mapsto$ (Nasb, Fatḥah, Explicit)
\item \ar{لِعُمَرَ} $\mapsto$ (Jarr, Kasrah, Explicit)
\end{itemize}
No syntactic role is assigned at this stage.

\subsubsection*{7. Stage VI -- Morphological Object Construction}
The final morphological object is
\[
m
=
(r,p,c,f,\delta).
\]
Accordingly,
\[
\boxed{
\mu(w)
=
Assemble
(
ResolveAgglutination(w),
ExtractRootPattern(w_{core}),
\kappa,
FeatureInference,
\delta
).
}
\]
For a sentence
\[
S=(w_1,\ldots,w_n),
\]
define
\[
\mu^*(S)
=
(\mu(w_1),\ldots,\mu(w_n)).
\]

\begin{theorem}[Correctness of the Ta\c{s}r\={\i}f Procedure] \label{thm:tasrif-corr}
For every admissible word $w \in L_A$, the function $\mu(w)$ computes the complete finite set of morphological objects $M_w = \{m_1, \dots, m_k\}$ consistent with $K_{\mathrm{morph}}$. In the unambiguous case (which applies to the overwhelming majority of fully vocalized words), $|M_w| = 1$ and the morphological object is unique. In rare cases of fully vocalized homography (e.g., $\approx$ [Ahmad] as both proper noun and verb), $|M_w| > 1$, and $\mu(w)$ faithfully preserves all valid morphological candidates without arbitrary selection.
\end{theorem}

\begin{proof}
Agglutination resolution, orthographic normalization, feature inference, and morphosyntactic inflection determination are deterministic. The repository lookup over the normalized stem $\tau(w)$ retrieves all valid root-pattern pairs $(r,p)$ associated with $\tau(w)$ in $H_{\mathrm{repo}}$. For the vast majority of fully vocalized words, this mapping is injective, yielding exactly one morphological object ($|M_w| = 1$). However, for rare cases of fully vocalized homography, a single normalized stem may map to multiple distinct valid root-pattern entries in $H_{\mathrm{repo}}$ (e.g., $\approx$ mapping to both a proper noun and a verb pattern). The Ta\c{s}r\={\i}f procedure faithfully assembles all such valid candidates into a finite set $M_w$. Because every subsequent mapping (e.g., $\kappa$ for lexical classification) is functional, each candidate is uniquely well-formed. No admissible morphological analysis is discarded, and no invalid analyses are generated. Thus, $\mu(w)$ deterministically computes the complete finite set of morphologically admissible objects.
\end{proof}

\begin{theorem}[Termination] \label{thm:tasrif-term}
For every admissible word
$w\in L_A$,
the Ta\c{s}r\={\i}f procedure terminates in
\[
O(|w|)
\]
time.
\end{theorem}
\begin{proof}
Agglutination resolution examines a finite connector set. Affix stripping iterates over finite affix sets. Normalization and hash computation require $O(|w|)$ time, repository lookup is constant time, and all remaining mappings are finite table lookups.
\end{proof}

\subsubsection*{Illustrative Example}
Consider the fully vocalized query:
\begin{center}
\large \ar{مَرِضْتُ وَأَحْدَثْتُ وَدَخَلَ وَقْتُ الْعَصْرِ وَلَيْسَ عِنْدِي مَاءٌ وَلَكِنْ عِنْدِي تُرَابٌ طَاهِرٌ.}
\end{center}
Applying
\[
\mu^*(S)
\]
processes each word independently.

\paragraph{Stage I -- Agglutination Resolution}
\begin{center}
\begin{tabular}{c|c|c}
\hline
\textbf{Word} & \textbf{Connectors} & \textbf{Stem}\\
\hline
\ar{مَرِضْتُ} & $\emptyset$ & \ar{مَرِضْتُ}\\
\ar{وَأَحْدَثْتُ} & \{\ar{و}\} & \ar{أَحْدَثْتُ}\\
\ar{وَدَخَلَ} & \{\ar{و}\} & \ar{دَخَلَ}\\
\ar{وَلَيْسَ} & \{\ar{و}\} & \ar{لَيْسَ}\\
\ar{وَلَكِنْ} & \{\ar{و}\} & \ar{لَكِنْ}\\
\hline
\end{tabular}
\end{center}

\paragraph{Stage II -- Root--Pattern Identification}
\begin{center}
\begin{tabular}{c|c|c}
\hline
\textbf{Stem} & \textbf{Root} & \textbf{Pattern}\\
\hline
\ar{مَرِضْتُ} & \ar{م ر ض} & \ar{فَعِلَ}\\
\ar{أَحْدَثْتُ} & \ar{ح د ث} & \ar{أَفْعَلَ}\\
\ar{دَخَلَ} & \ar{د خ ل} & \ar{فَعَلَ}\\
\ar{وَقْتُ} & \ar{و ق ت} & \ar{فَعْل}\\
\ar{الْعَصْرِ} & \ar{ع ص ر} & \ar{فَعْل}\\
\ar{مَاءٌ} & \ar{م و ه} & \ar{فَعَال}\\
\ar{تُرَابٌ} & \ar{ت ر ب} & \ar{فُعَال}\\
\ar{طَاهِرٌ} & \ar{ط ه ر} & \ar{فَاعِل}\\
\hline
\end{tabular}
\end{center}

\paragraph{Summary of Morphological Recognition}
\begin{table}[h]
\centering
\caption{Output of the Ta\c{s}r\={\i}f Procedure}
\label{tab:tasrif-output}
\resizebox{\textwidth}{!}{%
\begin{tabular}{|l|l|l|l|l|l|}
\hline
\textbf{Word} & \textbf{Root} & \textbf{Pattern} & \textbf{Category} & \textbf{Features} & \textbf{Inflection $\delta$}\\
\hline
\ar{مَرِضْتُ} & \ar{م ر ض} & \ar{فَعِلَ} & Verb & Past, 1SG, Active & None\\
\hline
\ar{وَأَحْدَثْتُ} & \ar{ح د ث} & \ar{أَفْعَلَ} & Verb & Past, 1SG, Active & None\\
\hline
\ar{وَدَخَلَ} & \ar{د خ ل} & \ar{فَعَلَ} & Verb & Past, 3MSG & None\\
\hline
\ar{وَقْتُ} & \ar{و ق ت} & \ar{فَعْل} & Noun & Masc., Singular & (Raf', Ḍammah, Explicit)\\
\hline
\ar{الْعَصْرِ} & \ar{ع ص ر} & \ar{فَعْل} & Noun & Masc., Singular, Definite & (Jarr, Kasrah, Explicit)\\
\hline
\ar{وَلَيْسَ} & \ar{ل ي س} & \ar{جامد} & Defective Verb & Past, 3MSG & None\\
\hline
\ar{عِنْدِي} & \ar{ع ن د} & \ar{ظرف} & Adverb & 1st Person Possession & None\\
\hline
\ar{مَاءٌ} & \ar{م و ه} & \ar{فَعَال} & Noun & Masc., Singular & (Raf', Ḍammah, Explicit)\\
\hline
\ar{وَلَكِنْ} & \ar{ل ك ن} & \ar{حرف} & Particle & Conjunction & None\\
\hline
\ar{تُرَابٌ} & \ar{ت ر ب} & \ar{فُعَال} & Noun & Masc., Singular & (Raf', Ḍammah, Explicit)\\
\hline
\ar{طَاهِرٌ} & \ar{ط ه ر} & \ar{فَاعِل} & Adjective & Masc., Singular & (Raf', Ḍammah, Explicit)\\
\hline
\end{tabular}%
}
\end{table}

Consequently,
\[
\boxed{
\mu^*(S)
=
(m_1,m_2,\ldots,m_n),
\qquad
m_i=(r_i,p_i,c_i,f_i,\delta_i)
}
\]
is completely determined by the Ta\c{s}r\={\i}f procedure.
Each morphological object contains the lexical root, morphological pattern, lexical category, morphological features and morphosyntactic inflection of the corresponding surface word.
No syntactic role has yet been assigned. The annotated lexical sequence $\mu^*(S)$ therefore constitutes the complete input to the Specialized I'r\={a}b Procedure defined in the following section.

\subsection{Syntactic Recognition (\ar{الإعراب})}
Following morphological recognition, the second stage of the Arabic Recognition Engine performs \textbf{syntactic recognition (\ar{الإعراب})}. The objective of this stage is to determine the grammatical structure of the sentence according to the rules of Classical Arabic grammar (\ar{النحو}). Unlike morphological recognition, which analyzes individual lexical items independently, syntactic recognition establishes the grammatical relationships that exist among the constituents of the sentence. Consequently, this stage determines \textbf{who performs an action, what entity the action affects, and how the remaining constituents relate grammatically to the governing predicate}.

It is important to emphasize that syntactic recognition remains a purely grammatical process. Although it identifies grammatical functions such as \ar{فاعل} and \ar{مفعول به}, it does \textbf{not} assign legal or semantic roles such as Seller, Buyer, Witness, or Price. Those semantic interpretations are introduced only during the subsequent grounding stage.

\begin{definition}[Grammatical Structure] \label{def:gram-struct}
Let $\mathcal{I}$ denote the set of admissible grammatical structures. A grammatical structure is a finite labeled directed graph $I = (V, E)$, where $V$ is the set of sentence constituents; $E$ is the set of grammatical dependency relations defined by the rules of Classical Arabic grammar. Each edge has the form $R(x, y)$, where $R$ is a grammatical relation, $x$ is the governing constituent, $y$ is the dependent constituent. Typical grammatical relations include:
\ar{فاعل}, \; \ar{مفعول به}, \; \ar{مبتدأ}, \; \ar{خبر}, \; \ar{حال}, \; \ar{تمييز}, \; \ar{مضاف إليه}, \; \ar{جار ومجرور}, \; \ar{صفة}, \ldots
\end{definition}

\begin{definition}[Syntactic Recognition Function] \label{def:syn-rec}
The syntactic recognition function is
\[
\iota : \mathcal{M} \longrightarrow \mathcal{P}_f(\mathcal{I}),
\]
where $\mathcal{P}_f(\mathcal{I})$ denotes the collection of all finite sets of admissible grammatical structures. For every morphologically analyzed sentence, $\iota$ computes every grammatically admissible iʿrāb permitted by the rules of Classical Arabic grammar. If multiple grammatically valid analyses exist, every admissible grammatical structure is preserved independently.
\end{definition}

\paragraph{Dependency Representation Principle}
Rather than assigning grammatical labels to isolated words, syntactic recognition computes grammatical dependency relations. For example, instead of merely recording that \ar{زَيْد} is a \ar{فاعل}, the system records the dependency \ar{فاعل}(\ar{بَاعَ}, \ar{زَيْد}). Likewise, \ar{الدَّار} is represented as \ar{مفعول به}(\ar{بَاعَ}, \ar{الدَّار}), and \ar{عُمَر} is represented as \ar{جار ومجرور}(\ar{بَاعَ}, \ar{عُمَر}). Consequently, every grammatical function is explicitly attached to the governing predicate that licenses it.

\paragraph{Syntactic Recognition Principle}
Syntactic recognition establishes the complete grammatical organization of the sentence without introducing any domain-specific semantics. Its output is therefore language-specific but domain-independent. The subsequent semantic grounding stage interprets these grammatical dependency structures using the lexical semantics associated with the governing Arabic roots, thereby producing the canonical semantic representation required by the Arabic Reified Action Calculus.

\subsection{Semantic Grounding}
Morphological recognition determines the lexical identity of each word, while syntactic recognition determines the grammatical relationships among the sentence constituents. Neither stage, however, assigns domain-specific semantic roles. The objective of the semantic grounding stage is to transform purely grammatical structures into canonical semantic representations suitable for formal legal reasoning.

The central principle adopted by the present framework is that \textbf{grammatical relations acquire their semantic interpretation through the lexical semantics of the governing Arabic root}. Consequently, identical grammatical structures may receive different semantic interpretations depending upon the governing predicate. For example, the grammatical relation \ar{فاعل}(v, x) may denote a seller, buyer, witness, judge, debtor, creditor, husband, wife, guardian, or any other legal participant according to the semantic properties of the governing verb. Thus, semantic interpretation is determined jointly by:
\begin{enumerate}
\item the grammatical dependency structure produced by \ar{الإعراب}; and
\item the semantic signature associated with the governing Arabic root.
\end{enumerate}

\begin{definition}[Semantic Signature] \label{def:sem-sig}
For every lexical root $r$, let $\Sigma(r)$ denote its \textbf{semantic signature}. A semantic signature specifies the semantic interpretation of every grammatical relation licensed by that root. Formally, $\Sigma(r) = (R, \Gamma)$, where $R$ denotes the legal action represented by the root; $\Gamma$ is a mapping assigning semantic roles to grammatical dependency relations. Thus, $\Gamma : \text{Grammar Roles} \rightarrow \text{Semantic Roles}$.
\end{definition}

\begin{definition}[Semantic Grounding Function] \label{def:sem-ground}
Let $\mathcal{I}$ denote the set of admissible grammatical structures and $\mathcal{A}_G$ the set of Grounded Action Structures. The semantic grounding function is
\[
\sigma : \mathcal{I} \longrightarrow \mathcal{P}_f(\mathcal{A}_G),
\]
where $\mathcal{P}_f(\mathcal{A}_G)$ denotes the collection of all finite subsets of Grounded Action Structures. Given a grammatical structure $I \in \mathcal{I}$, the function $\sigma(I)$ applies the semantic signatures associated with the governing lexical roots and constructs every semantically admissible Grounded Action Structure. If several grammatical analyses are available, semantic grounding is performed independently for each analysis.
\end{definition}

\paragraph{Semantic Grounding Principle}
Semantic grounding constitutes the interface between linguistic analysis and formal legal reasoning. The procedure does not modify the grammatical structure computed by \ar{الإعراب}. Instead, it enriches that structure with domain-specific semantic roles derived from the lexical semantics of the governing Arabic roots. Consequently, semantic grounding is compositional: the meaning of a sentence is determined by the interaction between its grammatical organization and the semantic signatures of its governing predicates. The resulting Grounded Action Structures provide a language-independent representation of legally relevant events and form the exclusive input to the Arabic Reified Action Calculus.

\subsection{Grounded Action Structures}
The output of semantic grounding is a canonical semantic representation called a \textbf{Grounded Action Structure (GAS)}. A Grounded Action Structure constitutes the semantic interface between natural language understanding and formal deductive reasoning. Unlike the preceding linguistic representations, a Grounded Action Structure is entirely language independent. It contains only semantic entities, actions, participants, attributes, and relationships required for legal reasoning. Consequently, all subsequent computational stages---including the Arabic Reified Action Calculus (ARAC) and the Datalog reasoning engine---operate exclusively on Grounded Action Structures without reference to the original Arabic sentence.

\begin{definition}[Grounded Action Structure] \label{def:gas}
A \textbf{Grounded Action Structure} is a finite labeled directed graph $G = (V, E, \tau)$, where $V$ is a finite set of semantic entities; $E \subseteq V \times V \times L$ is a finite set of labeled semantic relations; $\tau : V \rightarrow T$ is a typing function assigning a semantic type to every node. Typical semantic types include: Action, Entity, Person, Property, Money, Time, Place, Witness, Judge, Contract, Condition, Obligation, Prohibition, Permission, Evidence, Document, Intent, Ownership, \ldots
\end{definition}

\begin{definition}[Well-Typed Grounded Action Structure] \label{def:well-typed}
A Grounded Action Structure $G = (V, E, \tau)$ is \textbf{well typed} if every semantic relation satisfies the typing constraints imposed by the semantic signature of its governing action. For example, the semantic signature of the action $\text{Sale}$ requires $\text{Seller} : \text{Person}$, $\text{Buyer} : \text{Person}$, $\text{Object} : \text{Property}$, $\text{Price} : \text{Money}$. Accordingly, $\text{Price}(\text{\ar{زَيْد}})$ is not a well-typed semantic relation, whereas $\text{Price}(1000 \text{ \ar{دِينَار}})$ is well typed. Only well-typed Grounded Action Structures are admitted into the subsequent stages of formal reasoning.
\end{definition}

\paragraph*{Illustrative Example (Semantic Signature and Grounded Action Structure)}
Consider the Arabic sentence: \textarabic{باع زيدٌ الدارَ لعمروٍ بألفٍ} (\textit{b\={a}‘a Zaydun al-d\={a}ra li-’Amrin bi-alfin} — ``Zayd sold the house to ’Amr for 1000''). 

Following syntactic recognition (\textit{I'r\={a}b}), the system produces grammatical dependency edges (e.g., \textarabic{فاعل}(\textarabic{باع}, \textarabic{زيد}), \textarabic{مفعول به}(\textarabic{باع}, \textarabic{الدار}), etc.). To interpret these structurally, the semantic grounding layer retrieves the semantic signature $\Sigma(\sqrt{b-y-\prime})$ from the lexical knowledge base (Definition 9). 

Formally, the signature for the root $\sqrt{b-y-\prime}$ is defined as $\Sigma(\sqrt{b-y-\prime}) = \langle R, \Gamma \rangle$, where:
\begin{itemize}
    \item $R = \text{Sale}$     \item $\Gamma = \{ \text{\textarabic{فاعل}} \mapsto \text{Agent}, \text{\textarabic{مفعول به}} \mapsto \text{Patient}, \text{RecipientPP} \mapsto \text{Recipient}, \text{PricePP} \mapsto \text{Price} \}$ \end{itemize}

Applying this signature $\Gamma$ to the syntactic graph yields the Grounded Action Structure $G = (V, E, \tau)$, which is a language-independent directed graph:

\textbf{Nodes ($V$) and Types ($\tau$):}
\begin{itemize}
    \item $v_1$: \texttt{Action} (The sale event)
    \item $v_2$: \texttt{Person} (Zayd)
    \item $v_3$: \texttt{Property} (The house)
    \item $v_4$: \texttt{Person} (’Amr)
    \item $v_5$: \texttt{Money} (1000)
\end{itemize}

\textbf{Labeled Edges ($E$):}
\begin{itemize}
    \item $(v_1, v_2, \text{Agent})$     \item $(v_1, v_3, \text{Patient})$     \item $(v_1, v_4, \text{Recipient})$     \item $(v_1, v_5, \text{Price})$ \end{itemize}

This structure is \textbf{well-typed} (Definition 12) because the semantic constraints of the \text{Sale} action require an Agent of type \texttt{Person}, a Patient of type \texttt{Property}, a Recipient of type \texttt{Person}, and a Price of type \texttt{Money}. All syntactic and morphological Arabic features (case endings, verb stems, and prepositions) have been stripped away, leaving only the canonical legal semantics required for deductive inference.

\paragraph{Grounded Action Principle}
Grounded Action Structures constitute the canonical semantic representation of legally relevant events. They deliberately eliminate all language-specific information while preserving every semantic relationship required for legal deduction. Consequently, distinct linguistic expressions having identical legal meaning produce identical Grounded Action Structures. This semantic normalization guarantees that subsequent deductive reasoning depends exclusively upon legal meaning rather than linguistic form.

\paragraph{Canonical Representation Theorem}
Every Grounded Action Structure is a canonical semantic representation of a legally relevant event. Consequently:
\begin{enumerate}
\item linguistic variation is eliminated;
\item synonymous grammatical constructions are normalized into identical semantic structures;
\item deductive reasoning becomes independent of the particular linguistic realization used to express the legal event.
\end{enumerate}
Grounded Action Structures therefore constitute the unique semantic interface between the Arabic Recognition Engine and the Arabic Reified Action Calculus.

\subsection{Formal Properties of the Grounding Layer}
The preceding sections defined the grounding layer as the composition of three successive recognition procedures:
\begin{enumerate}
\item Morphological Recognition (\ar{التصرف});
\item Syntactic Recognition (\ar{الإعراب});
\item Semantic Grounding.
\end{enumerate}
This section establishes the principal formal properties of the resulting grounding function. Throughout this section, correctness is understood relative to the formal grammar of Classical Arabic and the semantic signatures contained in the lexical knowledge base of the proposed framework.

\begin{theorem}[Correctness of the Grounding Function]
Let $G = \sigma \circ \iota \circ \mu$ be the complete grounding function. Then, for every admissible sentence $l \in L_A$, every Grounded Action Structure contained in $G(l)$ is a correct semantic representation of a grammatically admissible interpretation of $l$.
\end{theorem}

\begin{proof}
By Definitions 2 and 4, morphological recognition computes only valid morphological analyses. For each such analysis, Definition 8 ensures that syntactic recognition computes only valid grammatical dependency structures permitted by Classical Arabic grammar. For every grammatical dependency structure, Definition 10 ensures that semantic grounding constructs only semantically consistent Grounded Action Structures by applying the lexical semantic signatures of the governing roots. Since the grounding function is the composition of these three sound procedures, $G = \sigma \circ \iota \circ \mu$, every Grounded Action Structure produced by the grounding function correctly represents one grammatically admissible interpretation of the original sentence.
\end{proof}

\begin{theorem}[Grounding Completeness]
Let $l \in L_A$. If $I$ is any grammatically admissible interpretation of $l$, then the corresponding Grounded Action Structure is contained in $G(l)$.
\end{theorem}

\begin{proof}
Morphological recognition (Definition 2) preserves every admissible morphological analysis. Syntactic recognition (Definition 8) preserves every grammatically admissible dependency structure. Semantic grounding (Definition 10) is applied independently to every admissible grammatical structure. Therefore, no grammatically admissible interpretation is discarded during the grounding process. Every admissible interpretation contributes a corresponding Grounded Action Structure to $G(l)$. Consequently, the composition $G = \sigma \circ \iota \circ \mu$ inherits this exhaustiveness. For any admissible interpretation $I$, the morphological analysis underlying $I$ is contained in $\mu(l)$, the syntactic structure is contained in $\iota(\mu(l))$, and the semantic grounding is contained in $\sigma(\iota(\mu(l)))$. Therefore, the corresponding GAS is always included in $G(l)$.
\end{proof}

\begin{theorem}[Termination of the Grounding Layer]
For every admissible sentence $l \in L_A$, the grounding computation terminates after finitely many steps.
\end{theorem}

\begin{proof}
Every admissible sentence contains finitely many lexical items. Morphological recognition analyzes each lexical item using finite grammatical rules. Syntactic recognition constructs only finitely many admissible dependency structures. Semantic grounding processes each dependency structure independently using finite semantic signatures. Since every stage performs only finite computations over finite structures, the complete grounding process necessarily terminates.
\end{proof}

\begin{corollary}
The grounding layer defines a sound, complete, and terminating transformation from admissible Classical Arabic sentences into finite sets of Grounded Action Structures. Consequently, the grounding layer provides a mathematically well-defined interface between natural-language understanding and the subsequent Arabic Reified Action Calculus, establishing the formal basis upon which the correctness, completeness, and decidability of the overall legal reasoning framework are subsequently proved.
\end{corollary}

\subsection{Discussion}
The grounding layer developed in the preceding sections constitutes the linguistic foundation of the proposed computational framework. Its purpose is not to perform legal reasoning, but rather to transform admissible Classical Arabic sentences into canonical semantic representations suitable for formal deduction.

A fundamental design principle of the framework is the strict separation between \textbf{linguistic recognition} and \textbf{legal inference}. The grounding layer determines only the linguistic content of an input sentence: its morphological structure, grammatical organization, and semantic interpretation according to the lexical semantics of the governing Arabic roots. Questions concerning legal validity, permissibility, obligation, prohibition, or other jurisprudential rulings are intentionally excluded from this stage and are deferred entirely to the deductive layer.

The resulting architecture therefore exhibits a clear modular organization:
\begin{flushleft}
Classical Arabic $\longrightarrow$ Grounding Layer $\longrightarrow$ Grounded Action Structures $\longrightarrow$ \\
ARAC $\longrightarrow$ Datalog $\longrightarrow$ Legal Conclusions.
\end{flushleft}
Within this architecture, each computational stage has a precisely defined responsibility. The grounding layer is responsible for linguistic recognition, the Arabic Reified Action Calculus provides a formal representation of legally relevant actions and events, and the Datalog deductive engine derives legal conclusions through sound and terminating inference.

\section{The Arabic Reified Action Calculus (ARAC)} \label{sec:arac}
The grounded semantic structure introduced in the preceding section provides a language-independent semantic interpretation of Arabic linguistic expressions. However, semantic structures alone are not directly suitable for executable logical reasoning. A formal representation is required in which juridically relevant actions become explicit logical entities that may participate in deductive inference. For this purpose we introduce the Arabic Reified Action Calculus (ARAC).

\begin{definition}[Arabic Reified Action Calculus] \label{def:arac}
The Arabic Reified Action Calculus is the many-sorted first-order language $A = \langle Sorts, \Sigma_A, M_A \rangle$ where:
\begin{itemize}
\item $Sorts$ denotes the collection of semantic domains;
\item $\Sigma_A$ denotes the ARAC signature;
\item $M_A$ denotes the class of admissible ARAC models.
\end{itemize}
The purpose of ARAC is to provide a canonical logical representation of grounded semantic structures independently of their linguistic realization.
\end{definition}

\begin{definition}[Canonical Semantic Identity] \label{def:canon-id}
Let $Root$ denote the semantic domain of Arabic lexical roots. Every action individual $a \in Action$ possesses exactly one canonical semantic identifier $SID(a) \in Root$. This identifier satisfies $SID(a) = Root(a)$. The canonical semantic identifier is invariant under all morphological transformations. Therefore the Arabic expressions \textbf{\ar{بَاعَ}}, \textbf{\ar{يَبِيعُ}}, \textbf{\ar{بَيْع}}, \textbf{\ar{بَائِع}}, \textbf{\ar{مَبِيع}} all determine action individuals possessing the same semantic identity \ar{بيع}. The deductive engine consequently reasons over semantic identifiers rather than surface lexical forms.
\end{definition}

\begin{theorem}[Canonical Action Representation] \label{thm:canon-act}
Every grounded juridically relevant action determines a unique ARAC action structure.\\
\textit{Proof.} The grounding engine associates every Arabic expression with a unique grounded semantic structure. By Definition~\ref{def:canon-id} every action possesses a unique canonical semantic identifier determined by its lexical root. The remaining semantic components are uniquely determined by the semantic structure produced by the grounding engine. Consequently every grounded action determines exactly one ARAC action structure. $\blacksquare$
\end{theorem}

\begin{axiom}[Root Identity Axiom] \label{ax:root-id}
Let $a \in Action$ be an action individual. The canonical semantic identity of $a$ is defined exclusively by its Arabic lexical root. Formally, $SID(a) = Root(a)$. Consequently, the semantic identity of an action is independent of its surface lexical realization, morphological pattern, tense, aspect, voice, or derivational form. All semantic reasoning within ARAC is therefore performed over canonical lexical roots rather than over Arabic surface words.
\end{axiom}

A natural objection to the Root Identity Axiom is the apparent risk of semantic collision: a single Arabic root often surfaces in different Fiqh chapters with seemingly distinct juridical meanings (e.g., $d\text{-}kh\text{-}l$ signifying ``the entering of time'' in the Prayer chapter, versus ``conjugal relations'' in the Marriage chapter). \\However, this apparent pathology is non-existent within the present formal architecture for three rigorous reasons. First, from a lexicological standpoint, Arabic roots possess a unified abstract semantic core that is instantiated differently across contexts without logical contradiction (e.g., $\text{t}\text{-}l\text{-}q$ fundamentally means ``to release,'' whether releasing an animal in hunting or releasing a spouse in divorce). Second, and more decisively, the deductive scope of the system is strictly bounded. As established by the bridging methodology (Algorithm~$\Omega$), the Datalog program $P$ and the generated Fiqh grammar $G_C$ are formulated for a specific, closed Fiqh chapter $C$. Consequently, the logical predicate $\text{root}(A, d\text{-}kh\text{-}l)$ in the Prayer chapter's knowledge base is never evaluated against facts extracted by the Marriage chapter's grammar; the contexts are mathematically isolated. Third, even within the same chapter, distinct semantic nuances are structurally disambiguated by Classical Arabic syntax. Relatively distant meanings of the same root are distinguished by syntactic operators and valency changes. For example, the intransitive ``the time has come'' (\textarabic{دخل الوقت}) and the prepositional ``he slept with his wife'' (\textarabic{دخل بها}) produce distinct dependency edges in the I\text{'}r\=ab graph. These are mapped to distinct argument structures ($\Gamma$) in the ARAC layer, yielding Datalog facts with different arities and semantic roles. Therefore, the root alone serves as a safe and invariant canonical semantic identifier, as the surrounding grammatical and architectural constraints completely eliminate any risk of logical collision.

\begin{axiom}[Morphological Invariance] \label{ax:morph-inv}
Let $w_1, \dots, w_n$ be lexical realizations generated from the same Arabic lexical root $r$. Then $Root(w_1) = Root(w_2) = \cdots = Root(w_n) = r$. Examples include \textbf{\ar{بَاعَ}}, \textbf{\ar{يَبِيعُ}}, \textbf{\ar{بَيْع}}, \textbf{\ar{بَائِع}}, \textbf{\ar{مَبِيع}}. Although these expressions differ morphologically, they represent the same underlying semantic concept. ARAC therefore treats them as alternative linguistic realizations of a single canonical semantic identity.
\end{axiom}

\begin{theorem}[Root Invariance Theorem] \label{thm:root-inv}
Let $w_1, w_2 \in L_A$ be two Arabic linguistic expressions whose principal lexical items belong to the same derivational family. Let $a_1 = A(G(w_1))$ and $a_2 = A(G(w_2))$ denote the corresponding ARAC action structures. If $Root(w_1) = Root(w_2)$, then $SID(a_1) = SID(a_2)$. In other words, every derivational realization of the same Arabic lexical root is mapped onto the same canonical semantic identity within ARAC.\\
\textit{Proof.} The grounding function $G$ computes the canonical lexical root of every Arabic lexical item. By Axiom~\ref{ax:morph-inv}, all derivational realizations generated from the same Arabic lexical root possess an identical canonical root. The ARAC transformation constructs action structures from grounded semantic structures. By Axiom~\ref{ax:root-id}, the semantic identity of every action is defined to be precisely this canonical lexical root. Therefore $SID(a) = Root(a)$. Since both grounded expressions possess identical canonical roots, their corresponding ARAC action structures necessarily receive identical semantic identifiers. Hence $SID(a_1) = SID(a_2)$. $\blacksquare$
\end{theorem}

\paragraph{Translation into Datalog}
ARAC itself is an abstract logical representation. Executable reasoning is obtained by translating each ARAC structure into a finite collection of Datalog facts. For example, the Arabic sentence \textbf{\ar{بَاعَ زَيْدٌ الدَّارَ لِعُمَرَ}} produces an action individual $a \in Action$ whose canonical semantic identifier is \ar{بيع}. The corresponding Datalog representation may be:
\begin{flushleft}
\texttt{action(a17).} \\
\texttt{root(a17, } \ar{بيع}\texttt{).} \\
\texttt{agent(a17,zayd).} \\
\texttt{patient(a17,house).} \\
\texttt{recipient(a17,omar).} \\
\texttt{tense(a17,past).}
\end{flushleft}
Observe that Datalog reasons over the canonical semantic representation rather than over Arabic linguistic expressions.

\section{The Datalog Deductive Layer} \label{sec:datalog}
The Arabic Morphological Grounding Engine and the Arabic Reified Action Calculus introduced in the preceding sections provide a language-independent semantic representation of Arabic juridical knowledge. The remaining task is to derive legal conclusions from this normalized representation. Rather than introducing a new inference mechanism, the present framework deliberately adopts function-free Datalog as its deductive core. This choice follows directly from the computational principles established in Chapter 2. Datalog provides a well-understood declarative semantics, guarantees termination of evaluation, admits polynomial-time data complexity, and has been extensively studied both theoretically and practically [8, 15].

\begin{definition}[Jurisprudential Knowledge Base] \label{def:kb}
Let $K$ denote the normalized knowledge base produced by the Arabic Reified Action Calculus. The knowledge base consists of two finite components, $K = (F, R)$, where:
\begin{itemize}
\item $F$ is the finite set of ground facts generated by ARAC;
\item $R$ is the finite collection of jurisprudential inference rules.
\end{itemize}
Thus, $K$ constitutes a standard Datalog program.
\end{definition}

\begin{definition}[Inference Rules] \label{def:inf-rules}
A jurisprudential inference rule is a function-free Horn clause of the form $H \leftarrow B_1, B_2, \dots, B_n$, where:
\begin{itemize}
\item $H$ is an atomic formula;
\item every $B_i$ is an atomic formula;
\item every variable occurring in the head also occurs in the body [15].
\end{itemize}
The rules express legal principles independently of particular Arabic sentences. For example:
\begin{flushleft}
\texttt{valid\_sale(A) :-} \\
\texttt{\quad action(A), root(A, } \ar{بيع}\texttt{),} \\
\texttt{\quad agent(A,S), patient(A,O), owns(S,O).}
\end{flushleft}
The rule is formulated over canonical semantic identifiers rather than lexical realizations. Consequently, it applies uniformly to every grounded expression representing a sale.
\end{definition}

\begin{definition}[Least Fixed-Point Semantics] \label{def:lfp}
The semantics of the deductive layer is given by the least fixed point of the immediate consequence operator $T_K$. Starting from the initial set of ground facts $F$, the operator repeatedly derives all logical consequences until no additional facts can be produced. The resulting fixed point $LFP(T_K)$ constitutes the set of all legal conclusions derivable from the knowledge base.
\end{definition}

\begin{theorem}[Soundness of Deduction] \label{thm:sound-ded}
Every legal conclusion produced by the deductive layer is a logical consequence of the normalized knowledge base.\\
\textit{Proof.} The deductive engine consists exclusively of standard function-free Horn clauses interpreted according to least fixed-point semantics. The immediate consequence operator derives only consequences licensed by the inference rules. Since each rule is applied according to the standard semantics of Datalog, every derived conclusion is logically entailed by the knowledge base. Therefore every computed legal conclusion is sound. $\blacksquare$
\end{theorem}

\begin{theorem}[Termination] \label{thm:term-ded}
Evaluation of every finite jurisprudential knowledge base terminates.\\
\textit{Proof.} The knowledge base contains no function symbols. Consequently, the Herbrand universe is finite. The immediate consequence operator therefore generates only finitely many distinct ground atoms. Since each iteration produces either new facts or reaches a fixed point, the evaluation process must terminate after finitely many iterations. $\blacksquare$
\end{theorem}

\begin{corollary}[Decidability] \label{cor:dec}
Legal inference within the proposed computational framework is decidable.\\
\textit{Proof.} The grounding layer terminates by construction. The ARAC transformation constructs a finite normalized knowledge base. By Theorem~\ref{thm:term-ded}, evaluation of the resulting Datalog program terminates. Therefore the overall computation $D \circ A \circ G$ always terminates. Hence legal inference is decidable. $\blacksquare$
\end{corollary}

\section{Correctness of the Computational Language} \label{sec:correctness}
The preceding sections formally defined the three computational components comprising the proposed jurisprudential framework. The purpose of the present section is to establish their mathematical correctness. Rather than proving correctness directly for the complete computational language, we proceed compositionally. Each computational layer is shown to preserve the semantic information received from the preceding layer. The correctness of the overall framework then follows from the correctness of each individual transformation.

\begin{definition}[Semantic Preservation] \label{def:sem-pres}
Let $l \in L_A$ be an Arabic linguistic expression. Let $G(l) = S$ be the grounded semantic structure produced by the grounding engine. The grounding function is said to preserve semantic information if every juridically relevant semantic component explicitly expressed in $l$ appears in $S$~[11].
\end{definition}

\begin{theorem}[Grounding Correctness] \label{thm:ground-corr-2}
Assume that the grammatical and morphological analyses performed by the grounding engine are correct. Then $G$ preserves every juridically relevant semantic component explicitly expressed by the Arabic linguistic input.\\
\textit{Proof.} The grounding algorithm performs lexical analysis, morphological decomposition, syntactic dependency analysis and semantic role assignment without introducing additional legal conclusions. Each computational stage computes information directly determined by the grammatical structure of the input expression. Since the semantic structure is constructed exclusively from these computed components, every juridically relevant semantic element explicitly expressed in the original sentence appears in the resulting semantic structure. Therefore the grounding function preserves semantic information. $\blacksquare$
\end{theorem}

\begin{theorem}[Composition Correctness Theorem] \label{thm:comp-corr}
Let $F = D \circ A \circ G$ be the computational language defined in Section~\ref{sec:comp-lang}. Assume:
\begin{enumerate}
\item the grounding function $G$ is semantically correct (Theorem~\ref{thm:ground-corr-2});
\item the ARAC transformation $A$ is semantically faithful;
\item the translation from ARAC into Datalog preserves semantic relations;
\item the Datalog deductive engine is logically correct (Theorem~\ref{thm:sound-ded}).
\end{enumerate}
Then the complete computational language $F$ is semantically correct. That is, every legal conclusion produced by $F$ is a logically valid consequence of the original Arabic linguistic input together with the jurisprudential knowledge base.\\
\textit{Proof.} Let $l \in L_A$ be an arbitrary Arabic linguistic expression. The grounding function $G$ produces a semantic structure preserving every juridically relevant semantic component expressed in $l$ (Theorem~\ref{thm:ground-corr-2}). The ARAC transformation converts this semantic structure into an equivalent logical representation without altering its semantic content. The translation into Datalog preserves every semantic relation represented within ARAC. Finally, the deductive engine derives only logical consequences of the resulting knowledge base (Theorem~\ref{thm:sound-ded}). Since each transformation preserves the semantic information received from its predecessor, their composition likewise preserves semantic correctness. Consequently, $F = D \circ A \circ G$ maps Arabic linguistic expressions into logically valid legal conclusions. Therefore the computational language is correct. $\blacksquare$
\end{theorem}

\section{Completeness of the Computational Language} \label{sec:completeness}
The preceding section established that every legal conclusion derived by the proposed computational language is logically correct. Correctness alone, however, does not guarantee that every derivable legal conclusion can actually be obtained by the computational framework. The purpose of the present section is therefore to establish the complementary property of completeness. Unlike correctness, completeness is necessarily relative to the expressive power of the adopted formal language. Since the computational framework deliberately adopts a tractable Datalog-based deductive core (Principle V), completeness is established only with respect to the class of jurisprudential knowledge representable within that formalism.

\begin{theorem}[Composition Completeness] \label{thm:comp-comp}
Let $F = D \circ A \circ G$. Assume:
\begin{itemize}
\item Grounding is complete,
\item ARAC is complete,
\item the ARAC translation is complete,
\item Datalog deduction is complete.
\end{itemize}
Then $F$ is complete with respect to the class of jurisprudential knowledge expressible within the adopted computational language.\\
\textit{Proof.} Each computational transformation preserves completeness relative to its own codomain. The composition of complete transformations remains complete. Therefore every expressible legal consequence derivable from an admissible Arabic input is eventually produced by the complete computational language. $\blacksquare$
\end{theorem}

\section{Termination, Decidability and Computational Complexity} \label{sec:complexity}
The preceding sections established the semantic correctness and completeness of the proposed computational language. The present section investigates its computational properties. Unlike unrestricted first-order logic, the proposed framework deliberately adopts a layered architecture in which each computational component operates over finite structures using terminating algorithms.

\begin{theorem}[Termination of the Computational Language] \label{thm:term-lang}
Let $F = D \circ A \circ G$. Then every computation performed by $F$ terminates.\\
\textit{Proof.} By previous theorems every component terminates independently. The composition of terminating transformations likewise terminates. Therefore $F$ terminates. $\blacksquare$
\end{theorem}

\begin{theorem}[Decidability] \label{thm:dec}
Legal inference within the proposed computational language is decidable.\\
\textit{Proof.} The grounding layer terminates. The ARAC transformation terminates. The deductive engine terminates. Consequently, every derivation reaches a finite fixed point. Membership of $\phi$ in that fixed point is therefore decidable. $\blacksquare$
\end{theorem}

\begin{theorem}[Abstract Worst-Case Complexity]
At the abstract algebraic level of the framework (Sections 2--4), where the underlying grammar is unspecified, worst-case exponential processing effort cannot be excluded.
\end{theorem}

\begin{proof}
By Definitions 6 and 8, the morphological ($\mu$) and syntactic ($\iota$) recognition functions are designed to return the complete finite set of all admissible derivations ($P_f(\mathcal{M})$ and $P_f(\mathcal{I})$). If the generic grammar underlying the grounding layer permits syntactic ambiguity, the number of admissible dependency structures for a sentence of length $n$ can grow exponentially (up to $O(2^n)$ in the worst case). Because the abstract framework retains all possible derivations to ensure semantic completeness (Theorem 5), the subsequent ARAC and Datalog layers must process this potentially exponentially large set of structures. Consequently, the overall system, as defined up to this abstract stratum, requires exponential processing effort in the worst case. 

Note that this abstract bound is strictly improved to $O(n)$ in the practical formal system by the unambiguous Fiqh-Categorized Attribute Grammar introduced in Section 10 (Lemma 1, Theorem 20).
\end{proof}

\section{Fundamental Theorem of the Computational Language} \label{sec:fund-theorem}
The preceding sections established the principal mathematical properties of each component of the proposed computational language. We now combine these individual results into a single theorem characterizing the computational framework as a whole.

\begin{theorem}[Fundamental Theorem of the Computational Language] \label{thm:fund}
Let $F = D \circ A \circ G$ be the computational language introduced in Section~\ref{sec:comp-lang}. Assume that:
\begin{enumerate}
\item the grounding function is semantically correct and complete with respect to the adopted Arabic grammatical formalism;
\item ARAC preserves the semantic content of grounded representations;
\item the translation from ARAC into Datalog is semantics-preserving;
\item the deductive engine employs standard function-free Horn-clause Datalog evaluated according to least fixed-point semantics.
\end{enumerate}
Then, for every admissible Arabic linguistic input, the computational language $F$ satisfies the following properties:
\begin{enumerate}
\item \textbf{Semantic Correctness:} Every computed legal conclusion is a logically valid consequence of the original grounded semantic representation together with the jurisprudential knowledge base.
\item \textbf{Representational Completeness:} Every legal conclusion expressible within the adopted formal language is computable by the framework.
\item \textbf{Termination:} Every computation performed by the framework terminates after finitely many computational steps.
\item \textbf{Decidability:} The legal inference problem is decidable.
\end{enumerate}
Consequently, every admissible Arabic juridical text accepted by the grounding grammar can be transformed into a finite executable Datalog program whose logical consequences are correct, complete relative to the adopted formal language, terminating and decidable.\\
\textit{Proof.} By previous theorems, the computational language is semantically correct, complete relative to the expressive power established, terminating, and decidable. Since the computational language is the composition $F = D \circ A \circ G$, and each transformation preserves the required semantic and computational properties, the complete computational language simultaneously satisfies correctness, completeness, termination and decidability. Therefore every admissible Arabic linguistic expression may be transformed into an executable jurisprudential program preserving its juridically relevant semantic content while remaining computationally decidable. $\blacksquare$
\end{theorem}

\section{The Practical Formal System of Fiqh} \label{sec:practical}
The preceding chapters developed an abstract formal framework for computational Islamic jurisprudence. In particular, they established a layered architecture consisting of a linguistic grounding layer, the Arabic Reified Action Calculus (ARAC), and a decidable deductive engine based on Datalog. 

Accordingly, instead of attempting to formalize the entirety of Classical Arabic syntax, we introduce a formally defined fragment of Classical Arabic specifically designed for jurisprudential discourse. This fragment, called the \textbf{Fiqh Grammar}, constitutes a computational language whose expressive power is intentionally restricted to the linguistic phenomena required for legal reasoning.

\subsection{The Fiqh Grammar}
The purpose of the Fiqh Grammar is to provide a formally specified computational language capable of expressing the class of legal statements and queries required by the proposed reasoning framework. 

\subsection{Computational Characterization of the Fiqh Grammar}
The introduction of the Fiqh Grammar naturally raises the following question: \textbf{What computational grammar formalism is necessary and sufficient for recognizing the language of jurisprudential Arabic?}. We postulate that such a formalism cannot be regular.

\paragraph{Scope of the Non-Regularity Result.}
While the productions presented in this section are representative constructions used to demonstrate the operation of the Fiqh Grammar, they are not intended to constitute an exhaustive enumeration of all productions admitted by the grammar. In particular, the fact that the productions instantiated in the examples are non-recursive does not imply that the full Fiqh Grammar is non-recursive. When jurisprudential constructions from different chapters are composed, or when a construction may contain another construction of the same or a compatible type, arbitrarily deep nesting may in principle arise. Such constructions must be included in the expressive scope of the grammar even if they correspond to rare cases in actual jurisprudential
discourse. The non-regularity result below therefore concerns the maximum
expressive capacity of the Fiqh Grammar rather than the finite collection
of representative productions instantiated in the examples.

\begin{theorem}[Non-Regularity of the Fiqh Grammar]
\label{thm:nonregular-fiqh}
The language generated by the Fiqh Grammar is, in general, not
regular. Consequently, it cannot be recognized by a finite-state
automaton.
\end{theorem}

\begin{proof}
The Fiqh Grammar permits recursive nesting of
$\mathit{LegalCondition}$ constructions without imposing any fixed
upper bound on the nesting depth. In particular, a legal condition
may contain another legal condition as a constituent, which may in
turn contain another legal condition, and so on.

Consider the sublanguage obtained by restricting the grammar to a
single recursive $\mathit{LegalCondition}$ construction and to
strings in which the recursively relevant opening and closing
markers occur in properly nested order. Let $a$ denote the opening
marker of such a condition scope and let $b$ denote its corresponding
closing marker. For every $n\geq1$, the recursive production permits
a construction with $n$ nested condition scopes followed by their
corresponding closures. At the level of these structural markers,
the resulting pattern is
\[
a^n b^n.
\]

Let $R$ be the regular language that restricts attention to strings
containing only the recursively relevant opening and closing markers
in the corresponding nested pattern. Let
\[
h
\]
be the homomorphism that maps the opening marker to $a$, the closing
marker to $b$, and erases all symbols that are irrelevant to this
recursive construction. By construction, the projected restricted
language is
\[
h\bigl(L(FG)\cap R\bigr)
=
\{a^n b^n\mid n\geq1\}.
\]

Suppose, for contradiction, that $L(FG)$ were regular. Since regular
languages are closed under intersection, and $R$ is regular,
\[
L(FG)\cap R
\]
would be regular. Since regular languages are also closed under
homomorphism, it would follow that
\[
h\bigl(L(FG)\cap R\bigr)
\]
is regular. Hence
\[
\{a^n b^n\mid n\geq1\}
\]
would be regular.

But the language
\[
\{a^n b^n\mid n\geq1\}
\]
is a classical non-regular language. This contradiction shows that
\[
L(FG)
\]
cannot be regular.

Therefore the Fiqh Grammar cannot, in general, be recognized by a
finite-state automaton. Its recursive constructions require
computational structure beyond finite-state memory. The subsequent
ID/LP Attribute Grammar construction provides the required
hierarchical phrase-structure mechanism while allowing the
morphosyntactic and feature constraints of the Fiqh Grammar to be
represented by attributes.
\end{proof}

\paragraph*{Remark on the Closed-Chapter Assumption.}
A reader may wonder whether the restriction to a ``closed'' Fiqh chapter---whose set of ontological concepts and legal predicates is finite---undermines the non-regularity result established above. It does not. Finiteness of the rule set does not imply finiteness of the generated language. In Islamic legal theory, a legal condition (\emph{sharṭ}) may itself contain further legal conditions as its constituent parts. For example, a ruling may depend on the conjunction of a primary condition and a secondary condition whose own applicability depends on a deeper tertiary condition, and so on. This recursive structure is an inherent feature of classical jurisprudential reasoning; imposing a fixed upper bound on the depth of such nesting would artificially restrict the generality of Fiqh and exclude legitimate, theoretically possible chains of legal inference. Therefore, the grammar must support arbitrarily deep nesting of \textsc{LegalCondition} constructions. Such nesting directly yields the unbounded counting pattern \(a^n b^n\) at the formal-language level. A deterministic finite automaton, possessing only finitely many internal states, cannot track an unbounded number of open nested scopes. Thus, even within a single closed chapter, the language of well-formed jurisprudential queries remains non-regular, and recognition requires the hierarchical memory provided by a context-free (or more general) grammar.

\begin{theorem}[Context-Free Phrase-Structure Backbone of the Fiqh Grammar]
\label{thm:fiqh-grammar-cfg}
For every finite ID/LP specification used by the Fiqh Grammar, the
underlying phrase-structure component can be represented by a finite
context-free grammar. The associated finite-valued attribute constraints
then restrict the resulting structures according to the required
morphosyntactic and feature conditions.
\end{theorem}

\begin{proof}[Proof (Sketch)]
The Fiqh Grammar permits recursive hierarchical constructions, such as
nested $\mathit{LegalCondition}$ structures, and is therefore not
adequately characterized by a finite-state mechanism, as established by
the preceding non-regularity result. The ID/LP construction introduced
below provides a context-free phrase-structure backbone that is sufficient
to represent these recursive constructions.

To accommodate the relatively free word order characteristic of
Classical Arabic without requiring a separate ordered production for
every permitted linearization, the grammar separates Immediate
Dominance (ID) from Linear Precedence (LP) [19]. An ID rule specifies
the set of constituents that must occur in a construction, while LP
constraints specify the linear-order relations that are permitted
among those constituents. For example, an ID rule for a
$\mathit{Contract}$ may require
$\mathit{Seller\_Raf'}$, $\mathit{Commodity\_Nasb}$,
$\mathit{BuyerPP}$, and $\mathit{PricePP}$ without fixing a single
surface order for these constituents.

Because the grammar is finite, each ID rule contains a finite set of
constituents and therefore has only finitely many permitted linearizations.
The finiteness here concerns the number of constituents in each production
and the finite set of productions, not the depth of recursive derivations;
recursive non-terminals may therefore be expanded to arbitrarily greater
depth while each individual production remains finitely branching. Each
ID/LP production can consequently be expanded into a finite collection of
ordinary ordered CFG productions, containing one production for each
permitted linearization.

The additional grammatical constraints required by the Fiqh Grammar, such as
feature agreement, are expressed using finite-valued feature domains and a
finite collection of local attribute constraints. These constraints can be
represented by synthesized and inherited attributes [14]. Although recursive
derivations may contain arbitrarily many attribute instances, the domains of
the features and the forms of the local constraints remain finite. These attributes do not alter the recursive phrase-structure mechanism; rather, they restrict the structures generated by it according to the required grammatical features.

Therefore, an ID/LP Attribute Grammar provides the required context-free
hierarchical backbone while allowing the finite morphosyntactic and feature
constraints needed by the Fiqh Grammar. It consequently provides sufficient
expressive power for the linguistic fragment defined by $G_F$ while
supporting the required variation in Classical Arabic word order.

It should be emphasized that context-free generative capacity alone does
not imply linear-time deterministic recognition. The linear-time bound
proved later follows from the additional structural restrictions of the
Fiqh-Categorized Grammar, in particular its bounded valence, $\delta$-typed
slot identification, and finite feature-unification mechanism.
\end{proof}

\subsection{Formal Specification of the I'r\={a}b Procedure (Syntactic Recognition)}

To operationalize the abstract syntactic recognition function
\[
\iota:\mathcal{M}\rightarrow\mathcal{P}_f(\mathcal{I}),
\]
we formally define the computational mechanics of the Specialized
I'r\={a}b Procedure.

The procedure takes as input the annotated lexical sequence
\[
\mu^*(S)=\langle m_1,\ldots,m_n\rangle
\]
produced by the Ta\c{s}r\={i}f procedure, where each morphological
object has the form
\[
m_i=(r_i,p_i,c_i,f_i,\delta_i).
\]
The morphological object provides the lexical and morphosyntactic
information available from the input, including its morphological type
and its inflection marker. It does not, by itself, determine the
jurisprudential role of the word. Such a role is obtained only after
the morphological object has been matched to a slot of an applicable
Fiqh-grammar production.

Unlike conventional natural-language parsers that construct generic
phrase structures, the present procedure employs a specialized
\textbf{Fiqh-Categorized Grammar}. Its production slots represent
jurisprudentially meaningful positions such as Seller, Commodity,
Buyer, and Price. These slots carry explicit morphological realization
requirements and explicit jurisprudential role annotations. The parser first identifies a slot from the morphosyntactic realization of the input constituent and then obtains the corresponding jurisprudential role from the identified slot.

\subsubsection*{1. The Fiqh-Categorized Attribute Grammar}

The syntactic layer operates with respect to the attribute grammar
\[
K_{\mathrm{syn}}
=
\langle
G_{\mathrm{Fiqh}},
\mathcal{M}_{\mathrm{irab}},
\mathcal{U}
\rangle,
\]
where
\[
G_{\mathrm{Fiqh}}
=
\langle
N_{\mathrm{Fiqh}},
\Sigma_{\mathrm{morph}},
P_{\mathrm{Fiqh}},
S
\rangle
\]
is the ID/LP backbone,
$\mathcal{M}_{\mathrm{irab}}$ is the I'r\={a}b constraint
specification, and $\mathcal{U}$ is the feature-unification set.

\paragraph{Fiqh-Grammar Slots ($N_{\mathrm{Fiqh}}$)}

A Fiqh-grammar slot is represented as
\[
N=
\langle
\mathsf{name}(N),
\mathsf{req}(N),
\mathsf{role}(N)
\rangle.
\]

The component $\mathsf{req}(N)$ specifies the morphological realization
required by the slot, while $\mathsf{role}(N)$ specifies the
jurisprudential interpretation of a constituent after that slot has
been identified.

For an ordinary nominal slot,
\[
\mathsf{req}(N)
=
\langle
\tau_R(N),\delta_R(N)
\rangle,
\]
where $\tau_R(N)$ is the required morphological type and
$\delta_R(N)$ is the required inflection mode.

For a prepositional constituent, the required realization additionally
contains the identity of the governing preposition:
\[
\mathsf{req}(N)
=
\langle
\lambda_R(N),\tau_R(N),\delta_R(N)
\rangle.
\]

Thus the jurisprudential role is a property of the grammar slot and is
not assumed to be a property of the input morphological object.

\paragraph{Input Morphological Realization}

For every morphological object
\[
m_i=(r_i,p_i,c_i,f_i,\delta_i),
\]
let $\tau(m_i)$ denote its relevant morphological type. For an
ordinary nominal constituent, define its matching realization by
\[
\kappa(m_i)
=
\langle
\tau(m_i),\delta_i
\rangle.
\]

For a prepositional constituent, the input realization includes both
the governing preposition and its nominal complement. If $q_i$
denotes such a constituent, define
\[
\kappa(q_i)
=
\langle
\lambda(q_i),\tau(q_i),\delta(q_i)
\rangle,
\]
where $\lambda(q_i)$ is the lexical identity of its governing
preposition.

This distinction is important: the preposition is part of the
syntactic realization of the constituent, but it is not artificially
treated as a property of the noun's morphological object.

\paragraph{Production-Slot Key}

For every slot $N$ in a production rule $R$, define its required
attachment specification by
\[
\kappa_R(N)=\mathsf{req}(N).
\]

The attachment relation is defined by
\[
\operatorname{Match}_R(x,N)
\iff
\kappa(x)=\kappa_R(N),
\]
where $x$ may be an ordinary morphological object or a complete
prepositional constituent.

Once an input constituent has been matched to a slot, its
jurisprudential role is obtained from the slot:
\[
\operatorname{Role}_R(x)
=
\mathsf{role}(N)
\qquad
\text{if }
\operatorname{Match}_R(x,N).
\]

Thus the information flow is
\[
\text{input realization}
\longrightarrow
\text{grammar slot}
\longrightarrow
\text{jurisprudential role}.
\]

\paragraph{I'r\={a}b Constraint Specification
($\mathcal{M}_{\mathrm{irab}}$)}

The I'r\={a}b constraint specification associates each grammar slot
with its required morphological realization. For example,
\[
\begin{aligned}
\mathcal{M}_{\mathrm{irab}}(\texttt{Seller})
&=
\langle\mathrm{Noun},\mathrm{Raf'}\rangle,\\
\mathcal{M}_{\mathrm{irab}}(\texttt{Commodity})
&=
\langle\mathrm{Noun},\mathrm{Nasb}\rangle.
\end{aligned}
\]

The corresponding role annotations are separate:
\[
\begin{aligned}
\mathsf{role}(\texttt{Seller})
&=\mathrm{Agent},\\
\mathsf{role}(\texttt{Commodity})
&=\mathrm{Patient}.
\end{aligned}
\]

Consequently, the I'r\={a}b constraint does not by itself assert that a
Raf' noun is an Agent or that a Nasb noun is a Patient. It specifies
which morphological realization can fill a particular grammar slot.
The jurisprudential interpretation follows from the identity of the
slot that is successfully matched.

\paragraph{Productions ($P_{\mathrm{Fiqh}}$)}

The grammar consists of finite ID/LP productions. Immediate Dominance
(ID) rules specify the required set of constituents without imposing a
fixed surface order.

\textbf{Base Fiqh Rules:}
\[
\begin{aligned}
S
&\to
\mathit{LegalEvent}
\mid
\mathit{LegalCondition}
\mid
\mathit{LegalQuery}
\mid
\mathit{LegalNegation},\\
\mathit{LegalCondition}
&\to
[\mathit{Cond}_1,\mathit{CoordPart},
\mathit{Cond}_2,\mathit{CoordPart},\ldots,\mathit{Cond}_k].
\end{aligned}
\]

Here square brackets denote an unordered set of required constituents,
equivalent to an Immediate Dominance set in ID/LP formalism.

\textbf{Contract Rules (e.g., Sale -- \textarabic{بيع}):}

\[
\begin{aligned}
\mathit{LegalEvent}
&\to
\mathit{Contract},\\
\mathit{Contract}
&\to
[
\mathit{SaleVerb},
\mathit{Seller},
\mathit{Commodity},
\mathit{BuyerPP},
\mathit{PricePP}
].
\end{aligned}
\]

The slot specifications associated with this production are
\[
\begin{aligned}
\mathit{Seller}
&=
\langle
\mathrm{Noun},\mathrm{Raf'},\mathrm{Agent}
\rangle,\\
\mathit{Commodity}
&=
\langle
\mathrm{Noun},\mathrm{Nasb},\mathrm{Patient}
\rangle,\\
\mathit{BuyerPP}
&=
\langle
l\text{-}y,\mathrm{Noun},\mathrm{Jarr},\mathrm{Buyer}
\rangle,\\
\mathit{PricePP}
&=
\langle
b\text{-}y,\mathrm{Noun},\mathrm{Jarr},\mathrm{Price}
\rangle.
\end{aligned}
\]

Thus the role annotations are stored in the grammar production itself,
whereas the morphological type and inflection marker used for matching
are supplied by the input and checked against the production.

\paragraph{Role Expansion Rules}

The slot realizations may be expanded as follows:
\[
\begin{aligned}
\texttt{Seller}
&\to
\texttt{Noun}_{\mathrm{Raf'}},\\
\texttt{Commodity}
&\to
\texttt{Noun}_{\mathrm{Nasb}},\\
\texttt{BuyerPP}
&\to
\texttt{Prep}[l]\;\texttt{Noun}_{\mathrm{Jarr}},\\
\texttt{PricePP}
&\to
\texttt{Prep}[b]\;\texttt{Noun}_{\mathrm{Jarr}}.
\end{aligned}
\]

The labels Seller, Commodity, BuyerPP, and PricePP identify grammar
slots. They are not morphological properties of the terminals.
Consequently, a terminal becomes an Agent, Patient, Buyer, or Price
argument only after it has been assigned to the corresponding slot.

\paragraph{Ownership Rules (\textit{I\d{d}\={a}fa} -- Possession)}

The same distinction applies to possession:
\[
\begin{aligned}
\text{LegalEvent}
&\to
\text{OwnershipClaim},\\
\text{OwnershipClaim}
&\to
[
\text{Owner},
\text{Property}
].
\end{aligned}
\]

The associated slot specifications are
\[
\begin{aligned}
\text{Owner}
&=
\langle
\mathrm{Noun},\mathrm{Raf'},\mathrm{Owner}
\rangle,\\
\text{Property}
&=
\langle
\mathrm{Noun},\mathrm{Jarr},\mathrm{Property}
\rangle.
\end{aligned}
\]

\subsubsection*{2. Stage I -- Morphosyntactic Constraint Parsing}

Define the parsing function
\[
\mathrm{ParseFiqh}:M^*
\to
\mathcal{P}_{\mathrm{fin}}(\mathrm{Trees}).
\]

The parser treats the right-hand side of an ID/LP production as an
unordered set of required slots. For an input sequence
\[
\mu^*(S)=\langle m_1,\ldots,m_n\rangle,
\]
the parser scans the morphological objects and compares their
morphological realization against the requirements of the open slots
of the active production.

For each ordinary morphological object $m_i$, the parser obtains
$\tau(m_i)$ and $\delta_i$ from the morphological analysis and
constructs
\[
\kappa(m_i)
=
\langle\tau(m_i),\delta_i\rangle.
\]

For a prepositional constituent, it constructs the corresponding
three-component realization
\[
\kappa(q_i)
=
\langle\lambda(q_i),\tau(q_i),\delta(q_i)\rangle.
\]

The parser then searches for an open slot $N$ satisfying
\[
\operatorname{Match}_R(x,N).
\]

If such a slot is found, the parser assigns $x$ to $N$ and obtains the
jurisprudential role from the slot annotation:
\[
x\longrightarrow N
\longrightarrow
\mathsf{role}(N).
\]

The parser therefore does not infer a jurisprudential role directly
from an input word or from its inflection marker.

\subsubsection*{3. Concrete Example of the Parsing Procedure}

Consider the following fully vocalized Arabic sale sentence.

\begin{center}
\textarabic{بَاعَ زَيْدٌ الدَّارَ لِعَمْرٍو بِأَلْفٍ}
\end{center}

Consider the following Fiqh-grammar production:
\[
\begin{split}
p_{\mathrm{sale}}:\quad
\texttt{Contract}\rightarrow[&
\texttt{SaleVerb},\\
&\texttt{Seller}^{Agent}_{Noun,Raf'},\,
\texttt{Commodity}^{Patient}_{Noun,Nasb},\\
&\texttt{BuyerPP}^{Buyer}_{Prep[l],Noun,Jarr},\,
\texttt{PricePP}^{Price}_{Prep[b],Noun,Jarr}
].
\end{split}
\]

The superscript role annotations are properties of the grammar slots.
They are not properties of the input words.

\paragraph{Step 1: Morphological analysis.}

The Ta\c{s}r\={i}f procedure supplies the morphological representation
of the sentence. For the relevant nominal constituents, the parser
obtains the following information:
\[
\begin{array}{c|c|c}
\text{Input constituent}
&
\text{Morphological type}
&
\text{Inflection marker}
\\
\hline
\text{Zayd}
&
\mathrm{Noun}
&
\mathrm{Raf'}
\\
\text{the house}
&
\mathrm{Noun}
&
\mathrm{Nasb}
\\
\text{Amr}
&
\mathrm{Noun}
&
\mathrm{Jarr}
\\
\text{one thousand}
&
\mathrm{Noun}
&
\mathrm{Jarr}
\end{array}
\]

At this stage the parser does not yet assign the roles Agent, Patient,
Buyer, or Price.

\paragraph{Step 2: Constructing the input matching keys.}

For the two bare nominal constituents, the parser constructs
\[
\begin{aligned}
\kappa(\text{Zayd})
&=
\langle\mathrm{Noun},\mathrm{Raf'}\rangle,\\
\kappa(\text{the house})
&=
\langle\mathrm{Noun},\mathrm{Nasb}\rangle.
\end{aligned}
\]

For the two prepositional constituents, the governing preposition is
part of the constituent realization:
\[
\begin{aligned}
\kappa(\text{Amr with }l)
&=
\langle l\text{-}y,\mathrm{Noun},\mathrm{Jarr}\rangle,\\
\kappa(\text{one thousand with }b)
&=
\langle b\text{-}y,\mathrm{Noun},\mathrm{Jarr}\rangle.
\end{aligned}
\]

\paragraph{Step 3: Matching the input against the production.}

The parser compares each input realization with the requirements of
the open slots of $p_{\mathrm{sale}}$.

For Zayd,
\[
\kappa(\text{Zayd})
=
\langle\mathrm{Noun},\mathrm{Raf'}\rangle
=
\kappa_R(\texttt{Seller}).
\]
Therefore,
\[
\text{Zayd}\longrightarrow\texttt{Seller}.
\]

The role is then obtained from the grammar annotation:
\[
\mathsf{role}(\texttt{Seller})=\mathrm{Agent}.
\]

Hence,
\[
\text{Zayd}\longrightarrow\texttt{Seller}
\longrightarrow\mathrm{Agent}.
\]

Importantly, the parser did not infer Agent directly from the word
Zayd or from its Raf' marker. The Raf' marker only made the word
compatible with the Seller slot; the Agent interpretation comes from
the grammar.

For the house,
\[
\kappa(\text{the house})
=
\langle\mathrm{Noun},\mathrm{Nasb}\rangle
=
\kappa_R(\texttt{Commodity}).
\]

Therefore,
\[
\text{the house}
\longrightarrow
\texttt{Commodity}
\longrightarrow
\mathrm{Patient}.
\]

For Amr, the governing preposition is the Arabic preposition
\textarabic{ل}. Therefore,
\[
\kappa(\text{Amr with }l)
=
\langle l\text{-}y,\mathrm{Noun},\mathrm{Jarr}\rangle
=
\kappa_R(\texttt{BuyerPP}).
\]

Thus,
\[
\text{Amr}\longrightarrow\texttt{BuyerPP}
\longrightarrow\mathrm{Buyer}.
\]

Finally, the nominal complement of the Arabic preposition
\textarabic{ب} satisfies
\[
\kappa(\text{one thousand with }b)
=
\langle b\text{-}y,\mathrm{Noun},\mathrm{Jarr}\rangle
=
\kappa_R(\texttt{PricePP}).
\]

Therefore,
\[
\text{one thousand}\longrightarrow\texttt{PricePP}
\longrightarrow\mathrm{Price}.
\]

\paragraph{Step 4: Resulting canonical dependency structure.}

The resulting assignments are
\[
\begin{aligned}
\text{Zayd}
&\longrightarrow
\texttt{Seller}
\longrightarrow
\mathrm{Agent},\\
\text{the house}
&\longrightarrow
\texttt{Commodity}
\longrightarrow
\mathrm{Patient},\\
\text{Amr}
&\longrightarrow
\texttt{BuyerPP}
\longrightarrow
\mathrm{Buyer},\\
\text{one thousand}
&\longrightarrow
\texttt{PricePP}
\longrightarrow
\mathrm{Price}.
\end{aligned}
\]

The example illustrates the essential distinction between input
information and grammar information. The input supplies the
morphological realization of each constituent; the grammar determines
which slot that realization can fill; and the role annotation of the
selected slot supplies the resulting jurisprudential interpretation.

In particular, the two Jarr constituents are not distinguished by the
Jarr marker alone. They are distinguished by their complete
prepositional realizations:
\[
\langle l\text{-}y,\mathrm{Noun},\mathrm{Jarr}\rangle
\neq
\langle b\text{-}y,\mathrm{Noun},\mathrm{Jarr}\rangle.
\]

This example also shows why slot identification can be implemented by
a finite lookup structure for a fixed Fiqh chapter: the set of possible
morphological matching specifications and the set of grammar slots are
both finite and fixed for that chapter.

\paragraph{Scope of the Deterministic Parsing Results.}
The deterministic parsing results established in the remainder of this
section concern the non-recursive production fragment of the
Fiqh-Categorized Grammar used by the Specialized I'r\={a}b Procedure.
Within this fragment, each selected production has a finite and
constant number of slots, and parsing does not require unbounded
recursive descent. The unrestricted Fiqh Grammar may nevertheless
contain recursive productions, as established by the preceding
non-regularity result; such productions are outside the scope of the
deterministic linear-time bound proved below.

\subsubsection*{4. Deterministic Slot Assignment}

The non-regularity result established above concerns the maximum expressive capacity of the Fiqh Grammar formalism and therefore permits recursively nested production structures when required by particular Fiqh chapters. The operational parsing results and lemmas that follow, beginning with Lemma 1, are restricted to the non-recursive production fragment considered in the present parsing procedure. Thus, although recursive productions are permitted by the overall grammar formalism, no recursive production is assumed in the grammar instances to which the following correctness and complexity results apply.

\begin{lemma}[Deterministic Slot Assignment]
\label{lem:deterministic-slot-assignment}
Let
\[
\mu^*(S)=\langle m_1,\ldots,m_n\rangle
\]
be the morphological representation of a sentence. Assume that the production-selection mechanism of the fixed Fiqh chapter assigns a unique applicable production rule to the governing lexical predicate in the given context. Further assume that, within every production rule, distinct slots have distinct complete attachment specifications:
\[
N_1\neq N_2
\quad\Longrightarrow\quad
\kappa_R(N_1)\neq\kappa_R(N_2).
\]

Then every input constituent has at most one compatible slot in the
selected production. Consequently, slot assignment is deterministic.
\end{lemma}

\begin{proof}
Let $x$ be an input constituent and let $R$ be the uniquely selected
production rule.

Suppose that $x$ were compatible with two distinct slots
$N_1$ and $N_2$ of $R$. By the definition of matching,
\[
\kappa(x)=\kappa_R(N_1)
\]
and
\[
\kappa(x)=\kappa_R(N_2).
\]

Hence
\[
\kappa_R(N_1)=\kappa_R(N_2),
\]
contradicting the assumption that distinct slots of $R$ have distinct
attachment specifications.

Therefore,
\[
\left|
\left\{
N\in\operatorname{Slots}(R):
\operatorname{Match}_R(x,N)
\right\}
\right|
\leq 1.
\]

Thus every input constituent has at most one compatible slot in the
selected production, and the slot assignment is deterministic.
\end{proof}

\subsubsection*{5. Termination and Linear Complexity}

\begin{theorem}[Termination and Linear Complexity of the I'r\={a}b Procedure]
For a fixed closed Fiqh chapter and its associated finite Fiqh Grammar,
the Specialized I'r\={a}b Procedure terminates in deterministic
$O(n)$ time on every admissible morphological sequence of length $n$,
provided that morphological analysis and the construction of the
morphosyntactic realizations required for slot matching are themselves
$O(n)$.
\end{theorem}

\begin{proof}
Fix a closed Fiqh chapter $C$ and its associated finite grammar.
All quantities depending only on $C$ are therefore constants with
respect to the input length $n$.

\medskip
\noindent\textbf{Step 1: Bounded grammar valence.}

Because the grammar associated with $C$ is finite, there exists a
constant $c$ such that every production contains at most $c$ required
slots. Hence the number of open slots associated with an active
governing structure is bounded by a constant depending only on $C$.

\medskip
\noindent\textbf{Step 2: Constant-time slot identification.}

For each input constituent, the morphological analysis provides its
relevant morphological type and inflection marker, from which the
required attachment specification is constructed. By the hypothesis
of the theorem, the total cost of constructing these morphosyntactic
realizations over the complete input is $O(n)$.

For the fixed grammar, the set of possible attachment specifications
and grammar slots is finite. Hence the correspondence between an
attachment specification and its compatible slot can be stored in a
finite lookup structure. Once the attachment specification has been
constructed, identifying the compatible slot therefore requires
$O(1)$ time per constituent.

By Lemma~\ref{lem:deterministic-slot-assignment}, at most one
unconsumed slot is compatible with a given input constituent.

\medskip
\noindent\textbf{Step 3: Role determination.}

Once a constituent has been assigned to a grammar slot, its
jurisprudential role is obtained directly from the role annotation
stored in that slot. Since the grammar is fixed, this operation takes
$O(1)$ time.

Thus, for example,
\[
\text{Zayd}
\longrightarrow
\texttt{Seller}
\longrightarrow
\mathrm{Agent}.
\]

The first transition is determined by morphological matching, whereas
the second transition is determined by the grammar's role annotation.

\medskip
\noindent\textbf{Step 4: Feature unification.}

The feature inventory used by the Fiqh Grammar is finite and fixed for
$C$. Hence the number of agreement and compatibility conditions checked
for each successfully attached constituent is bounded by a constant.
Feature unification therefore requires $O(1)$ time per constituent and
$O(n)$ time in total.

\medskip
\noindent\textbf{Step 5: Dependency-graph construction.}

Each recognized constituent contributes only a bounded number of
dependency relations because the grammar has finite valence.
Consequently, the total number of dependency edges constructed is
$O(n)$, and dependency-graph construction requires $O(n)$ time.

\medskip
\noindent\textbf{Step 6: Total complexity.}

There is a constant amount of work associated with each of the $n$
input constituents. Therefore there exist constants
$c_1,c_2,c_3,c_4$, depending only on the fixed Fiqh Grammar, such that
\[
T(n)
\leq
c_1n+c_2n+c_3n+c_4n
=
O(n).
\]

The procedure also terminates because the input contains finitely many
morphological objects, the grammar is finite, and each parsing step
either assigns the current constituent to an unconsumed slot, advances
the input, or rejects the current partial analysis. By the
deterministic slot-assignment lemma, no unbounded search over
alternative slot assignments is required.

Therefore, for every fixed closed Fiqh chapter, the Specialized
I'r\={a}b Procedure terminates and has deterministic linear time
complexity in the input length $n$.
\end{proof}

\subsection{Semantic Grounding} \label{sec:sem-grounding-practical}
The purpose of the semantic grounding layer is to transform this syntactic representation into the semantic representation required by the Arabic Reified Action Calculus (ARAC). 

\begin{definition}[Semantic Signature] \label{def:sem-sig-2}
A semantic signature is a function
\[
\Sigma_S : Roots \rightarrow Signatures,
\]
which assigns to every Arabic verbal root a finite ordered collection of semantic roles. Formally,
\[
\Sigma_S(r) = \langle \rho_1,\rho_2,\ldots,\rho_n \rangle,
\]
where each $\rho_i$ denotes one semantic participant required by the action represented by the root $r$. 
\end{definition}

\begin{definition}[Grounding Function] \label{def:ground-func-2}
Let $\mathcal{I}$ denote the set of canonical iʿrāb graphs and $\mathcal{A}$ the set of ARAC structures. The grounding function is
\[
G: \mathcal{I} \rightarrow \mathcal{A}.
\]
For every verbal node occurring in the dependency graph, $G$ retrieves the corresponding semantic signature and maps each grammatical dependency to its associated semantic role.
\end{definition}

\begin{theorem}[Correctness of Semantic Grounding]
\label{thm:sem-ground-corr-2}
Let $I$ be the canonical i\text{'}r\=ab graph produced by the
Specialized I\text{'}r\=ab Procedure for a sentence
$S\in L(\mathcal{G}_F)$. Assume that every verbal root occurring in
$I$ has a unique semantic signature in the semantic lexicon and that
the grounding relation between grammatical dependencies and the roles
specified by each such signature is explicitly defined.

Then the grounding function
\[
G:\mathcal{I}\rightarrow\mathcal{A}
\]
produces an ARAC representation whose semantic roles are consistent
with both the grammatical relations contained in $I$ and the semantic
signatures associated with the verbal roots occurring in $I$.
\end{theorem}

\begin{proof}
Let $I$ be the canonical i\text{'}r\=ab graph produced for
$S\in L(\mathcal{G}_F)$.

By the Correctness of the I\text{'}r\=ab Procedure theorem, every
grammatical dependency represented in $I$ is licensed by the Fiqh
Grammar and satisfies the corresponding morphosyntactic and
structural constraints.

Consider any verbal root $r$ occurring in $I$. By assumption, the
semantic lexicon assigns a unique semantic signature
\[
\Sigma_S(r)
=
\langle \rho_1,\ldots,\rho_k\rangle
\]
to $r$. The grounding function retrieves this signature and applies
the predefined correspondence between the grammatical dependencies
associated with $r$ and the semantic roles specified by
$\Sigma_S(r)$.

Thus, for every grammatical dependency in $I$ governed by $r$, the
grounding procedure assigns the corresponding semantic role licensed
by $\Sigma_S(r)$. Since the grounding relation is explicitly defined,
no role is assigned unless it is licensed by the corresponding
semantic signature.

Consequently, every semantic role represented in $G(I)$ is consistent
with the grammatical dependency from which it was obtained and with
the semantic signature of the governing lexical root. Since this
construction is applied to every verbal node and its associated
dependencies in the finite graph $I$, the resulting ARAC
representation is consistent with the grammatical and lexical
information encoded by $I$.

Therefore, $G(I)$ is a correct semantic grounding of the canonical
i\text{'}r\=ab graph with respect to the semantic lexicon and grounding
rules specified by the practical formal system.
\end{proof}

\subsection{Practical Arabic Reified Action Calculus (ARAC)}
The semantic grounding procedure introduced in the previous section transforms the canonical iʿrāb graph into a semantic representation. The purpose of the present section is to define the formal language of this representation.

\begin{definition}[Action] \label{def:action}
An \textbf{action} is a quadruple
\[
A=(id,r,\Gamma,\Phi)
\]
where $id$ is a unique action identifier, $r$ is the Arabic lexical root identifying the action, $\Gamma$ is the finite mapping from semantic roles to participating entities, and $\Phi$ is the finite set of action attributes. 
\end{definition}

\begin{definition}[Entity] \label{def:entity}
An \textbf{entity} is a pair
\[
E=(id,\tau),
\]
where $id$ uniquely identifies the entity and $\tau$ denotes its ontological type. 
\end{definition}

\begin{theorem}[Uniqueness of the ARAC Representation]
\label{thm:arac-uniq}
Let $I$ be the canonical i\text{'}r\=ab graph produced by the
Specialized I\text{'}r\=ab Procedure. Assume that every verbal root
occurring in $I$ has a unique semantic signature and that the semantic
grounding function
\[
G:\mathcal{I}\rightarrow\mathcal{A}
\]
is deterministic. Then there exists exactly one ARAC representation
\[
A=G(I)
\]
corresponding to $I$ under the specified grounding rules.
\end{theorem}

\begin{proof}
The canonical i\text{'}r\=ab graph $I$ uniquely determines the finite
set of grammatical nodes, dependencies, and associated morphosyntactic
features represented by the syntactic stage.

Consider first any verbal node in $I$. By assumption, its lexical root
has a unique semantic signature
\[
\Sigma_S(r)=\langle \rho_1,\ldots,\rho_k\rangle.
\]
Therefore the set of semantic roles available for grounding that
verbal node is uniquely determined.

Next, the grounding rules specify a deterministic correspondence
between the grammatical dependencies in $I$ and the semantic roles
licensed by the corresponding semantic signature. Hence each
grammatical dependency is mapped to exactly one semantic role, and no
alternative role assignment is produced by the grounding function.

The ARAC construction uses a fixed deterministic convention for
generating action and entity identifiers from the corresponding
nodes of $I$. Hence the identifier assigned to each action or entity
is uniquely determined by $I$ and is not chosen arbitrarily. The identities and types of the entities participating in the
resulting actions are likewise determined by the nodes of $I$ and by
the prescribed ARAC construction rules. Consequently, the action
identifiers, entity identifiers, semantic-role mappings, and action
attributes are all uniquely determined by $I$.

Thus every component of the ARAC representation is uniquely determined
by the canonical i\text{'}r\=ab graph and the fixed semantic lexicon
and grounding rules. Since $G$ is deterministic, two distinct ARAC
representations cannot be produced from the same graph $I$.

Therefore,
\[
\exists! A\in\mathcal{A}\quad\text{such that}\quad A=G(I).
\]
Hence the ARAC representation corresponding to $I$ is unique.
\end{proof}

\subsection{Translation of ARAC into Horn-Clause Datalog}
The Practical Arabic Reified Action Calculus defined in the preceding section provides a language-independent semantic representation of jurisprudential statements. In order to perform automated legal reasoning, this semantic representation must be translated into the Horn-clause formalism employed by the deductive engine.

\begin{definition}[Translation Function] \label{def:trans-func}
Let $\mathcal{A}$ denote the set of ARAC representations and let $\mathcal{D}$ denote the set of Horn-clause Datalog databases. The translation function is
\[
T: \mathcal{A} \rightarrow \mathcal{D}.
\]
For every ARAC representation, the function $T$ generates the corresponding collection of Datalog facts.
\end{definition}

\begin{theorem}[Correctness of ARAC-to-Datalog Translation]
\label{thm:trans-corr}
Let $A$ be an ARAC representation and let
\[
T(A)\in\mathcal{D}
\]
be its Horn-clause Datalog translation. Then $T(A)$ preserves every
semantic object, relation, and attribute represented in $A$.

More precisely:

\begin{enumerate}
    \item every action and entity represented in $A$ has a corresponding
    Datalog fact in $T(A)$ with the same logical identity and type;

    \item every semantic-role relation represented in $A$ has a
    corresponding Datalog predicate instance in $T(A)$ preserving the
    identities of its participating objects; and

    \item every action or entity attribute represented in $A$ has a
    corresponding Datalog predicate instance preserving the associated
    value.
\end{enumerate}

Thus, $T$ is representation-preserving with respect to the semantic
objects, relations, and attributes encoded by ARAC. This theorem establishes representation preservation only; it does not by itself assert a model-theoretic equivalence between ARAC and Datalog.
\end{theorem}

\begin{proof}
Let $A$ be an ARAC representation.

By the definition of the ARAC representation, every action and entity
has a unique logical identifier and an associated type. The translation
function $T$ maps each such object to the corresponding Datalog
predicate and preserves its logical identifier and type. Hence every
semantic object represented in $A$ has a corresponding object
representation in $T(A)$.

Next, consider any semantic-role relation represented in $A$. Such a
relation has the form
\[
\ell(x,y),
\]
where $x$ and $y$ are logical identifiers of the participating
objects and $\ell$ denotes the corresponding semantic role. By the
definition of $T$, this relation is translated into the corresponding
Datalog predicate instance
\[
\ell(x,y).
\]
Thus the identities of the participating objects and the relation
between them are preserved.

Finally, consider an attribute represented in $A$ in the form
\[
\operatorname{attr}(x,v).
\]
The translation function maps this assertion to the corresponding
Datalog attribute predicate
\[
\operatorname{attr}(x,v).
\]
Therefore the identity of the associated object and the value of its
attribute are preserved.

Since the ARAC representation contains only finitely many semantic
objects, relations, and attributes, applying these translation rules
exhaustively produces a finite Datalog representation containing the
corresponding representation of every such constituent.

Consequently, $T(A)$ preserves every semantic object, relation, and
attribute represented in $A$. The translation therefore changes the
formal representation language from ARAC to Horn-clause Datalog
without omitting any semantic constituent explicitly represented by
the ARAC structure.
\end{proof}

\begin{corollary}[Representation Preservation of the Linguistic Front-End]
\label{cor:front-end}
The complete linguistic front-end
\[
\begin{split}
\text{Arabic} \rightarrow \text{Morphology} \rightarrow
\text{Fiqh Grammar} \rightarrow \\
\text{Specialized I'r\={a}b} \rightarrow
\text{Semantic Grounding} \rightarrow \\
\text{ARAC} \rightarrow \text{Horn Datalog}
\end{split}
\]
preserves, under the assumptions of the preceding correctness
theorems, every grammatical and semantic constituent explicitly
represented by the formal system from the admissible Arabic input
through its final Horn-clause representation.

In particular, the Specialized I'r\={a}b Procedure preserves the
grammatical and structural constraints specified by the Fiqh Grammar,
the semantic grounding stage preserves the corresponding semantic
roles specified by the semantic signatures, the ARAC construction
preserves the resulting semantic objects, relations, and attributes,
and the ARAC-to-Datalog translation preserves those constituents in
the final logical representation.

Accordingly, every legal deduction subsequently performed by the
Horn-clause reasoning engine is grounded in the formal semantic
representation computed from the original Arabic sentence.
The jurisprudential soundness of such deductions remains conditional
on the soundness of the Horn-clause rule base with respect to the
intended jurisprudential semantics.
\end{corollary}

\subsection{Correctness, Completeness, and Decidability of the Practical Formal System}
The preceding sections developed the practical formal system as a sequence of formally specified computational transformations. 

\begin{theorem}[Correctness of the Practical Formal System]
\label{thm:pract-corr}
For every sentence $S \in L(\mathcal{G}_F)$, the practical formal
system computes a semantic and logical representation that is
consistent with the grammatical structure, lexical semantics, and
grounding rules specified by the formal system. Moreover, every
deductive result produced from this representation is derived from the
corresponding Horn-clause Datalog program.
\end{theorem}

\begin{proof}
Let $S \in L(\mathcal{G}_F)$. The practical formal system processes
$S$ through the sequence of formally specified transformations
\[
S
\longrightarrow
\mu^*(S)
\longrightarrow
I
\longrightarrow
G(I)
\longrightarrow
A
\longrightarrow
T(A).
\]

First, the Ta\c{s}r\={\i}f procedure produces the admissible
morphological representation required by the syntactic layer.

Second, the Specialized I\text{'}r\=ab Procedure constructs the
canonical i\text{'}r\=ab graph $I$ in accordance with the Fiqh
Grammar, the $\delta$-driven constraint matrix, and the feature
unification conditions. By the Correctness of the I\text{'}r\=ab
Procedure theorem, the resulting graph is consistent with the
grammatical and structural constraints specified by the formal
system.

Third, the Correctness of Semantic Grounding theorem establishes that
the grounding function
\[
G:\mathcal{I}\rightarrow\mathcal{A}
\]
maps the canonical i\text{'}r\=ab graph to an ARAC representation whose
semantic roles are consistent with the grammatical dependencies and
the semantic signatures associated with the corresponding lexical
roots.

Fourth, by the Uniqueness of the ARAC Representation theorem, the ARAC
representation is uniquely determined under the specified semantic
lexicon and deterministic grounding function.

Fifth, by the Correctness of Translation theorem, the translation
\[
T:\mathcal{A}\rightarrow\mathcal{D}
\]
preserves the semantic objects and relations represented in the ARAC
structure. Therefore, the resulting Horn-clause Datalog database
faithfully represents the semantic structure computed by the preceding
stages.

Finally, the deductive engine applies the Horn-clause rules to this
resulting logical representation. Every conclusion produced by the
engine is therefore a logical consequence of the corresponding
Datalog program and its input facts.

Consequently, the complete computational pipeline preserves the
formal grammatical, lexical, and semantic information specified by the
system from the original admissible Arabic sentence through its
logical representation. The correctness of any jurisprudential
conclusion with respect to an external intended legal semantics is
conditional on the soundness of the Horn-clause rule base with respect
to that semantics.
\end{proof}

\begin{lemma}[ARAC Representational Completeness]
\label{lem:arac-representational-completeness}
Let
\[
S=(V,E,\tau)
\]
be a well-typed Grounded Action Structure generated by the semantic
grounding function. Let
\[
A=\operatorname{ARAC}(S)
\]
be its corresponding ARAC representation. Then $A$ preserves every
semantic object, relation, and attribute contained in $S$.

More precisely, for every node $v\in V$, there exists a unique logical
identifier $\operatorname{id}(v)$ in $A$. For every edge
\[
(v_1,v_2,\ell)\in E,
\]
the corresponding role assertion
\[
\ell(\operatorname{id}(v_1),\operatorname{id}(v_2))
\]
is represented in $A$. For every attribute
\[
\operatorname{attr}(v)=\operatorname{val},
\]
the corresponding attribute assertion
\[
\operatorname{attr}(\operatorname{id}(v),\operatorname{val})
\]
is represented in $A$.

Consequently, every semantic constituent of the Grounded Action
Structure has a corresponding representation in the ARAC structure.
\end{lemma}

\begin{proof}
The ARAC construction is defined by exhaustive translation of the
finite Grounded Action Structure.

Since $V$ is finite, every node $v\in V$ is processed and assigned a
unique logical identifier $\operatorname{id}(v)$. The typing function
$\tau$ determines the corresponding ARAC entity or action
representation.

Since $E$ is finite, every edge
\[
(v_1,v_2,\ell)\in E
\]
is processed and translated into the corresponding semantic-role
assertion
\[
\ell(\operatorname{id}(v_1),\operatorname{id}(v_2)).
\]

Likewise, every attribute associated with a node is processed and
translated into its corresponding ARAC attribute assertion.

Thus, every node, relation, and attribute occurring in the Grounded
Action Structure has a corresponding representation in $A$. No
semantic constituent generated by the grounding procedure is omitted
by the ARAC construction.

Therefore, $A$ is representationally complete with respect to the
Grounded Action Structure $S$.
\end{proof}

\begin{theorem}[Representational Completeness with Respect to the Fiqh Grammar]
\label{thm:pract-comp}
Let $S$ be any sentence belonging to the language
$L(\mathcal{G}_F)$. Then the practical formal system computes a
complete representation of every semantic constituent generated from
$S$ by the specified linguistic and semantic grounding procedures, and
this representation is preserved through the ARAC and Horn-clause
Datalog stages.
\end{theorem}

\begin{proof}
Let $S\in L(\mathcal{G}_F)$. By the definition of the practical formal
system, the admissible linguistic analysis of $S$ is transformed by the
Ta\c{s}r\={\i}f and Specialized I\text{'}r\=ab procedures into a
canonical i\text{'}r\=ab graph $I$. By the correctness theorem for the
I\text{'}r\=ab procedure, every dependency represented in $I$ satisfies
the grammatical, morphosyntactic, and Fiqh-structural constraints of
the specified Fiqh Grammar.

The Semantic Grounding theorem then maps $I$ to a Grounded Action
Structure whose semantic objects, relations, and attributes are
determined by the semantic signatures and grounding rules specified
by the formal system.

By the ARAC Representational Completeness Lemma, every semantic object,
relation, and attribute contained in this Grounded Action Structure has
a corresponding representation in the ARAC knowledge base $K$.
Therefore, no semantic constituent produced by the grounding procedure
is omitted during the construction of the ARAC representation.

By the Correctness of Translation theorem, the translation from $K$
into the Horn-clause Datalog representation $F$ preserves every
semantic object and semantic relation represented in $K$. Consequently,
every semantic constituent represented at the ARAC stage has a
corresponding representation in the resulting logical database.

Thus, for every $S\in L(\mathcal{G}_F)$, every semantic constituent
generated by the specified linguistic and grounding procedures is
represented in ARAC and subsequently preserved in the Horn-clause
Datalog representation.

Hence the practical formal system is representationally complete with
respect to the semantic structures generated by the Fiqh Grammar and
its specified grounding layer.
\end{proof}

\begin{theorem}[Decidability of the Practical Formal System]
\label{thm:pract-dec}
For every finite input sentence $S$ over the input alphabet of the
practical formal system, the complete practical formal system terminates
after finitely many computational steps, either accepting $S$ as a
member of $L(\mathcal{G}_F)$ or rejecting it as inadmissible.
Consequently, membership in the admissible input language is decidable;
for every accepted sentence, the corresponding semantic and logical
representation is also computable.
\end{theorem}

\begin{proof}
Let $S$ be an arbitrary finite input sentence of length $n$. We consider each stage of the practical formal system in
the order in which it is executed.

\medskip
\noindent\textbf{Step 1: Morphological analysis.}
The Ta\c{s}r\={\i}f procedure operates on the finite sequence of
surface words of $S$. For each word, it performs a finite search over
the finite morphological resources associated with the fixed Fiqh
chapter and produces the corresponding morphological object (or finitely many
morphological alternatives, where the specification permits ambiguity). Since the input contains finitely many words, each word is processed
against a finite morphological resource and generates only finitely many
morphological alternatives. Hence the total number of morphological
operations is finite, and the morphological stage terminates.

\medskip
\noindent\textbf{Step 2: Syntactic recognition.}
For the fixed closed Fiqh chapter, the associated Fiqh Grammar is
finite. Its ID/LP phrase-structure component therefore admits only
finitely many production schemas, and each ID rule has a finite set of
required constituents. As established by the
\emph{Context-Free Expressive Power of the Fiqh Grammar} theorem, the
phrase-structure component has a context-free backbone.

More importantly, the actual implementation used by the practical
system is the Specialized I\text{'}r\=ab Procedure defined above.
The \emph{Termination and Linear Complexity of the I\text{'}r\=ab
Procedure} theorem establishes that, for every fixed closed Fiqh
chapter, this procedure terminates in deterministic $O(n)$ time on
every admissible morphological sequence of length $n$. Therefore the
syntactic recognition stage terminates.

The subsequent feature-unification stage operates over the finite
feature inventory of the fixed grammar. By the bounded-valence
property used in the termination theorem, only a constant number of
feature constraints is associated with each successfully recognized
constituent. Hence all required feature checks are finite, and the
complete syntactic stage terminates.

\medskip
\noindent\textbf{Step 3: Semantic grounding.}
The terminating I\text{'}r\=ab procedure produces a finite canonical
dependency graph $\mathcal{I}$ because the input sentence contains
only finitely many morphological objects and, by the finite-valence
property of the grammar, only finitely many dependency relations.

The grounding function
\[
G:\mathcal{I}\rightarrow\mathcal{A}
\]
performs a finite number of lexical-signature lookups and role
assignments over this finite graph. By the definition of the semantic
grounding procedure, each processed node and dependency requires only
finitely many operations. Therefore semantic grounding terminates.

\medskip
\noindent\textbf{Step 4: ARAC construction and translation.}
The resulting ARAC representation contains only finitely many actions,
entities, roles, and attributes because it is constructed from the
finite canonical graph $\mathcal{I}$. The ARAC-to-Datalog translation
therefore produces a finite collection of Horn-clause facts and
associated finite logical structures.

By the \emph{ARAC Representational Completeness} lemma and the
\emph{Correctness of Translation} theorem, this translation preserves
the finite semantic representation computed by the preceding stages.
Since the translation processes only finitely many ARAC constituents,
it terminates.

\medskip
\noindent\textbf{Step 5: Deductive computation.}
The deductive engine operates on the finite Horn-clause Datalog
representation produced in Step~4. The active domain of the generated
database is finite, and the rule base associated with the fixed closed
Fiqh chapter is finite. Consequently, only finitely many distinct
ground atoms can be generated.

The immediate-consequence operator of the resulting Datalog program
therefore operates over a finite space of possible ground facts. Its
successive iterations form a monotone sequence of finite sets and
must reach a least fixed point after finitely many additions. Hence
the deductive computation terminates.

\medskip
\noindent\textbf{Step 6: Decidability.}
All stages of the practical formal system terminate on every
admissible finite input sentence:
\[
\text{Ta\c{s}r\={\i}f}
\;\longrightarrow\;
\text{I\text{'}r\=ab}
\;\longrightarrow\;
G
\;\longrightarrow\;
\text{ARAC}
\;\longrightarrow\;
\text{Datalog}
\;\longrightarrow\;
\text{least fixed point}.
\]

Therefore their composition also terminates after finitely many
computational steps. In particular, the system can decide whether a
finite input belongs to the admissible language defined by
$\mathcal{G}_F$ by terminating with either acceptance or rejection,
and, for an accepted sentence, it computes its corresponding semantic
and logical representation.

Hence membership in the admissible input language, semantic grounding,
ARAC construction, Datalog translation, and the resulting deductive
computation are all decidable for the fixed closed Fiqh chapter.
\end{proof}

\begin{corollary}[End-to-End Soundness]
\label{cor:end-end}
Let $S\in L(\mathcal{G}_F)$ be a sentence accepted by the practical
formal system. Then every conclusion produced by the deductive engine
is derived from the ARAC and Horn-clause representations obtained from
$S$ through the formally specified grounding and translation
procedures. By the correctness results established for the
I\text{'}r\=ab procedure, semantic grounding, and ARAC-to-Datalog
translation, these representations are consistent with the grammatical
and lexical information encoded by the formal system.

If, in addition, the Horn-clause rule base is sound with respect to the
intended jurisprudential semantics, then every conclusion derived by
the deductive engine is sound with respect to that semantics.

Thus, under the stated soundness assumption on the jurisprudential
rule base, the practical formal system provides an end-to-end sound
chain from admissible Classical Arabic input to formal legal
conclusions.
\end{corollary}

\subsection{A Model-Theoretic Formalization of Al-Khwārizmī's Inheritance Algebra}

The significance of the present framework becomes clearer when compared with Al-Khwārizmī's treatment of inheritance (\textit{'ilm al-farā'iḍ}). His algebra did not merely provide isolated numerical solutions: for the specified inheritance configurations, it provided a general computational method capable of resolving the corresponding class of cases. The underlying idea is therefore the complete computational treatment of a delimited jurisprudential domain. To formalize Al-Khwārizmī's inheritance arithmetic
(\textit{'ilm al-farā'iḍ}) within mathematical logic, we represent its
algebraic principles by a fixed-dimensional theory of linear rational
arithmetic over the field of rational numbers $\mathbb{Q}$ [23]. For a fixed
inheritance configuration, the relevant variables represent the shares
of the distinct claimant categories, while the legal constraints are
expressed as rational linear equalities and inequalities.

\subsubsection{Fourier--Motzkin Elimination on a Historical Bequest}

Al-Khwārizmī's \textit{Kitāb al-Waṣāyā} considers cases in which
third-party legacy bequests (\textit{waṣiyyah}) interact with the
standard inheritance shares. Islamic jurisprudence imposes the
constraint that an unapproved bequest cannot exceed one third of the
estate.

Consider a classical problem in which a man dies leaving two sons and
wills to a stranger a legacy equal to one third of the estate minus one
fourth of what one son receives. Normalize the gross estate to
$G=1$. Let $x\in\mathbb{Q}$ denote the share of one son and let
$w\in\mathbb{Q}$ denote the value of the bequest. The system is

\begin{align}
2x+w &= 1
&&\text{(Estate Conservation)},\\
w &= \frac{1}{3}-\frac{1}{4}x
&&\text{(Bequest Definition)},\\
w &\leq \frac{1}{3}
&&\text{(Legal Upper Bound)}.
\end{align}

The equality constraints may be represented as pairs of linear
inequalities. In particular, the system contains the upper and lower
bounds

\begin{equation}
\begin{aligned}
w &\leq 1-2x,
&
w &\geq 1-2x,\\
w &\leq \frac{1}{3}-\frac14x,
&
w &\geq \frac{1}{3}-\frac14x,\\
w &\leq \frac13.
\end{aligned}
\end{equation}

Fourier--Motzkin elimination eliminates $w$ by combining each lower
bound with each upper bound. One of the generated Fourier--Motzkin constraints is

\begin{equation}
1-2x
\leq
\frac13-\frac14x.
\end{equation}

Applying the algebraic operations of \textit{al-jabr} and
\textit{al-muqābala} gives

\begin{equation}
1-\frac13
\leq
2x-\frac14x,
\end{equation}
and hence

\begin{equation}
\frac23
\leq
\frac74x.
\end{equation}

The equality constraints of the original system then determine

\begin{equation}
x=\frac{8}{21}.
\end{equation}

Substitution gives

\begin{equation}
w
=
\frac13-\frac14\left(\frac{8}{21}\right)
=
\frac{5}{21}.
\end{equation}

Thus each son receives $\frac{8}{21}$ of the estate and the stranger
receives $\frac{5}{21}$. The legal constraint is satisfied since

\begin{equation}
\frac{5}{21}\leq\frac13.
\end{equation}

The example illustrates that the inheritance calculation can be
represented as a finite system of rational linear constraints and
resolved by a terminating elimination procedure.

\subsubsection{Model-Theoretic Completeness}

The algebraic representation also admits a model-theoretic
characterization [24]. For a fixed inheritance configuration, the relevant
structure is not an arbitrary finite-dimensional ordered vector space:
the claimant coordinates and their order are part of the specified
formal structure.

The mathematical development below establishes three distinct formal
properties: model uniqueness, logical completeness of the complete
theory of the intended model, and algorithmic decidability of the
corresponding rational linear constraint systems.

First, the proof of model uniqueness establishes that every admissible
fixed-dimensional inheritance model is structurally isomorphic to the
rational coordinate space $\mathbb{Q}^d$. The coordinate map
$\psi:\mathcal{M}\to\mathbb{Q}^d$ is defined by the unique rational
coordinates of each element with respect to the distinguished basis
$e_1,\ldots,e_d$. Because $\psi$ preserves addition, rational scalar
multiplication, the distinguished basis vectors, and coordinatewise
ordering, it is an isomorphism.

Second, the proof of algorithmic decidability demonstrates that a
computational engine can always resolve an inheritance scenario in
finite time. This is achieved by mapping all legal constraints to finite
systems of linear rational equalities and inequalities. Such systems are
decidable via quantifier elimination procedures, specifically
Fourier--Motzkin elimination [25]. Because an inheritance configuration
contains a finite number of variables, the algorithm systematically
projects out variables one by one and is guaranteed to terminate after
finitely many steps, determining whether the system is satisfiable and,
when the constraints determine a unique allocation, computing that
allocation exactly.

Third, the final synthesis establishes what we term \emph{domain-relative computational completeness} for the broader Fiqh framework. This is logically constructed by combining the exhaustive case generation of the \textit{al-\c{s}abr wa-al-taq\c{s}\={\i}m} procedure with the terminating properties of function-free Horn-clause Datalog [8]. By satisfying Proposition 2 (Complete Legislation) and Proposition 1 (Consistent Legislation) from prior work [1], the framework ensures that the jurisprudential question space is exhaustively covered by reasoned rulings. The inductive procedure generates a complete axiom set $A_{CC}$ for a closed chapter, which the deductive engine evaluates to its least fixed point. The completeness result for the fixed inheritance theory should not be interpreted as a general completeness claim for arbitrary Fiqh theories. For a closed Fiqh chapter $CC$, completeness of the axiom set $A_{CC}$ with respect to its specified finite case space, together with completeness of the Horn deductive engine for the specified rule class, ensures that every ruling represented in $A_{CC}$ is derivable by the engine.

\subsubsection{Formal Language of Inheritance Equations}

For a fixed inheritance configuration with $d$ claimant categories, let
$\mathbb{Q}$ be the fixed coefficient field. We use a one-sorted
first-order language whose domain is the space of inheritance
allocations. The language is

\[
\mathcal{L}_{d}
=
\left\langle
0,+,\leq,\{f_q\}_{q\in\mathbb{Q}},
e_1,\ldots,e_d
\right\rangle,
\]

where $0$ is the zero allocation, $+$ is vector addition, $\leq$ is the
coordinatewise order, $f_q$ is the unary operation corresponding to
multiplication by the fixed rational scalar $q$, and
$e_1,\ldots,e_d$ are distinguished basis vectors corresponding to the
$d$ claimant coordinates.

The intended structure is

\[
\mathcal{V}_d
=
\left(
\mathbb{Q}^d,
0,+,\leq_{\mathrm{cw}},
\{f_q\}_{q\in\mathbb{Q}},
e_1,\ldots,e_d
\right),
\]

where

\[
e_i=(0,\ldots,0,1,0,\ldots,0)
\]

and

\[
(x_1,\ldots,x_d)
\leq_{\mathrm{cw}}
(y_1,\ldots,y_d)
\iff
\bigwedge_{i=1}^{d}x_i\leq y_i.
\]

The admissible fixed-dimensional inheritance models are required to
satisfy the following two coordinate conditions. First, every element
has a representation over the fixed coefficient field $\mathbb{Q}$:

\[
\forall x\;
\exists q_1,\ldots,q_d\in\mathbb{Q}
\quad
x=q_1e_1+\cdots+q_de_d.
\]

Second, the distinguished vectors are linearly independent:

\[
q_1e_1+\cdots+q_de_d=0
\quad\Longrightarrow\quad
q_1=\cdots=q_d=0,
\qquad
q_1,\ldots,q_d\in\mathbb{Q}.
\]

These two conditions define the admissible fixed-dimensional
inheritance models considered below. They are structural conditions
over the fixed coefficient field $\mathbb{Q}$ and are not additional
first-order quantification over the inheritance domain.

\subsubsection{Uniqueness of the Fixed-Dimensional Inheritance Model}

\textbf{Uniqueness of the Fixed-Dimensional Inheritance Model}
Let $\mathcal{M}$ be a structure in the language $\mathcal{L}_d$
satisfying the vector-space axioms over the fixed coefficient field
$\mathbb{Q}$, the coordinatewise-order axioms, and the two fixed
coordinate conditions stated above. Then

\[
\mathcal{M}\cong\mathcal{V}_d=\mathbb{Q}^d.
\]

In particular, the structure is uniquely determined up to isomorphism
by its fixed coefficient field and its $d$ distinguished basis
coordinates.

\begin{proof}
Let $M$ be the domain of $\mathcal{M}$. By the spanning condition, for
every $x\in M$ there exist rational numbers
$q_1,\ldots,q_d\in\mathbb{Q}$ such that

\[
x=q_1e_1+\cdots+q_de_d.
\]

By linear independence of
$e_1,\ldots,e_d$, this representation is unique. Hence we may define
the coordinate map

\[
\phi:M\longrightarrow\mathbb{Q}^d
\]

by

\[
\phi(x)=(q_1,\ldots,q_d),
\]

where $(q_1,\ldots,q_d)$ is the unique rational coordinate tuple of
$x$.

The map $\phi$ is surjective because, for every
$(q_1,\ldots,q_d)\in\mathbb{Q}^d$, the element

\[
q_1e_1+\cdots+q_de_d
\]

belongs to $M$ and is mapped by $\phi$ to
$(q_1,\ldots,q_d)$. It is injective because two elements having the same
coordinate tuple have the same representation in the basis
$e_1,\ldots,e_d$ and are therefore equal.

It remains to show that $\phi$ preserves the structure. If
$x,y\in M$ have coordinate representations

\[
x=\sum_{i=1}^{d}q_i e_i,
\qquad
y=\sum_{i=1}^{d}r_i e_i,
\]

then

\[
x+y
=
\sum_{i=1}^{d}(q_i+r_i)e_i,
\]

and hence

\[
\phi(x+y)
=
(q_1+r_1,\ldots,q_d+r_d)
=
\phi(x)+\phi(y).
\]

Similarly, for every $q\in\mathbb{Q}$,

\[
qx
=
\sum_{i=1}^{d}(qq_i)e_i,
\]

so

\[
\phi(f_q(x))
=
q\phi(x).
\]

Finally, by the coordinatewise-order axiom,

\[
x\leq y
\iff
q_i\leq r_i
\quad\text{for every }i,
\]

which is equivalent to

\[
\phi(x)\leq_{\mathrm{cw}}\phi(y).
\]

Moreover, $\phi(e_i)$ is the $i$th standard basis vector of
$\mathbb{Q}^d$ for each $i=1,\ldots,d$. Thus $\phi$ is a bijection preserving zero, addition, rational scalar
multiplication, the distinguished basis vectors, and the order
relation. Therefore

\[
\mathcal{M}\cong\mathcal{V}_d=\mathbb{Q}^d.
\]

This proves the theorem.
\end{proof}

\subsubsection{The Complete Theory of the Fixed-Dimensional Model}

Let
\[
T_d=\operatorname{Th}(\mathcal{V}_d)
\]
denote the first-order theory of the intended fixed-dimensional
inheritance structure $\mathcal{V}_d$.

By definition, for every sentence $\varphi$ of the inheritance
language,
\[
\varphi\in T_d
\iff
\mathcal{V}_d\models\varphi.
\]

For every sentence $\varphi$, classical semantics gives
\[
\mathcal{V}_d\models\varphi
\qquad\text{or}\qquad
\mathcal{V}_d\models\neg\varphi.
\]

Hence
\[
T_d\models\varphi
\qquad\text{or}\qquad
T_d\models\neg\varphi.
\]

By the completeness theorem for first-order logic,
\[
T_d\vdash\varphi
\qquad\text{or}\qquad
T_d\vdash\neg\varphi.
\]

Thus $T_d$ is complete.

\subsubsection{Algorithmic Decidability}

\textbf{Decidability of the Inheritance Theorem}
\label{thm:inheritance-decidability}
For every fixed inheritance configuration, the corresponding system of
rational linear constraints is decidable.

\begin{proof}
Every inheritance condition in the fixed configuration is represented
by a finite system of linear equalities and inequalities over
$\mathbb{Q}$. Such systems admit quantifier elimination and
satisfiability procedures for linear rational arithmetic. Fourier--Motzkin elimination determines the resulting rational
solution set, which may be empty, a singleton, or contain multiple
solutions. In the singleton case, the unique rational allocation is
thereby determined exactly.

Consequently, the procedure determines whether the resulting system is
satisfiable and, whenever the constraints determine a unique
allocation, computes its exact rational solution.
\end{proof}

Therefore, the inheritance framework has two distinct formal
properties:
\[
\text{logical completeness}
\qquad\text{and}\qquad
\text{algorithmic decidability}.
\]

The first follows from the definition of
$T_d=\operatorname{Th}(\mathcal{V}_d)$ together with the completeness
theorem for first-order logic, while the second follows from the
decidable nature of linear rational arithmetic.

\subsubsection{Relation to the Complete Formalization of a Fiqh Subchapter}

The significance of this result for the present framework is
methodological. Al-Khwārizmī's treatment of inheritance does not merely provide isolated numerical examples. For a specified inheritance configuration, the relevant legal cases are represented within a finite algebraic structure and solved by a general computational procedure.

The present Fiqh framework adopts the same general objective at a
different level of formal abstraction. For a closed Fiqh chapter
$CC$, relative to its specified ontology, the \textit{al-sabr
wa-al-taqsīm} procedure exhaustively generates the relevant case space. The resulting Horn Fiqh axiom set $A_{CC}$ is therefore complete with respect to that specified finite space of cases.

This connects directly with the two propositions established in the
2016 paper as already discussed in section P.2.1.

Proposition~2 provides completeness of the legislation over the
specified question space. The exhaustive \textit{al-sabr
wa-al-taqsīm} construction provides completeness of the corresponding case space and hence of $A_{CC}$. Proposition~1 (in its re-formulation in this work) ensures that the resulting rulings are associated with an explicit Fiqh-rule rather than being introduced as unexplained exceptions.

If the Horn deductive engine is complete for the rule class specified by the framework, then every ruling encoded by the complete axiom set $A_{CC}$ is derivable by the engine. Thus, for every admissible question $M$ in the closed chapter, there exists a derivable ruling $R$ such that

\[
M\in SF_{CC}
\quad\Longrightarrow\quad
\exists R\;
\bigl(
\operatorname{RulingOf}(M,R)
\land
A_{CC}\vdash_{\mathrm{Horn}}R
\bigr).
\]

Accordingly, the combination of exhaustive case generation, complete
axiomatization, and complete deduction establishes a
\emph{domain-relative computational completeness} for the specified
closed Fiqh subchapter. The claim is not that all of Fiqh, or all
mathematics, has thereby been formalized. Rather, the claim is that a
delimited Fiqh domain can be completely represented and computationally
resolved relative to its specified ontology, question space, axioms,
and rule engine.

This is analogous in methodological spirit to Al-Khwārizmī's treatment
of inheritance. In both cases, the objective is to provide a complete
computational method for an entire delimited class of jurisprudential
problems rather than to formalize isolated examples. The formal
machinery, however, is different: Al-Khwārizmī's inheritance method is
expressed through algebraic equations over rational quantities, whereas
the present framework uses exhaustive case generation,
Horn-clause rules, and a terminating deductive engine.

The relation to Gödel's completeness theorem is likewise distinct.
Gödel's theorem concerns the equivalence between semantic validity and
formal provability for first-order logic. The completeness established
here is instead relative to a specified closed Fiqh domain: the
relevant cases are exhaustively generated, the resulting axioms are
complete for that case space, and the deductive engine is complete for
those axioms. It is therefore a \emph{domain-relative computational
completeness}, not a claim of universal logical completeness.

The resulting hierarchy of completeness is consequently:

\begin{equation}
\begin{aligned}
\text{Complete question space}
&\longrightarrow
\text{Complete axiom set }A_{CC}\\
&\longrightarrow
\text{Complete Horn deduction}\\
&\longrightarrow
\text{Complete computational model of }CC.
\end{aligned}
\end{equation}

In this precise sense, the framework seeks to achieve for a closed
Fiqh subchapter what Al-Khwārizmī achieved for the inheritance problems
within his algebraic treatment: a complete, finite, and computationally
executable formalization of a delimited jurisprudential domain.

\section{End-to-End Execution Example}

The preceding sections established the theoretical foundations and the
formal realization of the proposed computational system. This section
demonstrates the complete execution of the system on a single
jurisprudential query, tracing the input through Ta\c{s}r\={i}f,
Specialized I'r\={a}b, semantic grounding and ARAC translation, and
finally the Datalog deductive engine.

The example is deliberately chosen so that the syntactic stage
illustrates the attachment mechanism: morphological
information supplied by the input is matched against the realization
requirements of Fiqh-grammar slots, while jurisprudential roles are
obtained from the matched slots themselves.

\subsection*{1. Original Jurisprudential Query}

Consider the following fully vocalized Classical Arabic query submitted
by a user:

\begin{center}
\textarabic{أَنَا مَرِيضٌ وَأَحْدَثْتُ وَدَخَلَ وَقْتُ الْعَصْرِ وَلَيْسَ عِنْدِي مَاءٌ وَلَكِنْ عِنْدِي تُرَابٌ طَاهِرٌ هَلْ يَجُوزُ التَّيَمُّمُ؟}
\end{center}

Informally, the user states:

\begin{quote}
``I am sick, I incurred impurity, the time of Asr has entered, I do not
have water, but I have pure dust. Is Tayammum permissible?''
\end{quote}

The query contains five factual components followed by a legal query:

\begin{enumerate}
    \item the speaker is sick;
    \item the speaker has incurred minor ritual impurity;
    \item the time of Asr has entered;
    \item water is unavailable;
    \item clean earth is available;
    \item the user asks whether Tayammum is permissible.
\end{enumerate}

\subsection*{2. Phase 1: Ta\c{s}r\={i}f (Morphological Recognition)}

The system first processes the input sentence word-by-word using the
Ta\c{s}r\={i}f procedure. The result is the annotated morphological
sequence
\[
\mu^*(S)=\langle m_1,\ldots,m_n\rangle,
\]
where each morphological object has the form
\[
m_i=(r_i,p_i,c_i,f_i,\delta_i).
\]

For the constituents relevant to the subsequent syntactic and semantic
analysis, the morphological analysis supplies information of the
following form:

\begin{itemize}

\item \textbf{Clause 1:}
\textarabic{مَرِيضٌ}

\[
\left\langle
\text{\textarabic{م-ر-ض}},
\text{Adjective},
\ldots,
\left\langle
\mathrm{Raf'},\mathrm{\dhammah},\mathrm{Explicit}
\right\rangle
\right\rangle
\]

\item \textbf{Clause 2:}
\textarabic{أَحْدَثْتُ}

\[
\left\langle
\text{\textarabic{ح-د-ث}},
\text{Verb},
\ldots,
\left\langle
\mathrm{None},\mathrm{Sukun},\mathrm{Explicit}
\right\rangle
\right\rangle
\]

\item \textbf{Clause 3:}
\textarabic{دَخَلَ وَقْتُ الْعَصْرِ}

\[
\begin{aligned}
\text{\textarabic{دَخَلَ}}
&:
\left\langle
\text{\textarabic{د-خ-ل}},
\text{Verb},
\ldots,
\left\langle\mathrm{None},\ldots\right\rangle
\right\rangle,
\\
\text{\textarabic{وَقْتُ}}
&:
\left\langle
\text{\textarabic{و-ق-ت}},
\text{Noun},
\ldots,
\left\langle
\mathrm{Raf'},\mathrm{\dhammah},\mathrm{Explicit}
\right\rangle
\right\rangle,
\\
\text{\textarabic{الْعَصْرِ}}
&:
\left\langle
\text{\textarabic{ع-ص-ر}},
\text{Noun},
\ldots,
\left\langle
\mathrm{Jarr},\mathrm{kasrah},\mathrm{Explicit}
\right\rangle
\right\rangle.
\end{aligned}
\]

\item \textbf{Clause 4:}
\textarabic{لَيْسَ عِنْدِي مَاءٌ}

\[
\begin{aligned}
\text{\textarabic{لَيْسَ}}
&:
\left\langle
\text{\textarabic{ل-ي-س}},
\text{DefectiveVerb},
\ldots,
\left\langle\mathrm{None},\ldots\right\rangle
\right\rangle,
\\
\text{\textarabic{مَاءٌ}}
&:
\left\langle
\text{\textarabic{م-و-ه}},
\text{Noun},
\ldots,
\left\langle
\mathrm{Raf'},\mathrm{\dhammah},\mathrm{Explicit}
\right\rangle
\right\rangle.
\end{aligned}
\]

\item \textbf{Clause 5:}
\textarabic{تُرَابٌ طَاهِرٌ}

\[
\begin{aligned}
\text{\textarabic{تُرَابٌ}}
&:
\left\langle
\text{\textarabic{ت-ر-ب}},
\text{Noun},
\ldots,
\left\langle
\mathrm{Raf'},\mathrm{\dhammah},\mathrm{Explicit}
\right\rangle
\right\rangle,
\\
\text{\textarabic{طَاهِرٌ}}
&:
\left\langle
\text{\textarabic{ط-ه-ر}},
\text{Adjective},
\ldots,
\left\langle
\mathrm{Raf'},\mathrm{\dhammah},\mathrm{Explicit}
\right\rangle
\right\rangle.
\end{aligned}
\]

\item \textbf{Clause 6:}
\textarabic{هَلْ يَجُوزُ التَّيَمُّمُ؟}

\[
\begin{aligned}
\text{\textarabic{يَجُوزُ}}
&:
\left\langle
\text{\textarabic{ج-و-ز}},
\text{Verb},
\ldots,
\left\langle
\mathrm{Raf'},\mathrm{\dhammah},\mathrm{Explicit}
\right\rangle
\right\rangle,
\\
\text{\textarabic{التَّيَمُّمُ}}
&:
\left\langle
\text{\textarabic{ي-م-م}},
\text{VerbalNoun},
\ldots,
\left\langle
\mathrm{Raf'},\mathrm{\dhammah},\mathrm{Explicit}
\right\rangle
\right\rangle.
\end{aligned}
\]

\end{itemize}

At this stage, the morphological analysis supplies morphological types,
features, roots, and inflectional information. It does \emph{not}
itself assign jurisprudential roles such as Agent, Patient, Resource,
or Action.

\subsection*{3. Phase 2: I\text{'}r\={a}b (Syntactic Recognition via Fiqh Grammar)}

The parser receives $\mu^*(S)$ and applies the Fiqh-Categorized
Attribute Grammar defined in Section~3.

The essential point is that the parser does not infer a jurisprudential
role directly from an input word or from its $\delta$ marker. Instead,
it matches the morphological realization supplied by the input against
the realization required by an open grammar slot. The jurisprudential
role is then obtained from the annotation of the matched slot.

For completeness, the relevant fragment of the Fiqh Grammar used in
the present example is displayed before the individual inferences are
performed. This permits the reader to compare the operational example
directly with the formal grammar introduced earlier.

\paragraph{Coordination and higher-level query structure.}

The complete input is organized by a higher-level legal-condition
construction:

\[
\begin{aligned}
S
&\rightarrow
\mathit{LegalCondition}\;\mathit{LegalQuery},
\\
\mathit{LegalCondition}
&\rightarrow
[
\mathit{Cond}_1,\mathit{CoordPart},
\mathit{Cond}_2,\mathit{CoordPart},
\ldots,
\mathit{Cond}_5
].
\end{aligned}
\]

The coordination parts are realized by the Arabic conjunctions
\textarabic{و} and \textarabic{وَلَكِنْ}. The individual conditions are
instantiated by the following productions.

\paragraph{Legal-state production.}

\[
\mathit{LegalState}
\rightarrow
[
\mathit{Pronoun},
\mathit{StatePredicate}
].
\]

The relevant slot is specified by
\[
\mathit{StatePredicate}
=
\left\langle
\mathrm{Adjective},
\mathrm{Raf'},
\mathrm{HealthState}
\right\rangle.
\]

Thus the realization
\[
\left\langle
\mathrm{Adjective},
\mathrm{Raf'}
\right\rangle
\]
is the matching information supplied by the input, whereas
$\mathrm{HealthState}$ is the interpretation stored in the grammar
slot.

\paragraph{Impurity-event production.}

The second condition is represented by the event structure
\[
\mathit{LegalEvent}
\rightarrow
[
\mathit{EventVerb},
\mathit{EventParticipant}
].
\]

The relevant event predicate is associated with the lexical root
\texttt{h-d-th}, while the participant interpretation is supplied by
the corresponding grammar specification. The resulting semantic
interpretation is the state transition represented later as
\texttt{minor\_impurity}.

\paragraph{Temporal-event production.}

\[
\mathit{LegalEvent}
\rightarrow
[
\mathit{EventVerb},
\mathit{EventTime},
\mathit{TemporalBound}
].
\]

The relevant slot specifications are
\[
\begin{aligned}
\mathit{EventTime}
&=
\left\langle
\mathrm{Noun},
\mathrm{Raf'},
\mathrm{TimeEntry}
\right\rangle,
\\
\mathit{TemporalBound}
&=
\left\langle
\mathrm{IdafaComplement},
\mathrm{Jarr'},
\mathrm{AsrTime}
\right\rangle.
\end{aligned}
\]

\paragraph{Negated-resource production.}

\[
\mathit{LegalNegation}
\rightarrow
[
\mathit{NegVerb},
\mathit{PossessiveFrame},
\mathit{Resource}
].
\]

The resource slot is specified as
\[
\mathit{Resource}
=
\left\langle
\mathrm{Noun},
\mathrm{Raf'},
\mathrm{Water}
\right\rangle.
\]

\paragraph{Available-resource production.}

\[
\mathit{LegalState}
\rightarrow
[
\mathit{PossessiveFrame},
\mathit{Resource},
\mathit{Modifier}
].
\]

The relevant slot specifications are
\[
\begin{aligned}
\mathit{Resource}
&=
\left\langle
\mathrm{Noun},
\mathrm{Raf'},
\mathrm{CleanEarth}
\right\rangle,
\\
\mathit{Modifier}
&=
\left\langle
\mathrm{Adjective},
\mathrm{Raf'},
\mathrm{PurityModifier}
\right\rangle.
\end{aligned}
\]

\paragraph{Legal-query production.}

\[
\mathit{LegalQuery}
\rightarrow
[
\mathit{IntPart},
\mathit{ModalVerb},
\mathit{Action}
].
\]

The relevant slots are
\[
\begin{aligned}
\mathit{ModalVerb}
&=
\left\langle
\mathrm{Verb},
\mathrm{Raf'},
\mathrm{PermissionPredicate}
\right\rangle,
\\
\mathit{Action}
&=
\left\langle
\mathrm{VerbalNoun},
\mathrm{Raf'},
\mathrm{TayammumAction}
\right\rangle.
\end{aligned}
\]

\paragraph{Slot matching.}

Consider, for example, the constituent
\textarabic{مَرِيضٌ}. Its input realization is
\[
\kappa(\text{\textarabic{مَرِيضٌ}})
=
\left\langle
\mathrm{Adjective},
\mathrm{Raf'}
\right\rangle.
\]

The parser does not conclude from this key alone that the word means
$\mathrm{HealthState}$. Rather, it compares the key with the
requirements of the open slots of the applicable production. When it
matches
\[
\left\langle
\mathrm{Adjective},
\mathrm{Raf'}
\right\rangle
=
\kappa_R(\mathit{StatePredicate}),
\]
the constituent is assigned to the
$\mathit{StatePredicate}$ slot. The role or semantic interpretation
is then obtained from the slot annotation.

The same mechanism applies to the nominal constituents. For example,
\textarabic{مَاءٌ} supplies
\[
\kappa(\text{\textarabic{مَاءٌ}})
=
\left\langle
\mathrm{Noun},
\mathrm{Raf'}
\right\rangle.
\]

The grammar determines which open resource slot this realization can
fill, and the interpretation associated with that slot is then
obtained from the grammar. Likewise,
\textarabic{تُرَابٌ} supplies the same basic morphological realization:
\[
\kappa(\text{\textarabic{تُرَابٌ}})
=
\left\langle
\mathrm{Noun},
\mathrm{Raf'}
\right\rangle.
\]

The distinction between the water resource and the clean-earth
resource is therefore not obtained by treating the lexical root as a
jurisprudential role. It is obtained from the applicable production
and the corresponding slot specification. The adjectival constituent \textarabic{طَاهِرٌ} is matched to the modifier slot
\[
\mathit{Modifier}
=
\left\langle
\mathrm{Adjective},
\mathrm{Raf'},
\mathrm{PurityModifier}
\right\rangle,
\]
after which the feature-unification stage verifies its agreement with
\textarabic{تُرَابٌ}. The production rule used for Water does not include such an adjectival constituent, since Fiqh-contexts usually use "Water" to mean "pure Water".

For the query
\textarabic{هَلْ يَجُوزُ التَّيَمُّمُ؟},
the parser obtains
\[
\kappa(\text{\textarabic{يَجُوزُ}})
=
\left\langle
\mathrm{Verb},
\mathrm{Raf'}
\right\rangle
\] and \[
\kappa(\text{\textarabic{التَّيَمُّمُ}})
=
\left\langle
\mathrm{VerbalNoun},
\mathrm{Raf'}
\right\rangle.
\]

\begin{center}
\begin{minipage}{\linewidth}
\centering

\resizebox{\linewidth}{!}{%
\begin{tikzpicture}[
    node distance=1.0cm and 1.5cm,
    box/.style={
        draw,
        rounded corners,
        align=center,
        text width=2.9cm,
        minimum height=8mm,
        inner sep=3pt,
        font=\footnotesize,
        fill=blue!5
    },
    every edge/.style={
        draw,
        ->,
        >=Stealth,
        thick
    },
    every edge/.append style={
        font=\scriptsize
    }
]

\node[box] (c1) {Coord \textarabic{و}};
\node[box, right=of c1] (ana) {Pronoun \textarabic{أنا}};
\node[box, right=of ana] (marid) {StatePredicate\\
\textarabic{مَرِيضٌ}\\
$(\mathrm{Adjective},\mathrm{Raf'})$};
\draw (c1) edge node[above] {Atf} (ana);
\draw (ana) edge node[above] {State} (marid);

\node[box, below=1.1cm of c1] (c2) {Coord \textarabic{و}};
\node[box, right=of c2] (ahdath) {EventVerb\\
\textarabic{أَحْدَثْتُ}};
\draw (c1) edge node[left] {Atf} (c2);
\draw (c2) edge node[above] {Event} (ahdath);

\node[box, below=1.1cm of c2] (c3) {Coord \textarabic{و}};
\node[box, right=of c3] (dakhala) {EventVerb\\
\textarabic{دَخَلَ}};
\node[box, right=of dakhala] (waqt) {EventTime\\
\textarabic{وَقْتُ}\\
$(\mathrm{Noun},\mathrm{Raf'})$};
\node[box, right=of waqt] (asr) {TemporalBound\\
\textarabic{الْعَصْرِ}\\
$(\mathrm{Noun},\mathrm{Jarr'})$};
\draw (c2) edge node[left] {Atf} (c3);
\draw (c3) edge node[above] {Event} (dakhala);
\draw (dakhala) edge node[above] {TimeEntry} (waqt);
\draw (waqt) edge node[above] {Idafa} (asr);

\node[box, below=1.1cm of c3] (c4) {Coord \textarabic{و}};
\node[box, right=of c4] (laysa) {LegalNegation\\
\textarabic{لَيْسَ}};
\node[box, right=of laysa] (indi1) {PossessiveFrame\\
\textarabic{عِنْدِي}};
\node[box, right=of indi1] (ma) {Resource\\
\textarabic{مَاءٌ}\\
$(\mathrm{Noun},\mathrm{Raf'})$};
\draw (c3) edge node[left] {Atf} (c4);
\draw (c4) edge node[above] {Negation} (laysa);
\draw (laysa) edge node[above] {Possession} (indi1);
\draw (indi1) edge node[above] {Resource} (ma);

\node[box, below=1.1cm of c4] (c5) {Coord \textarabic{وَلَكِنْ}};
\node[box, right=of c5] (indi2) {PossessiveFrame\\
\textarabic{عِنْدِي}};
\node[box, right=of indi2] (turab) {Resource\\
\textarabic{تُرَابٌ}\\
$(\mathrm{Noun},\mathrm{Raf'})$};
\node[box, right=of turab] (tahir) {Modifier\\
\textarabic{طَاهِرٌ}\\
$(\mathrm{Adjective},\mathrm{Raf'})$};
\draw (c4) edge node[left] {Atf} (c5);
\draw (c5) edge node[above] {Possession} (indi2);
\draw (indi2) edge node[above] {Resource} (turab);
\draw (turab) edge node[above] {Modifier} (tahir);

\node[box, below=1.1cm of c5] (c6) {IntPart \textarabic{هَلْ}};
\node[box, right=of c6] (yajuz) {ModalVerb\\
\textarabic{يَجُوزُ}\\
$(\mathrm{Verb},\mathrm{Raf'})$};
\node[box, right=of yajuz] (tayammum) {Action\\
\textarabic{التَّيَمُّمُ}\\
$(\mathrm{VerbalNoun},\mathrm{Raf'})$};
\draw (c5) edge node[left] {Seq} (c6);
\draw (c6) edge node[above] {Query} (yajuz);
\draw (yajuz) edge node[above] {Action} (tayammum);

\end{tikzpicture}%
}

\vspace{2pt}

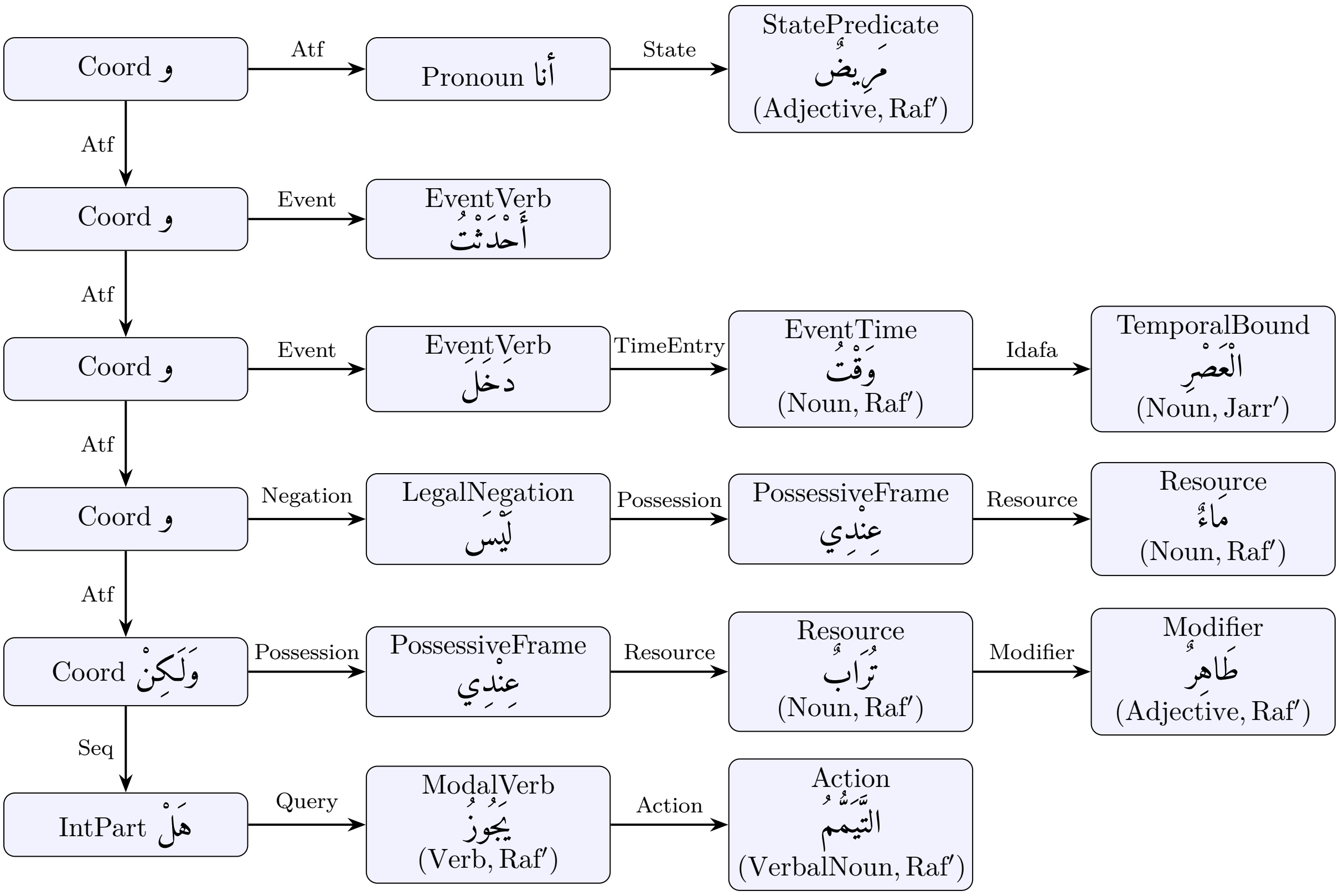
\textit{Figure 2: Visual Representation of the Syntactic Dependency
Graph $\mathcal{I}$}

\end{minipage}
\end{center}

The figure represents the result of the syntactic stage after slot
matching and feature validation. The labels such as
$\mathit{Resource}$, $\mathit{Action}$, and
$\mathit{StatePredicate}$ are therefore derived from grammar-slot
annotations rather than directly from the input words.

\subsection*{4. Phase 3: Semantic Grounding and ARAC Translation}

The validated syntactic graph $\mathcal{I}$ is now passed to the
semantic grounding component. At this stage, the system transforms
the syntactically recognized constituents into the domain-level
entities, states, resources, temporal facts, and actions that
constitute the input to ARAC.

This transformation is not performed by the I'r\={a}b procedure itself.
The I'r\={a}b procedure establishes the syntactic identity of each
constituent by matching its morphological realization against a
Fiqh-grammar slot. Semantic grounding subsequently combines this
syntactic information with the lexical knowledge associated with the
recognized roots and lexical entries.

The semantic grounding process therefore uses three sources of
information:

\begin{enumerate}
    \item the lexical root and morphological information supplied by
    Ta\c{s}r\={i}f;
    \item the semantic lexical entry or \emph{root-signature} associated
    with the recognized lexical item; and
    \item the jurisprudential role or interpretation supplied by the
    Fiqh-grammar slot to which the constituent was assigned.
\end{enumerate}

The distinction between these sources is essential. A root-signature
does not by itself determine the complete jurisprudential role of a
constituent, and an I'r\={a}b slot does not by itself supply the
lexical meaning of the word. The grounding component combines the two.

The resulting information flow can be represented as
\[
\boxed{
\begin{aligned}
\text{Morphological object}
&\longrightarrow
\text{Lexical entry/root-signature}\\
&\longrightarrow
\text{Grammar slot}
\longrightarrow
\text{Grounded semantic object}
\longrightarrow
\text{ARAC}.
\end{aligned}}
\]

More explicitly, for a recognized constituent $x$, let
$\operatorname{Lex}(x)$ denote its lexical entry and let
$\operatorname{Sig}(x)$ denote the semantic root-signature associated
with that entry. Let $\operatorname{Slot}(x)$ denote the Fiqh-grammar
slot obtained during I'r\={a}b. The grounding operation may then be
represented schematically as
\[
\operatorname{Ground}(x)
=
\Gamma\!\left(
\operatorname{Sig}(x),
\operatorname{Slot}(x),
\operatorname{Features}(x)
\right),
\]
where $\Gamma$ is the semantic grounding mapping defined by the
knowledge base of the relevant Fiqh chapter.

Thus, the root-signature supplies the lexical-semantic trigger, while
the grammar slot supplies the jurisprudential interpretation required
to determine how that trigger is instantiated in the current sentence.

\paragraph{4.1 Lexical Entries and Root-Signatures.}

For the present Tayammum example, the semantic lexicon contains entries
for the lexical items occurring in the query. Schematically, the
relevant entries may be represented as follows:

\begin{center}
\small
\begin{tabularx}{0.98\linewidth}{>{\raggedright\arraybackslash}X
                                  >{\raggedright\arraybackslash}X
                                  >{\raggedright\arraybackslash}X}
\textbf{Lexical item}
&
\textbf{Root}
&
\textbf{Root-signature}
\\ \hline
\textarabic{مَرِيضٌ}
&
\textarabic{م-ر-ض}
&
\texttt{SICKNESS}
\\
\textarabic{أَحْدَثْتُ}
&
\textarabic{ح-د-ث}
&
\texttt{RITUAL\_IMPURITY}
\\
\textarabic{دَخَلَ}
&
\textarabic{د-خ-ل}
&
\texttt{ENTRY}
\\
\textarabic{وَقْت}
&
\textarabic{و-ق-ت}
&
\texttt{TIME}
\\
\textarabic{الْعَصْر}
&
\textarabic{ع-ص-ر}
&
\texttt{ASR}
\\
\textarabic{مَاء}
&
\textarabic{م-و-ه}
&
\texttt{WATER}
\\
\textarabic{تُرَاب}
&
\textarabic{ت-ر-ب}
&
\texttt{EARTH}
\\
\textarabic{طَاهِر}
&
\textarabic{ط-ه-ر}
&
\texttt{PURITY}
\\
\textarabic{يَجُوز}
&
\textarabic{ج-و-ز}
&
\texttt{PERMISSION}
\\
\textarabic{التَّيَمُّم}
&
\textarabic{ي-م-م}
&
\texttt{TAYAMMUM}
\end{tabularx}
\normalsize
\end{center}

The table is schematic: the purpose here is to expose the operation of
the grounding layer rather than to reproduce the complete lexical
database. The important point is that the semantic category
\texttt{WATER}, for example, is obtained from the lexical knowledge
associated with \textarabic{مَاء}, whereas the fact that this lexical
item fills the jurisprudential \texttt{Resource} slot is obtained from
the Fiqh Grammar.

Consequently, the system does not use a rule of the form
\[
\texttt{Noun}+\mathrm{Raf'}
\longrightarrow
\texttt{Water}.
\]
Such a rule would incorrectly make the semantic interpretation a
property of the morphological realization.

Instead, the information is combined as follows:
\[
\begin{aligned}
\text{\textarabic{مَاءٌ}}
&\xrightarrow{\text{Ta\c{s}r\={i}f}}
\left(\text{\textarabic{م-و-ه}},\mathrm{Noun},\mathrm{Raf'}\right)
\\
&\xrightarrow{\text{Lexicon}}
\texttt{WATER}
\\
&\xrightarrow{\text{Fiqh slot}}
\texttt{Resource}
\\
&\xrightarrow{\text{Grounding}}
\texttt{Water resource}.
\end{aligned}
\]

The same mechanism applies to \textarabic{تُرَابٌ}. Its root-signature
is associated with the earth/material concept, while the Fiqh Grammar
identifies its occurrence as the resource relevant to the present
jurisprudential construction. The adjective \textarabic{طَاهِرٌ}
contributes the purity property through its own lexical entry and
through its attachment to the resource constituent.

\paragraph{4.2 Grounding the Individual Constituents.}

The transformation can now be followed constituent by constituent.

\medskip
\noindent\textbf{Clause 1: \textarabic{أَنَا مَرِيضٌ}.}

The Ta\c{s}r\={i}f layer has already established that
\textarabic{مَرِيضٌ} is an adjective with the realization
\[
\langle\mathrm{Adjective},\mathrm{Raf'}\rangle.
\]

The I'r\={a}b stage assigns it to the
$\mathit{StatePredicate}$ slot, whose jurisprudential interpretation is
$\mathrm{HealthState}$. Independently, the lexical entry for the root
\textarabic{م-ر-ض} supplies the root-signature
\[
\operatorname{Sig}(\text{\textarabic{مَرِيض}})
=
\texttt{SICKNESS}.
\]

The grounding component combines these pieces of information:
\[
\begin{aligned}
\text{\textarabic{مَرِيضٌ}}
&\xrightarrow{\text{Lexicon}}
\texttt{SICKNESS}
\\
&\xrightarrow{\text{Grammar slot }StatePredicate}
\texttt{HealthState}
\\
&\xrightarrow{\text{ARAC grounding}}
\texttt{sick}(p_1).
\end{aligned}
\]

The pronoun \textarabic{أنا}, recognized as the first-person speaker,
supplies the discourse individual $p_1$. The resulting grounded fact is
therefore
\[
\texttt{health\_state}(p_1,\texttt{sick}).
\]

\medskip
\noindent\textbf{Clause 2: \textarabic{وَأَحْدَثْتُ}.}

The lexical root \textarabic{ح-د-ث} is associated in the semantic lexicon with the
root-signature
\[
\operatorname{Sig}(\text{\textarabic{أَحْدَثْتُ}})
=
\texttt{RITUAL\_IMPURITY}.
\]

The verbal morphology identifies the constituent as an event-bearing
verb and the syntactic analysis attaches it to the relevant event slot.
The first-person morphology identifies the actor with the discourse
individual $p_1$.

The grounding component therefore instantiates the lexical
root-signature as a state of the current person:
\[
\text{\textarabic{أَحْدَثْتُ}}
\longrightarrow
\texttt{RITUAL\_IMPURITY}
\longrightarrow
\texttt{minor\_impurity}(p_1).
\]

The normalized ARAC fact is consequently
\[
\texttt{purification\_state}(p_1,\texttt{minor\_impurity}).
\]

The important point is that the root-signature identifies the semantic
phenomenon associated with the lexical item, whereas the grammatical
analysis determines how that phenomenon is related to the current
participant.

\medskip
\noindent\textbf{Clause 3: \textarabic{وَدَخَلَ وَقْتُ الْعَصْرِ}.}

The verb \textarabic{دَخَلَ} has the lexical root-signature
\[
\operatorname{Sig}(\text{\textarabic{دَخَلَ}})
=
\texttt{ENTRY}.
\]

The noun \textarabic{وَقْتُ} has the lexical signature
\[
\operatorname{Sig}(\text{\textarabic{وَقْت}})
=
\texttt{TIME},
\]
while \textarabic{الْعَصْرِ} has the lexical signature
\[
\operatorname{Sig}(\text{\textarabic{الْعَصْر}})
=
\texttt{ASR}.
\]

The I'r\={a}b stage has already established the dependency
\[
\text{\textarabic{دَخَلَ}}
\longrightarrow
\mathit{EventTime}
\longrightarrow
\text{\textarabic{وَقْتُ الْعَصْرِ}},
\]
with the Id\={a}fa relation connecting
\textarabic{وَقْتُ} and
\textarabic{الْعَصْرِ}.

The grounding component therefore composes the lexical signatures with
the syntactic structure:
\[
\texttt{TIME}
+
\texttt{ASR}
\longrightarrow
\texttt{asr\_time},
\]
and the entry signature of \textarabic{دَخَلَ} supplies the temporal
transition:
\[
\texttt{ENTRY}
+
\texttt{asr\_time}
\longrightarrow
\texttt{current\_time(asr)}.
\]

Thus the syntactic relation established by I'r\={a}b is not itself the
final semantic fact. It provides the structural information required
by the grounding component to combine the lexical signatures into the
ARAC-level temporal fact
\[
\texttt{current\_time(asr)}.
\]

\medskip
\noindent\textbf{Clause 4: \textarabic{وَلَيْسَ عِنْدِي مَاءٌ}.}

The lexical entry for \textarabic{مَاء} has root-signature
\[
\operatorname{Sig}(\text{\textarabic{مَاء}})
=
\texttt{WATER}.
\]

The I'r\={a}b stage assigns \textarabic{مَاءٌ} to the
$\mathit{Resource}$ slot of the negated-resource production. The
lexical and syntactic information are therefore combined as
\[
\begin{aligned}
\text{\textarabic{مَاءٌ}}
&\xrightarrow{\text{Lexicon}}
\texttt{WATER}
\\
&\xrightarrow{\text{Resource slot}}
\texttt{Water}
\\
&\xrightarrow{\text{\textarabic{لَيْسَ} + PossessiveFrame}}
\texttt{Unavailable(Water)}.
\end{aligned}
\]

Hence the grounded ARAC fact is
\[
\texttt{unavailable(water)}.
\]

Notice that the noun's Raf' realization does not produce the semantic
category \texttt{Water}. That category comes from the lexical
root-signature. The grammar determines that the lexical entity is
functioning as the relevant resource in this construction, and the
negation/possession structure determines its availability status.

\medskip
\noindent\textbf{Clause 5:
\textarabic{وَلَكِنْ عِنْدِي تُرَابٌ طَاهِرٌ}.}

The lexical entry for \textarabic{تُرَاب} supplies the root-signature
\[
\operatorname{Sig}(\text{\textarabic{تُرَاب}})
=
\texttt{EARTH},
\]
while the entry for \textarabic{طَاهِر} supplies
\[
\operatorname{Sig}(\text{\textarabic{طَاهِر}})
=
\texttt{PURITY}.
\]

The Fiqh Grammar has already established the syntactic relation
\[
\text{\textarabic{تُرَابٌ}}
\longrightarrow
\mathit{Resource}
\longleftarrow
\text{\textarabic{طَاهِرٌ}}
\]
through the corresponding resource and modifier slots.

The semantic grounding component therefore combines the two lexical
signatures with the grammatical modification relation:
\[
\texttt{EARTH}
+
\texttt{PURITY}
\longrightarrow
\texttt{CLEAN\_EARTH}.
\]

The possessive structure headed by \textarabic{عِنْدِي} then establishes
availability to the current person:
\[
\texttt{CLEAN\_EARTH}
+
\texttt{Possession}(p_1)
\longrightarrow
\texttt{available(clean\_earth)}.
\]

Thus the normalized fact is
\[
\texttt{available(clean\_earth)}.
\]

This step illustrates particularly clearly why semantic grounding
cannot be reduced either to morphology or to syntactic slot matching.
The lexical signature of \textarabic{تُرَاب} supplies the material
concept, the lexical signature of \textarabic{طَاهِر} supplies the
purity property, and the syntactic graph establishes that the latter
modifies the former. Their composition produces the domain-level
resource \texttt{clean\_earth}.

\medskip
\noindent\textbf{Clause 6:
\textarabic{هَلْ يَجُوزُ التَّيَمُّمُ؟}.}

The lexical entry for \textarabic{يَجُوزُ} has the root-signature
\[
\operatorname{Sig}(\text{\textarabic{يَجُوز}})
=
\texttt{PERMISSION},
\]
while the entry for \textarabic{التَّيَمُّم} has
\[
\operatorname{Sig}(\text{\textarabic{التَّيَمُّم}})
=
\texttt{TAYAMMUM}.
\]

The I'r\={a}b stage has already assigned these constituents to the
$\mathit{ModalVerb}$ and $\mathit{Action}$ slots respectively.
Consequently, the grounding component constructs the semantic query
\[
\texttt{Permissible(Tayammum)}.
\]

The interrogative particle \textarabic{هَلْ} marks the resulting
proposition as a query rather than as an asserted fact. Thus it is not
added to the factual ARAC state; instead, it determines the target
predicate submitted to the deductive engine.

\paragraph{4.3 From the Grounded Constituents to ARAC.}

The preceding steps can be summarized by the following grounding table.

\begin{center}
\fontsize{9pt}{10pt}\selectfont
\begin{tabularx}{0.98\linewidth}{>{\raggedright\arraybackslash}X
                                  >{\raggedright\arraybackslash}X
                                  >{\raggedright\arraybackslash}X
                                  >{\raggedright\arraybackslash}X}
\textbf{Input}
&
\textbf{Root-signature}
&
\textbf{Grammar interpretation}
&
\textbf{Grounded result}
\\ \hline
\textarabic{مَرِيضٌ}
&
\texttt{SICKNESS}
&
\texttt{HealthState}
&
\texttt{sick}(p\_1)
\\
\textarabic{أَحْدَثْتُ}
&
\texttt{RITUAL\_IMPURITY}
&
\texttt{Event}
&
\texttt{minor\_impurity}(p\_1)
\\
\textarabic{وَقْتُ الْعَصْرِ}
&
\texttt{TIME}+\texttt{ASR}
&
\texttt{EventTime}
&
\texttt{current\_time(asr)}
\\
\textarabic{مَاءٌ}
&
\texttt{WATER}
&
\texttt{Resource}
&
\texttt{water}
\\
\textarabic{لَيْسَ ... مَاءٌ}
&
\texttt{NEGATION}+\texttt{WATER}
&
\texttt{LegalNegation}
&
\texttt{unavailable(water)}
\\
\textarabic{تُرَابٌ طَاهِرٌ}
&
\texttt{EARTH}+\texttt{PURITY}
&
\texttt{Resource}+\texttt{Modifier}
&
\texttt{clean\_earth}
\\
\textarabic{عِنْدِي تُرَابٌ طَاهِرٌ}
&
\texttt{POSSESSION}+\texttt{CLEAN\_EARTH}
&
\texttt{PossessiveFrame}
&
\texttt{available(clean\_earth)}
\\
\textarabic{يَجُوزُ التَّيَمُّمُ}
&
\texttt{PERMISSION}+\texttt{TAYAMMUM}
&
\texttt{LegalQuery}
&
\texttt{permissible(tayammum)}
\end{tabularx}
\normalsize
\end{center}

This table makes explicit the division of responsibility between the
three relevant knowledge sources. Ta\c{s}r\={i}f supplies the root and
morphological realization; the semantic lexicon maps recognized lexical
items to root-signatures; and the Fiqh Grammar supplies the
jurisprudential interpretation of their syntactic attachment. The
grounding component composes these results into the domain predicates
used by ARAC.

The resulting normalized ARAC facts are:

\begin{verbatim}
person(p1).
health_state(p1, sick).
purification_state(p1, minor_impurity).
current_time(asr).
resource(water).
unavailable(water).
resource(clean_earth).
available(clean_earth).
\end{verbatim}

The target query is represented separately as

\begin{verbatim}
permissible(tayammum).
\end{verbatim}

The distinction between facts and query is important: the factual
sentences establish the current legal state, while the interrogative
sentence specifies the proposition whose derivability is to be tested.

The transformation from the syntactic graph to the ARAC graph can now
be visualized directly. Each node in the semantic graph corresponds to
a grounded domain entity, state, or temporal object, while each edge
corresponds to a semantic relation obtained by interpreting a
syntactic dependency together with the relevant lexical
root-signatures.


\begin{center}
\resizebox{\linewidth}{!}{%
\begin{tikzpicture}[
    node distance=1.8cm and 2cm,
    box/.style={
        draw,
        circle,
        align=center,
        minimum size=14mm,
        fill=green!5
    },
    edgelabel/.style={
        font=\footnotesize,
        inner sep=2pt
    },
    statuslabel/.style={
        font=\footnotesize,
        inner sep=2pt
    },
    every edge/.style={
        draw,
        ->,
        >=Stealth,
        thick
    }
]

\node[box] (p1) {Person ($p_1$)};
\node[box, below left=of p1] (hs1) {Sick ($hs_1$)};
\node[box, below=of p1] (ps1) {Impurity ($ps_1$)};
\node[box, below right=of p1] (t1) {Time Asr ($t_1$)};
\node[box, right=2.5cm of p1] (r1) {Water ($r_1$)};
\node[box, right=2.5cm of t1] (r2) {Earth ($r_2$)};

\draw (p1) edge node[left, edgelabel] {Health} (hs1);
\draw (p1) edge node[right, edgelabel] {Purification} (ps1);
\draw (p1) edge node[above, edgelabel] {Context} (t1);
\draw (p1) edge node[above, edgelabel] {Resource} (r1);
\draw (p1) edge node[above, pos=0.7, edgelabel] {Resource} (r2);

\node[
    draw,
    dashed,
    fit=(r1),
    inner sep=2pt,
    label={[statuslabel]below:Unavailable}
] {};

\node[
    draw,
    dashed,
    fit=(r2),
    inner sep=2pt,
    label={[statuslabel]below:Available}
] {};

\end{tikzpicture}%
}
\end{center}

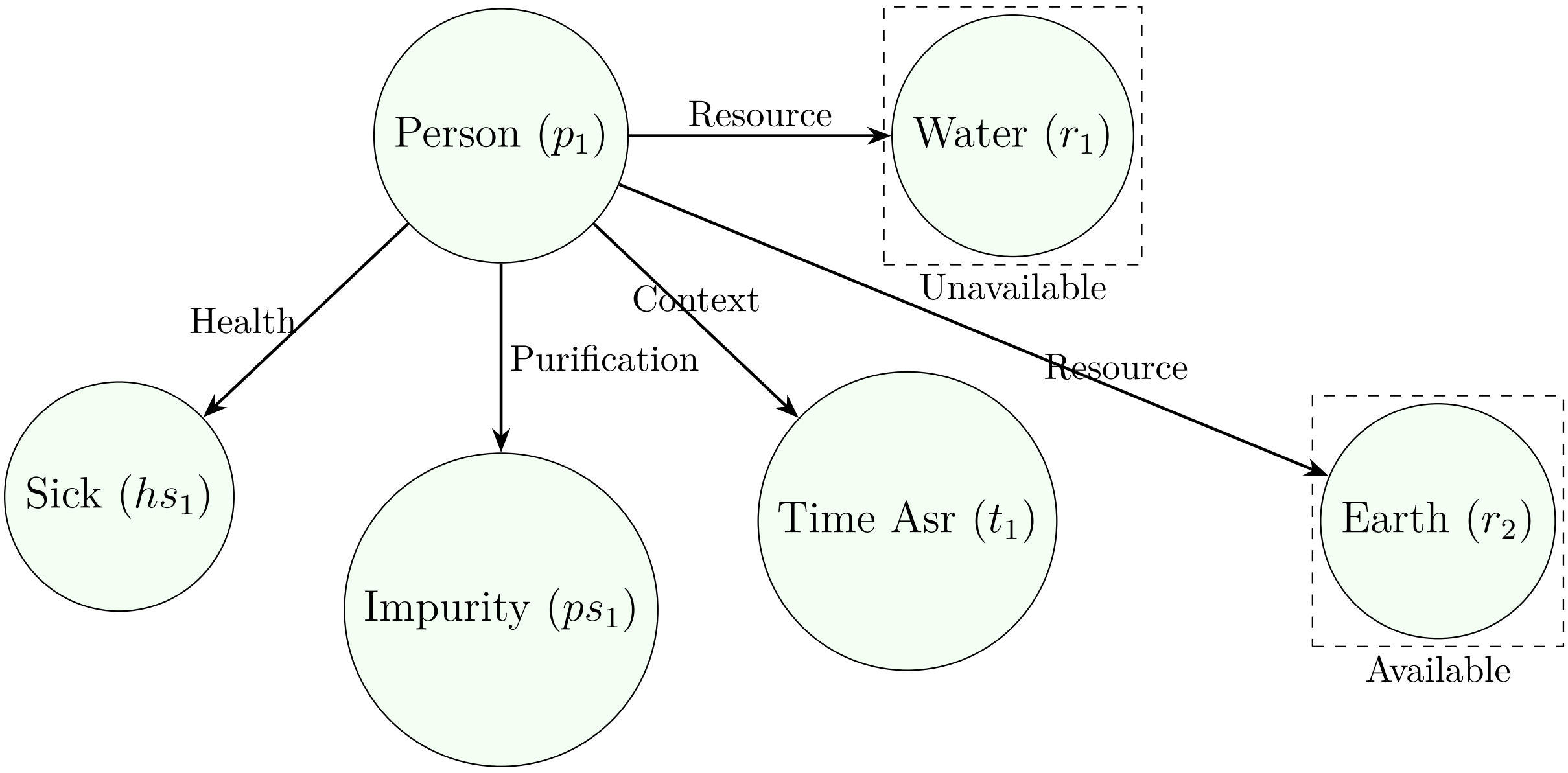
\textit{Figure 3: Visual Representation of the ARAC Semantic Graph}

The figure represents the result of semantic grounding after the
syntactic dependencies have been interpreted through the lexical
root-signatures and the corresponding Fiqh-grammar roles. For example,
the node \textit{Water} is obtained from the lexical signature of
\textarabic{مَاء}, while its status as a resource and its relation to
the person are supplied by the syntactic and jurisprudential
structure. Likewise, the \textit{Earth} node results from the lexical
signature of \textarabic{تُرَاب}, while its interpretation as the
available clean-earth resource results from combining the lexical
signature of \textarabic{طَاهِر} with the modifier relation and the
possessive structure.

Thus Figure~3 is not merely a relabeling of Figure~2. It is the result
of a semantic transformation in which lexical root-signatures are
combined with the syntactic dependencies and grammar-derived
interpretations to produce the reified domain objects and relations
required by ARAC.

\subsection*{5. Phase 4: Datalog Deductive Engine}

The Datalog engine combines the grounded ARAC facts with the
jurisprudential Horn clauses associated with the closed Tayammum
chapter.

The relevant legal rule is:

\begin{verbatim}
permissible(tayammum) :-
    health_state(P, sick),
    purification_state(P, minor_impurity),
    current_time(asr),
    unavailable(water),
    available(clean_earth).
\end{verbatim}

The deductive engine evaluates the query
\[
\texttt{permissible(tayammum)}.
\]

Its execution can be represented by the following sequence:

\begin{enumerate}
    \item \textbf{Goal:}
    evaluate \texttt{permissible(tayammum)}.

    \item \textbf{Match 1:}
    the rule requires
    \texttt{health\_state(P,sick)}.
    The knowledge base contains
    \texttt{health\_state(p1,sick)}, giving
    \(P=p_1\).

    \item \textbf{Match 2:}
    the rule requires
    \texttt{purification\_state(p1,minor\_impurity)}.
    The knowledge base contains the corresponding fact.

    \item \textbf{Match 3:}
    the rule requires
    \texttt{current\_time(asr)}.
    The corresponding fact is present.

    \item \textbf{Match 4:}
    the rule requires
    \texttt{unavailable(water)}.
    The corresponding fact is present.

    \item \textbf{Match 5:}
    the rule requires
    \texttt{available(clean\_earth)}.
    The corresponding fact is present.

    \item \textbf{Fire Rule:}
    all premises are satisfied, so the Horn clause fires.
\end{enumerate}

The resulting conclusion is

\begin{center}
\texttt{permissible(tayammum).}
\end{center}

The system therefore returns:

\begin{center}
\large
\textarabic{نَعَمْ، يَجُوزُ لَكَ التَّيَمُّمُ}
\end{center}

(Yes, it is permissible for you to perform Tayammum.)

\subsection*{6. Summary of Pipeline Integrity}

The example demonstrates the complete information flow through the
formal system:

\[
\begin{aligned}
\text{Arabic Input}
&\longrightarrow
\text{Ta\c{s}r\={i}f}
\longrightarrow
\text{Morphological Realizations}
\\
&\longrightarrow
\text{Fiqh-Grammar Slot Matching}
\longrightarrow
\text{Jurisprudential Role/Interpretation}
\\
&\longrightarrow
\text{Semantic Grounding}
\longrightarrow
\text{ARAC}
\longrightarrow
\text{Datalog}
\longrightarrow
\text{Legal Conclusion}.
\end{aligned}
\]

In particular, the syntactic stage does not assume jurisprudential
roles as properties of the input words. Morphological type and
inflectional realization are supplied by Ta\c{s}r\={i}f; the Fiqh
Grammar determines which slot a constituent can fill; the semantic
grounding component combines the resulting slot interpretation with
lexical and syntactic information to construct the ARAC representation;
and the Datalog engine performs the final legal conclusion.

The distinction is important even when several constituents have the
same inflectional realization. For example, both
\textarabic{مَاءٌ} and \textarabic{تُرَابٌ} are nouns marked
\(\mathrm{Raf'}\). The fact that one is interpreted as Water and the
other as CleanEarth is therefore not obtained by treating their
inflection marker as a semantic role. Rather, the applicable grammar
production determines the relevant slot, and the slot annotation
supplies the interpretation.

Similarly, where two prepositional constituents have identical
inflectional markers, their governing prepositions form part of the
complete realization used for slot matching. Thus the distinction
between such slots is determined by the complete attachment
specification rather than by \(Jarr\) alone.
\\

As a simple counterfactual check, suppose the user had instead stated:

\begin{center}
\textarabic{وَلَكِنْ عِنْدِي مَاءٌ طَاهِرٌ}
\end{center}

(``But I have pure water'').

The Ta\c{s}r\={i}f and I'r\={a}b stages would still establish the
appropriate morphological realizations and grammatical structure.
The semantic grounding stage, however, would map the relevant resource
to \texttt{Water} rather than \texttt{CleanEarth}, producing
\begin{center}
\texttt{available(water).}
\end{center}

This does not satisfy the Tayammum rule's premise

\begin{center}
\texttt{unavailable(water).}
\end{center}

Consequently, the Horn clause does not fire, and the system does not derive
\begin{center}
\texttt{permissible(tayammum)}
\end{center}

The example therefore illustrates the separation of responsibilities between the layers: Ta\c{s}r\={i}f supplies morphological information, the Fiqh Grammar performs
syntactic slot recognition and obtains the corresponding
jurisprudential interpretations, semantic grounding composes these
interpretations with lexical and syntactic information to construct
ARAC facts, and the Datalog engine performs the final legal
conclusion.

\subsection*{7. Step-by-Step Resolution of the Example}

The complete resolution of the query can be summarized as a sequence of
successive disambiguation and interpretation steps. The important point
is that no single layer determines the legal meaning by itself; the final
interpretation results from the successive combination of morphological,
grammatical, lexical, and semantic information.

\begin{enumerate}

\item \textbf{Vocalization determines the morphological realization.}

The fully vocalized input supplies the inflectional information required
by the syntactic layer. For example,
\textarabic{مَاءٌ}, \textarabic{تُرَابٌ}, and
\textarabic{طَاهِرٌ} are explicitly recognized as
\[
\begin{aligned}
\text{\textarabic{مَاءٌ}}
&\Rightarrow
(\mathrm{Noun},\mathrm{Raf'}),\\
\text{\textarabic{تُرَابٌ}}
&\Rightarrow
(\mathrm{Noun},\mathrm{Raf'}),\\
\text{\textarabic{طَاهِرٌ}}
&\Rightarrow
(\mathrm{Adjective},\mathrm{Raf'}).
\end{aligned}
\]

Likewise, the \textarabic{Jarr'} marking of
\textarabic{الْعَصْرِ} identifies it as the dependent constituent of
the Id\={a}fa construction
\textarabic{وَقْتُ الْعَصْرِ}. Thus vocalization resolves morphological
and syntactic ambiguities that would remain possible in an
unvocalized representation.

\item \textbf{Morphological realizations are matched against open
Fiqh-Grammar slots.}

The parser does not assign a jurisprudential meaning directly from a
root, word, or inflectional marker. Instead, the morphological
realization of each constituent is compared with the realization
requirements of the open slots of the applicable Fiqh-Grammar
production.

For example,
\[
\kappa(\text{\textarabic{مَرِيضٌ}})
=
\langle\mathrm{Adjective},\mathrm{Raf'}\rangle
\]
matches the realization required by the
$\mathit{StatePredicate}$ slot. Similarly,
\[
\kappa(\text{\textarabic{مَاءٌ}})
=
\langle\mathrm{Noun},\mathrm{Raf'}\rangle
\]
matches the resource slot of the negated-resource production, while
\textarabic{تُرَابٌ} matches the resource slot of the available-resource
production.

The crucial point is that the same morphological realization can occur
in different semantic contexts. Thus
\[
\langle\mathrm{Noun},\mathrm{Raf'}\rangle
\]
does not mean ``water'', ``earth'', or any other particular
jurisprudential concept. The applicable production and its slot
annotations determine the grammatical interpretation.

\item \textbf{The matched grammar slot supplies the jurisprudential
interpretation.}

Once a constituent has been matched to a slot, the annotation of that
slot supplies the relevant jurisprudential role or interpretation.
Thus the $\mathit{StatePredicate}$ slot identifies
\textarabic{مَرِيضٌ} as a health-state constituent; the
$\mathit{Resource}$ slot identifies \textarabic{مَاءٌ} and
\textarabic{تُرَابٌ} as resources in their respective constructions;
and the $\mathit{Action}$ slot identifies
\textarabic{التَّيَمُّمُ} as the action being queried.

Consequently, jurisprudential interpretation is obtained from the
grammar construction in which a constituent occurs, rather than being
read directly from its morphological form.

\item \textbf{The lexical entry and root-signature determine the lexical
semantic content.}

The grammar slot alone does not tell the system what the recognized
lexical item denotes. The semantic lexicon supplies this information
through the lexical entry and its root-signature.

Thus:
\[
\begin{aligned}
\text{\textarabic{مَاء}}
&\longrightarrow \texttt{WATER},\\
\text{\textarabic{تُرَاب}}
&\longrightarrow \texttt{EARTH},\\
\text{\textarabic{طَاهِر}}
&\longrightarrow \texttt{PURITY},\\
\text{\textarabic{الْعَصْر}}
&\longrightarrow \texttt{ASR},\\
\text{\textarabic{التَّيَمُّم}}
&\longrightarrow \texttt{TAYAMMUM}.
\end{aligned}
\]

The root-signature therefore supplies the lexical-semantic identity
which cannot be obtained from the morphological realization alone.

\item \textbf{Semantic grounding combines lexical meaning with
grammatical interpretation.}

The grounding component now combines the lexical signatures with the
grammar-derived interpretation and the syntactic dependencies.

For example,
\[
\texttt{EARTH}+\texttt{PURITY}
\longrightarrow
\texttt{CLEAN\_EARTH},
\]
because the lexical signature of \textarabic{طَاهِرٌ} is attached
syntactically to the resource denoted by \textarabic{تُرَابٌ}.

Likewise,
\[
\texttt{TIME}+\texttt{ASR}
\longrightarrow
\texttt{asr\_time},
\]
and the \texttt{ENTRY} predicate associated with
\textarabic{دَخَلَ} yields
\[
\texttt{current\_time(asr)}.
\]

The possessive and negative syntactic relations then transform the
grounded lexical entities into legal-state facts:
\[
\begin{aligned}
\text{\textarabic{لَيْسَ عِنْدِي مَاءٌ}}
&\longrightarrow
\texttt{unavailable(water)},\\
\text{\textarabic{عِنْدِي تُرَابٌ طَاهِرٌ}}
&\longrightarrow
\texttt{available(clean\_earth)}.
\end{aligned}
\]

\item \textbf{The grounded facts are reified in ARAC.}

The resulting semantic objects and relations are represented as ARAC
facts describing the current legal state:
\[
\begin{gathered}
\texttt{health\_state(p1,sick)},\\
\texttt{purification\_state(p1,minor\_impurity)},\\
\texttt{current\_time(asr)},\\
\texttt{unavailable(water)},\\
\texttt{available(clean\_earth)}.
\end{gathered}
\]

The interrogative constituent
\textarabic{هَلْ يَجُوزُ التَّيَمُّمُ؟} supplies the target query
\[
\texttt{permissible(tayammum)}.
\]

\item \textbf{The Datalog rule performs the final legal inference.}

The deductive engine checks whether all premises of the relevant
Tayammum rule are present in the grounded ARAC state. Since all five
required premises are satisfied, the rule fires and derives
\[
\texttt{permissible(tayammum)}.
\]

Thus the final legal conclusion is obtained only after the information
has passed through all preceding layers.

\end{enumerate}

The example therefore illustrates a cumulative resolution process:
\[
\boxed{
\begin{array}{c}
\text{Vocalization}\\
\Downarrow\\
\text{Morphological realization}\\
\Downarrow\\
\text{Fiqh-Grammar production and slot matching}\\
\Downarrow\\
\text{Jurisprudential role/interpretation}\\
\Downarrow\\
\text{Lexical root-signature}\\
\Downarrow\\
\text{Semantic grounding}\\
\Downarrow\\
\text{ARAC facts}\\
\Downarrow\\
\text{Datalog inference}\\
\Downarrow\\
\text{Legal conclusion}
\end{array}}
\]

In particular, vocalization resolves the morphological distinctions
needed for syntactic recognition; the Fiqh Grammar determines how the
recognized forms function within the jurisprudential construction; the
lexical root-signatures provide their lexical-semantic content; and
semantic grounding combines these sources into the domain-level facts
on which the deductive engine operates.

\section{Bridging the Gap Between Induction and Deduction in Computational Fiqh}

The preceding sections demonstrated how the Fiqh-Categorized Grammar ($G_{\mathrm{Fiqh}}^\delta$) is used to recognize a user's Classical Arabic input. However, an epistemological question remains: \textit{Where do these grammar productions come from?} If the grammar rules are hardcoded, the system suffers from a knowledge acquisition bottleneck and lacks a principled connection to the substantive law. 

This section bridges the gap between the inductive extraction of legal rules and their deductive application. We introduce a formal method that takes a Horn-clause program $P$—representing the Fiqh rulings $R$ of a closed chapter $C$ extracted via the inductive method of \textit{al-\d{S}abr wa al-Taq\c{s}\={\i}m}—and abstracts a specialized Fiqh-grammar $G_C$ capable of parsing any query $Q$ related to $C$.

\subsection{The Inductive Source: From Truth Tables to Horn Clauses}
Let $C$ be a closed Fiqh chapter (e.g., Tayammum). Following the methodology of \textit{al-\d{S}abr wa al-Taq\c{s}\={\i}m}, a truth table is constructed over the ontological concepts operative in $C$ [2]. An extraction algorithm deduces the minimal positive and negative legal rulings $R$. 

As established in prior work, these rulings $R$ are formally translated into a function-free Horn-clause program $P$. For example, the permissibility of Tayammum is encoded as a rule where the body contains premises regarding health state, impurity, time, water availability, and earth availability.

\subsection{The Ontology-to-Grammar Abstraction Algorithm}
To parse a natural language query $Q$ that triggers these rules, we must map the logical predicates of $P$ to the inflection-constrained non-terminals of $G_{\mathrm{Fiqh}}^\delta$. We define a formal abstraction function $\Omega: P \to G_C$ via Algorithm 1.

\begin{algorithm}[H]
\caption{Fiqh-Grammar Abstraction from Horn Clauses ($\Omega$)}
\label{alg:omega}
\begin{algorithmic}[1]
\Require Horn-clause program $P$, Ontological Meta-Map $M_{onto}$
\Ensure Fiqh-Grammar Productions $P^\delta_{Fiqh}$

\State $P^\delta_{Fiqh} \leftarrow \emptyset$

\State $P^\delta_{Fiqh} \leftarrow P^\delta_{Fiqh}
\cup
\{S \rightarrow \text{LegalQuery},
  S \rightarrow \text{LegalCondition}\}$

\For{each clause $H \leftarrow B_1,B_2,\ldots,B_k$ in $P$}

    \State $\text{CondList} \leftarrow [\ ]$

    \For{each premise $B_i$ in the body}

        \State Look up $B_i.\text{predicate}$ in $M_{onto}$

        \State $\text{NT}_i,\text{Rules}_i
        \leftarrow M_{onto}(B_i.\text{predicate})$

        \Comment{$\text{NT}_i$: $\delta$-typed non-terminal}

        \State $P^\delta_{Fiqh}
        \leftarrow
        P^\delta_{Fiqh}\cup\text{Rules}_i$

        \State Append $\text{NT}_i$ to $\text{CondList}$

    \EndFor

    \If{$H$ is a query target (e.g., \textit{permissible})}

        \State $P^\delta_{Fiqh}
        \leftarrow
        P^\delta_{Fiqh}
        \cup
        \{
        \text{LegalQuery}
        \rightarrow
        [\text{IntPart},
        \text{ModalVerb\_Raf'},
        \text{Action\_Raf'}]
        \}$

    \EndIf

    \State $P^\delta_{Fiqh}
    \leftarrow
    P^\delta_{Fiqh}
    \cup
    \left\{
    \text{LegalCondition}
    \rightarrow
    [\text{CondList}]_{\mathrm{ID}}
    \right\}$

\EndFor

\State \Return $P^\delta_{Fiqh}$

\end{algorithmic}
\end{algorithm}

Here $[\text{CondList}]_{\mathrm{ID}}$ denotes an Immediate Dominance
(ID) right-hand side: all non-terminals contained in
$\text{CondList}$ are required constituents of
$\texttt{LegalCondition}$, but their linear order is not specified by
the ID rule. Their surface ordering is subsequently constrained by the
Linear Precedence (LP) relations of the Fiqh-Categorized Grammar.

For each Horn clause processed by $\Omega$, a distinct
\texttt{LegalCondition} production is generated whose right-hand side
contains exactly the $\delta$-typed non-terminals corresponding to the
premises of that clause. The productions generated for different Horn
clauses are retained as separate alternatives in the resulting grammar.
The logical head of the clause is not encoded as a syntactic category;
it remains part of the Horn program $P$ and is evaluated by the
deductive engine after the linguistic premises have been grounded.

The Ontological Meta-Map $\mathcal{M}_{\mathrm{onto}}$ relates a predicate argument type to a Fiqh syntactic non-terminal indexed by a required $\delta.\text{mode}$. For instance, it maps Agent/Subject roles to \texttt{NP\_Raf'}, Patient roles to \texttt{NP\_Nasb}, and Resource/Instrument roles to \texttt{NP\_Jarr}.

\subsection{Concrete Example: Generating the Tayammum Grammar}
We demonstrate Algorithm 1 by generating the grammar for the chapter of Tayammum ($C = \text{Tayammum}$). The execution of the algorithm, displaying the input, the intermediate Meta-Map lookups, and the final assembled output, is visualized in Figure 1.

\setcounter{figure}{3}
\begin{figure}[H]
\centering
\begin{tikzpicture}[
    node distance=1.2cm,
    inputbox/.style={draw, rectangle, rounded corners, align=left, text width=11cm, minimum height=2cm, fill=blue!5, inner sep=10pt},
    mapbox/.style={draw, rectangle, rounded corners, align=left, text width=11cm, minimum height=4cm, fill=green!5, inner sep=10pt},
    outputbox/.style={draw, rectangle, rounded corners, align=left, text width=11cm, minimum height=2.5cm, fill=orange!10, inner sep=10pt},
    arrow/.style={->, >=Stealth, thick, draw=gray!80}
]

\node[inputbox] (input) {
    \textbf{Phase 1: Input -- Horn-Clause Program $P$} \\[5pt]
    \texttt{permissible(tayammum) :-} \\
    \texttt{\ \ health\_state(P, sick),} \\
    \texttt{\ \ purification\_state(P, minor\_impurity),} \\
    \texttt{\ \ current\_time(asr),} \\
    \texttt{\ \ unavailable(water),} \\
    \texttt{\ \ available(clean\_earth).}
};

\node[mapbox, below=of input] (map) {
    \textbf{Phase 2: Intermediate -- Ontology-to-Grammar Abstraction ($\Omega$ via $\mathcal{M}_{\mathrm{onto}}$)} \\[5pt]
    $B_1$: \texttt{health\_state} \quad $\xmapsto{\mathcal{M}_{\mathrm{onto}}}$ \quad \texttt{LegalState\_1 $\to$ Pronoun StatePredicate\_Raf'} \\
    $B_2$: \texttt{purification\_state} \quad $\xmapsto{\mathcal{M}_{\mathrm{onto}}}$ \quad \texttt{LegalEvent\_1 $\to$ ReflexiveActionVerb} \\
    $B_3$: \texttt{current\_time} \quad $\xmapsto{\mathcal{M}_{\mathrm{onto}}}$ \quad \texttt{LegalEvent\_2 $\to$ TimeVerb TimeAgent\_Raf' TimeBound\_Jarr} \\
    $B_4$: \texttt{unavailable} \quad $\xmapsto{\mathcal{M}_{\mathrm{onto}}}$ \quad \texttt{LegalNegation $\to$ NegVerb Adverb Resource\_Raf'} \\
    $B_5$: \texttt{available} \quad $\xmapsto{\mathcal{M}_{\mathrm{onto}}}$ \quad \texttt{LegalState\_2 $\to$ Adverb Resource\_Raf' Modifier\_Raf'}
};

\node[outputbox, below=of map] (output) {
    \textbf{Phase 3: Output -- Assembled Top-Level Fiqh-Grammar Productions (ID/LP Format)} \\[5pt]
    \texttt{LegalQuery $\to$ [IntPart ModalVerb\_Raf' Action\_Raf']} \\
    \texttt{LegalCondition $\to$ [LegalState\_1 LegalEvent\_1} \\
    \texttt{\quad LegalEvent\_2 LegalNegation LegalState\_2]$_{\mathrm{ID}}$}
};

\draw[arrow] (input) -- (map);
\draw[arrow] (map) -- (output);

\end{tikzpicture}
\caption{Phases of Grammar Abstraction via Algorithm $\Omega$}
\end{figure}

The execution of Algorithm $\Omega$ proceeds in the following exact steps, mapping the data structures from Phase 1 to Phase 3 as visualized in Figure 1:

\begin{enumerate}
    \item \textbf{Initialization (Lines 1-2):} The algorithm begins by initializing the grammar rule set $P_{\mathrm{Fiqh}}^\delta$ with the root productions $S \to \text{LegalQuery}$ and $S \to \text{LegalCondition}$. 
    \item \textbf{Premise Iteration \& Meta-Map Lookup (Lines 4-9):} The algorithm reads the input Horn clause $P$ (Phase 1) and iterates through its body premises. For each premise $B_i$, it queries the Ontological Meta-Map $\mathcal{M}_{\mathrm{onto}}$. 
    \begin{itemize}
        \item For $B_1$ (\texttt{health\_state}), $\mathcal{M}_{\mathrm{onto}}$ returns the non-terminal \texttt{LegalState\_1} and its expansion rule. This rule is added to $P_{\mathrm{Fiqh}}^\delta$, and \texttt{LegalState\_1} is appended to the $\text{CondList}$ data structure.
        \item This process repeats for $B_2$ through $B_5$, generating the intermediate rules shown in Phase 2. Each generated non-terminal (\texttt{LegalEvent\_1}, \texttt{LegalEvent\_2}, \texttt{LegalNegation}, \texttt{LegalState\_2}) is sequentially appended to $\text{CondList}$.
    \end{itemize}
    \item \textbf{Query Rule Generation (Lines 10-12):} The algorithm inspects the head of the clause $H$. Because $H$ is \texttt{permissible}, it recognizes this as a Fatwa target and generates the rule \texttt{LegalQuery $\to$ IntPart ModalVerb\_Raf' Action\_Raf'}.
    \item \textbf{Coordination Assembly (Lines 13-17):} For the current
Horn clause, the algorithm constructs a \texttt{CondSet} containing all
non-terminals in $\text{CondList}$, together with the required
\texttt{CoordPart} connectors. It then adds a single
\texttt{LegalCondition} production whose right-hand side represents the
complete set of premises of that clause. Productions generated for
different Horn clauses remain separate grammar alternatives.
    
    \item \textbf{Final Output:} The fully assembled coordination rule is added to $P_{\mathrm{Fiqh}}^\delta$ (Phase 3). The algorithm returns the complete set of generated productions, which forms the specialized Fiqh-grammar $G_C$ for chapter $C$.
\end{enumerate}

\subsection{Properties of the Bridging Method}
This inductive-to-deductive abstraction yields three critical formal properties:

\begin{definition}[Ontological Meta-Map]
Let $\mathcal{P}$ be the finite set of predicate symbols in the Datalog
program $P$. Let $N_\delta$ be the finite set of $\delta$-typed non-terminals
(e.g., \texttt{LegalState\_1}, \texttt{LegalEvent\_2}). Let $Q$ be the
finite set of context-free production rules. The Ontological Meta-Map is
a total function
\[
\mathcal{M}_{\mathrm{onto}} :
\mathcal{P} \rightarrow (N_\delta \times Q^*)
\]
such that for every predicate $p\in\mathcal{P}$,
\[
\mathcal{M}_{\mathrm{onto}}(p)=(N_p,\rho_p),
\]
where $N_p\in N_\delta$ is the unique $\delta$-typed non-terminal
associated with $p$, and $\rho_p\in Q^*$ is a finite sequence of
production rules expanding $N_p$.

The predicate-to-non-terminal assignment is injective:
\[
p_1\neq p_2
\quad\Longrightarrow\quad
N_{p_1}\neq N_{p_2}.
\]
Thus each predicate symbol has exactly one associated non-terminal, and
distinct predicate symbols are assigned distinct non-terminals. The
mapping $\mathcal{M}_{\mathrm{onto}}$ is therefore injective on predicate
symbols and preserves the one-to-one correspondence between logical
predicates and syntactic constituents.
\end{definition}

\begin{theorem}[Grammar-Legal Consistency]
Let $P$ be a function-free Horn-clause program representing the legal
rulings of a closed Fiqh chapter $C$. Let
$M_{\mathrm{onto}}$ be the Ontological Meta-Map of Definition~24,
mapping each predicate symbol occurring in $P$ to its corresponding
$\delta$-typed non-terminal and production rules. Let
$G_C=\Omega(P)$ be the Fiqh grammar generated by Algorithm~1. Then
$G_C$ is strictly consistent with $P$. In particular, a query $Q$ is
accepted by $G_C$ if and only if $Q$ provides the necessary
morphosyntactic material to instantiate the premises of some Horn
clause in $P$.
\end{theorem}

\begin{proof}
Let
\[
C_i:\quad H \leftarrow B_1,\ldots,B_k
\]
be any clause in $P$. By Definition~24, every predicate occurring in
the body of $C_i$ has a unique corresponding $\delta$-typed
non-terminal and production rules under $M_{\mathrm{onto}}$. Therefore,
Algorithm~1 generates a grammar constituent for every premise
$B_j$, without omission or conflation of distinct predicates.

The body of the clause is a conjunction, and conjunction is independent
of the order in which its premises are written. Algorithm~1 therefore
represents the corresponding constituents as required slots of an
ID/LP rule. Thus the generated grammar preserves the structural
correspondence
\[
B_1\wedge\cdots\wedge B_k
\quad\longleftrightarrow\quad
[N_1,\ldots,N_k],
\]
where $N_j$ is the non-terminal associated with $B_j$.

Suppose first that $Q\in L(G_C)$. A successful parse of $Q$ fills all
required slots of a generated clause structure. Each filled slot
corresponds, through $M_{\mathrm{onto}}$, to one of the premises
$B_1,\ldots,B_k$. Hence the morphological and syntactic material of
$Q$ instantiates every premise required by the corresponding Horn
clause.

Conversely, suppose that $Q$ provides the necessary morphosyntactic
material to instantiate the premises of a clause
\[
H \leftarrow B_1,\ldots,B_k.
\]
For every $B_j$, the corresponding non-terminal $N_j$ and its
production rules are present in $G_C$ by the construction of
$\Omega$. Since the generated clause structure is represented in
ID/LP form, the supplied constituents may occur in any permitted
surface order. The $\delta$-typed non-terminals ensure that each
constituent satisfies the required morphological and syntactic
constraints. Consequently, all required slots can be filled and the
query $Q$ has a valid parse in $G_C$.

Therefore,
\[
\begin{aligned}
Q\in L(G_C)
\quad\Longleftrightarrow\quad
&Q\text{ provides the necessary morphosyntactic material}\\
&\text{to instantiate the premises of some clause in }P.
\end{aligned}
\]

Thus $G_C$ is strictly consistent with the logical structure of $P$.
\end{proof}

\begin{theorem}[Decidability and Linear Complexity of the Generated Grammar]
Let $G_C=\Omega(P)$ be the Fiqh grammar generated by Algorithm~1 from
a finite function-free Horn-clause program $P$ and the Ontological
Meta-Map $M_{\mathrm{onto}}$ of Definition~24. Assume that each
governing predicate has a fixed finite valence $c$, independent of the
input length $n$, and that the $\delta$-typed slot structure provides
constant-time identification of the unique compatible open slot.
Then:
\begin{itemize}
    \item $L(G_C)$ is decidable;
    \item every query $Q$ of length $n$ can be parsed deterministically
    in $O(n)$ time.
\end{itemize}
\end{theorem}

\begin{proof}
Since $P$ and $M_{\mathrm{onto}}$ are finite, Algorithm~1 generates a
finite set of grammar productions. By Theorem~19, the resulting
production system is an ID/LP grammar equivalent to a context-free
grammar. The attribute evaluation stage performs only finite
unification and agreement checks. Therefore, membership in
$L(G_C)$ is decidable.

For the complexity bound, consider an input query
\[
Q=(m_1,\ldots,m_n).
\]
The parser scans the input from left to right. For each terminal
$m_i$, its $\delta_i.\mathrm{mode}$ identifies at most one compatible
open slot. Since every governing predicate has valence at most the
fixed constant $c$, locating and checking the corresponding slot
requires $O(c)=O(1)$ operations. The required feature-unification
checks are also finite and therefore take $O(1)$ time per terminal.

The construction of the dependency graph processes each terminal once
and introduces at most a constant number of edges per terminal.
Consequently, graph construction and the associated finite
unification checks require $O(n)$ time.

Thus the total running time is
\[
O(n)+O(n)=O(n).
\]

Hence $L(G_C)$ is decidable, and the deterministic parser recognizes
every query in $O(n)$ time.
\end{proof}

\subsection{Discussion}
This bridging methodology resolves the knowledge acquisition bottleneck of computational Fiqh. Rather than a linguist manually writing syntax rules for every possible legal scenario, the Fiqh-grammar is dynamically abstracted from the inductively extracted Horn clauses of a Madhhab. 

The \textit{al-\d{S}abr wa al-Taq\c{s}\={\i}m} algorithm acts as the inductive Knowledge Engineer, mining structural rules from a closed chapter. $\Omega$ acts as the linguistic projector, mapping those abstract legal concepts to the surface morphology of Classical Arabic. Finally, the $\delta$-driven parser and Datalog engine act as the deductive Mufti, applying these rules to unstructured real-world Arabic scenarios. This establishes an end-to-end computational pipeline from the inductive extraction of Fiqh rules to their linguistic realization and subsequent deductive application, relative to the correctness properties established for the inductive and deductive components of the framework.

\section{Challenges and Future Work}

While the mathematical foundations of the proposed formal system are rigorously established, the transition from abstract theory to a practical, executable system introduces a distinct class of engineering and knowledge-representation challenges. Because the architecture makes strict assumptions regarding closed chapters, total bijections, and fully vocalized input, the practical implementation bottlenecks are highly localized. This section outlines the most significant of these challenges and defines the trajectory for future empirical work.

\subsection{The Knowledge Acquisition Bottleneck: \texorpdfstring{$P$}{P} and \texorpdfstring{$M_{onto}$}{M\_onto}}
The bridging methodology (Algorithm $\Omega$) shifts the burden of system development from manually writing complex syntax rules to writing Datalog programs ($P$) and defining the Ontological Meta-Map ($M_{onto}$) for a closed Fiqh chapter $C$. While mathematically sound, this presents a formidable knowledge engineering task.

First, the inductive extraction of legal rulings via \textit{al-Ṣabr wa al-Taqşīm} requires constructing truth tables over the ontological concepts operative in $C$ [2]. For simple chapters (e.g., Tayammum), this is highly tractable. However, for complex chapters (e.g., \textit{Bay‘} or \textit{Nikah}), which may involve dozens of interacting legal variables, conditions, and exceptions, the truth table grows exponentially. Challenges of this nature are discussed and partially dealt with in our other publications on the subject.

Second, the knowledge engineer must define $M_{onto}$ as a total
injective mapping, assigning every logical predicate in $P$ to a
specific Arabic morphosyntactic non-terminal ($\delta$-typed).

\subsection{Lexical Grounding and Synonymy Management}
The most profound practical challenge in the system lies in managing lexical synonymy without compromising computational tractability or epistemological soundness. In Islamic legal theory (\textit{Uṣūl al-Fiqh}), a legal ruling (\textit{Hukm}) operates on a canonical \textit{concept} (e.g., ``the entry of prayer time''), not on a specific subset of Arabic surface expressions. A closed Fiqh chapter inherently contains synonymy clusters; for example, the concept of prayer time entering may be expressed as \textarabic{دخل وقت الصلاة}, \textarabic{جاء وقت الصلاة}, or \textarabic{أذن المؤذن}.

If the system attempts to handle this synonymy at the Datalog level by writing multiple Horn clauses for every permutation of synonyms, it faces an exponential rule explosion. Furthermore, this approach is epistemologically flawed because it conflates the logical ruling with the linguistic expression.

The correct architectural solution, provisioned by the stratified design of this paper, is to handle synonymy at the Semantic Grounding layer (ARAC). The Ontological Meta-Map ($M_{onto}$) provides the structural template (e.g., \texttt{[TimeVerb, Time\-Agent\_Raf', Time\-Bound\_Jarr]}), ensuring the parser remains strictly structural and $O(n)$ unambiguous. The Lexical Knowledge Base, accessed via the semantic signatures $\Sigma(r)$ (Definitions 9 and 19), must be populated with chapter-specific synonymy clusters.

When ARAC receives the syntactic dependency graph, it reads the canonical roots. If the roots match an ontological synonymy cluster, ARAC normalizes the expression into a single, language-independent Grounded Action Structure (GAS), outputting a canonical Datalog fact (e.g., \texttt{prayer\-\_time\-\_entered(a1)}). Consequently, the Datalog rules remain clean, simple, and free of synonym permutations. The practical challenge, therefore, is the lexical engineering required to build these synonymy clusters---a task analogous to compiling a specialized legal dictionary (\textit{Mu‘jam}) for each Fiqh chapter.

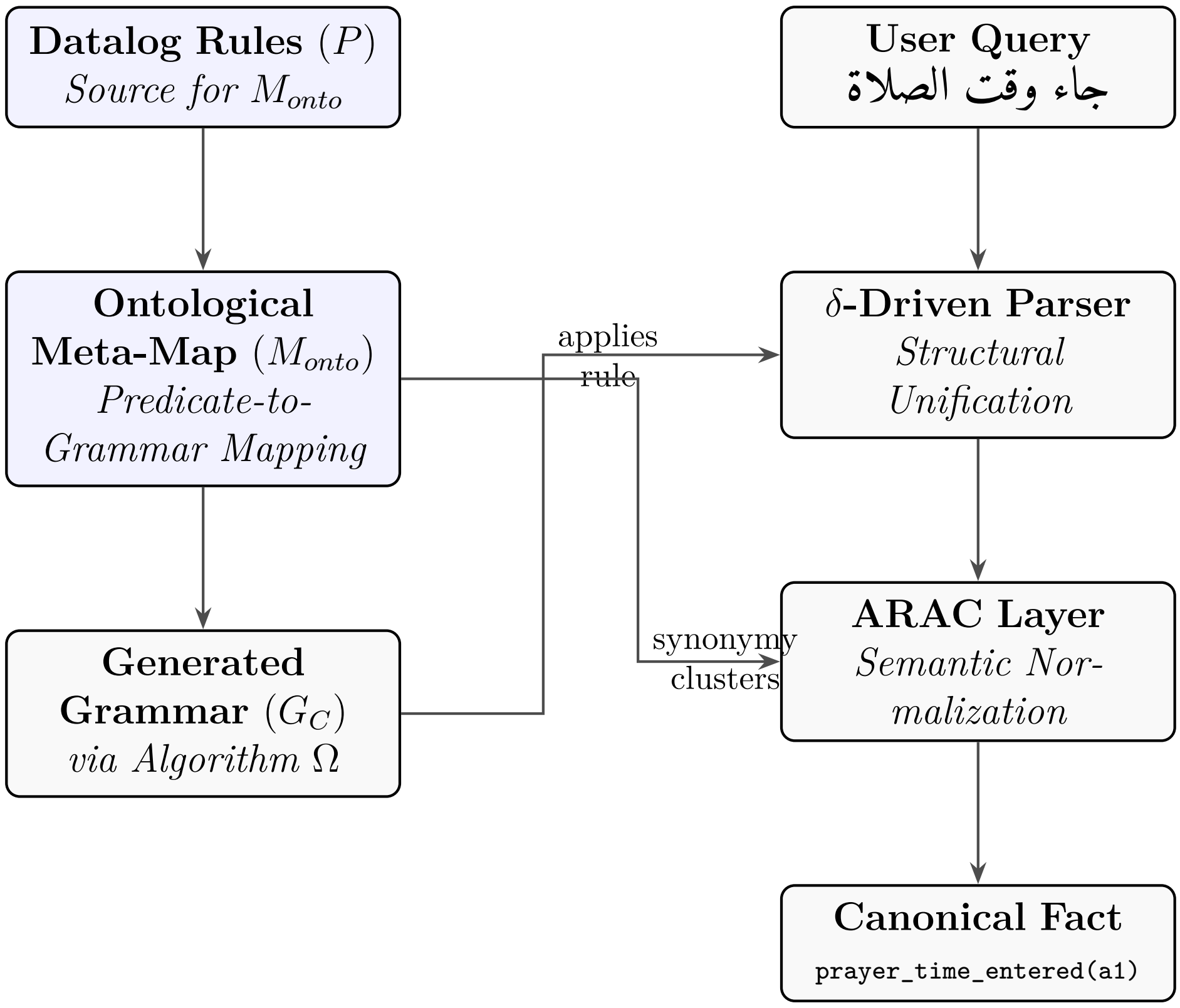
\begin{figure}[H]
\centering
\begin{tikzpicture}[
    node distance=1.5cm and 4cm,
    box/.style={draw, rectangle, rounded corners, align=center, text width=3.8cm, thick, fill=gray!5, inner sep=5pt},
    arrow/.style={->, >=Stealth, thick, draw=black!70}
]

\node[box, fill=blue!5] (datalog) {\textbf{Datalog Rules} ($P$)\\\textit{Source for $M_{onto}$}};

\node[box, fill=blue!5, below=of datalog] (monto) {\textbf{Ontological Meta-Map} ($M_{onto}$)\\\textit{Predicate-to-Grammar Mapping}};

\node[box, below=of monto] (grammar) {\textbf{Generated Grammar} ($G_C$)\\\textit{via Algorithm $\Omega$}};

\node[box, right=of datalog] (query) {\textbf{User Query}\\\textarabic{جاء وقت الصلاة}};
\node[box, below=of query] (parser) {\textbf{$\delta$-Driven Parser}\\\textit{Structural Unification}};
\node[box, below=of parser] (arac) {\textbf{ARAC Layer}\\\textit{Semantic Normalization}};
\node[box, below=of arac] (fact) {\textbf{Canonical Fact}\\\scriptsize\texttt{prayer\_time\_entered(a1)}};

\draw[arrow] (datalog) -- (monto);
\draw[arrow] (monto) -- (grammar);

\draw[arrow] (query) -- (parser);
\draw[arrow] (parser) -- (arac);
\draw[arrow] (arac) -- (fact);

\draw[arrow] (grammar.east) -- ++(1.5cm,0) |- node[pos=0.5, right, font=\footnotesize, align=center] {applies\\rule} (parser.west);

\draw[arrow] (monto.east) -- ++(2.5cm,0) |- node[pos=0.5, right, font=\footnotesize, align=center] {synonymy\\clusters} (arac.west);

\end{tikzpicture}
\caption{Separation of Design-Time and Run-Time phases. The Datalog program ($P$) acts strictly as the source for the Ontological Meta-Map ($M_{onto}$) during grammar generation. At Run-Time, the parser applies the generated grammar, and ARAC uses the synonymy clusters to normalize the user query into a canonical fact.}
\label{fig:monto_arch}
\end{figure}

\subsection{The Input Barrier: Full Vocalization}
The mathematical guarantees of unambiguity (Lemma 1) and deterministic $O(n)$ parsing rely entirely on the input being fully vocalized Classical Arabic (\textit{mushakkal}). Modern Arabic keyboards and standard user inputs rarely provide full diacritics.

To make the system practically accessible, a highly accurate automatic diacritizer must be integrated as a pre-processing step before the Taşrīf procedure [9]. However, this introduces a point of brittleness: if the diacritizer assigns an incorrect case ending (e.g., a \textit{Kasrah} instead of a \textit{Fatḥah}), the $\delta$-driven parser will deterministically accept the wrong legal role, potentially yielding an incorrect fatwa. Future work must explore robust error-recovery mechanisms or probabilistic $\delta$-weighting at the morphological boundary to mitigate this input noise.

\subsection{Scaling and Empirical Validation}
Future work will proceed in phased empirical validation. Phase 1 requires building a proof-of-concept system restricted entirely to a single, small, closed chapter (e.g., Tayammum), manually populating the Datalog rules and the semantic synonymy clusters to prove the end-to-end pipeline. Phase 2 will focus on integrating large-scale traditional Arabic morphological databases into the $H_{repo}$ hash repository. Finally, Phase 3 will test the limits of Algorithm $\Omega$ when scaling to larger, more complex chapters, evaluating the maintainability of the generated ID/LP grammars and the performance of the deterministic dependency parser on extensive real-world jurisprudential queries.

\section{Conclusion}

This paper has presented a mathematically precise, computationally tractable, and formally verifiable computational framework for automated Islamic jurisprudence. The three-layer stratified architecture---consisting of the abstract algebraic composition (Section~2), the general grammatical and semantic grounding layer (Sections~3--4), and the concrete Fiqh-Categorized Attribute Grammar (Sections~10--12)---has been rigorously developed and shown, under its stated assumptions, to satisfy the following properties:

\begin{itemize}
\item \textbf{Correctness:} Every legal conclusion derived by the system is a logical consequence of the input and the jurisprudential knowledge base (Theorems~12, 17, 24).

\item \textbf{Relative Completeness:} Every legal conclusion expressible within the adopted formal language and derivable from the represented jurisprudential rules is computable by the framework (Theorems~13, 25, 27).

\item \textbf{Termination:} Every computation performed by the system terminates after finitely many steps (Theorems~6, 10, 14, 17, 20, 28).

\item \textbf{Decidability:} The legal inference problem for the adopted formal system is decidable (Theorems~15, 26, 28).

\item \textbf{Polynomial Tractability:} The deterministic $\delta$-driven parser achieves $O(n)$ complexity (Theorems~20, 28), fulfilling Principle~V.

\end{itemize}

The word ``completeness'' in these statements must be understood relative to the formally specified domain. The paper does not claim that all of Islamic jurisprudence has been formalized, nor that every conceivable legal question has already been encoded. Rather, it establishes a general formal mechanism by which a \emph{closed Fiqh chapter}, once its ontology, admissible questions, legal rules, and formal grammar have been specified, can be represented and processed in a complete and decidable manner relative to that specification.

This distinction is essential because the present framework separates two logically different tasks. The first is the \emph{inductive extraction and representation} of jurisprudential rules; the second is the \emph{deductive application} of those rules to particular cases. The present work is principally concerned with the formal correctness, completeness, termination, and decidability of the second task, while providing a formal bridge to the first through the methodology of \textit{al-\d{S}abr wa al-Taq\c{si}{\i}m}.

\subsection*{On Consistent and Complete Legislation}

The distinction between \emph{Consistent Legislation} and \emph{Complete Legislation}, introduced in the earlier 2016 formulation of the computational Fiqh program, is particularly important for interpreting the present results.

The original formulation stated the two principles in the following direct form:

\begin{quote}
\textbf{Consistent Legislation:} In Islamic Jurisprudence, there exists a Fiqh-rule for the deduction of any legal verdict, deduction being done using a formal logic system.

\medskip

\textbf{Complete Legislation:} For every query or legal Fiqh-question there exists a legal verdict.
\end{quote}

The present work gives these principles a more precise computational interpretation. Consistent Legislation is represented by the requirement that a legal conclusion produced by the system be supported by an explicit rule or reason within the jurisprudential knowledge base. Complete Legislation, within a closed formal domain, becomes the requirement that every admissible question in the specified question space be associated with a legal ruling.

Thus, if $SF_{CC}$ denotes the set of admissible questions of a closed Fiqh chapter $CC$, and $QR_{CC}$ denotes its set of represented rulings, Complete Legislation is expressed as:

\begin{equation}
\forall M\in SF_{CC},\quad
\exists R\in QR_{CC}:
M\longrightarrow R.
\end{equation}

The present framework does not assume this property merely because a formal system has been constructed. Rather, it provides the machinery by which the property can be represented, tested, and, when the relevant rule repository has been completely generated, mechanically applied.

This gives a precise interpretation to the earlier 2016 Entscheidungsproblem. The question is not simply whether an abstract logical system exists, but whether the specified jurisprudential domain can be represented by a formal language and rule system for which the relevant questions, rules, and deductions are all effectively decidable.

\subsection*{On the Relation Between Inductive Extraction and Deductive Application}

The bridging methodology introduced in Section~12 formalizes the connection between the inductive extraction of legal rulings through \textit{al-\d{S}abr wa al-Taq\c{si}{\i}m} and their deductive application through the Fiqh grammar and Datalog engine.

A potential concern is that this methodology assumes the existence of a Horn-clause program $P$ representing the legal rulings of a closed Fiqh chapter. This assumption should not be understood as an evasion of the formalization problem. It is a deliberate modularization of two distinct tasks.

The first task is to determine, from the relevant scriptural, linguistic, and jurisprudential evidence, the rules that constitute the legal knowledge of the chapter. The second is to construct a formal mechanism capable of applying those rules correctly and completely to admissible queries.

The formalization of jurisprudential rulings into Horn-clause programs remains a substantial research task in its own right. The methodology of \textit{al-\d{S}abr wa al-Taq\c{si}{\i}m} provides the proposed basis for this inductive stage. The present paper contributes the formal machinery that connects its output to a terminating, decidable, and formally analyzable deductive system.

The distinction is therefore:

\begin{equation}
\text{Inductive extraction of rules}
\longrightarrow
P
\longrightarrow
\text{formal deduction}.
\end{equation}

The present paper establishes the correctness and computational properties of the right-hand side and provides the formal bridge connecting it to the left-hand side. To require the present paper itself to formalize every chapter of Islamic jurisprudence would conflate the construction of the formal computational language with the independent task of populating that language with the complete jurisprudential knowledge of every possible domain.

\subsection*{On the Meaning of Completeness in the Present Framework}

The results of this paper should therefore be understood as establishing a hierarchy of related but distinct completeness properties.

At the logical level, the relevant formal theories may possess syntactic completeness under the model-theoretic conditions established for their respective structures.

At the computational level, the adopted inference mechanism is complete relative to the represented Horn rules and the formal grammar.

At the jurisprudential level, a closed chapter is complete relative to its explicitly defined ontology, question space, and rule repository when every admissible question in that space is represented by an applicable ruling.

These levels may be summarized as:

\begin{equation}
\begin{aligned}
\text{Logical completeness}
&\longrightarrow
\text{completeness of the formal theory},\ \\
\text{Inference completeness}
&\longrightarrow
\text{completeness of deduction from represented rules},\ \\
\text{Relative Fiqh completeness}
&\longrightarrow
\text{coverage of the specified closed jurisprudential domain}.
\end{aligned}
\end{equation}

None of these statements entails that all of Islamic jurisprudence has already been formalized. Rather, they establish the conditions under which a specified jurisprudential domain can be transformed into a formally complete computational object.

\subsection*{On the Scalability and Compositionality of Unambiguity}

The unambiguity of the Fiqh-Categorized Attribute Grammar is proven in Lemma~1. The proof is structural: every production rule assigns distinct, non-repeating $\delta$ indices to each non-terminal slot, and every terminal possesses a unique $\delta$ mode.

This property is compositional. Any extension of the grammar that adheres to the same construction pattern---assigning distinct indices to slots---preserves unambiguity. The property is therefore not contingent on the particular set of rules currently included in the grammar, but on the rule-construction mechanism itself.

Consequently, as the grammar is expanded to cover additional Fiqh chapters, unambiguity is preserved provided that each extension satisfies the established construction conditions. The mathematical proof need not be reconstructed from first principles for every new chapter; rather, the preservation follows from the same structural condition.

\subsection*{On Implementation and Empirical Validation}

This paper is primarily a formalization paper. Its purpose is to establish the mathematical foundations of a computational system for Islamic jurisprudence.

Formal verification precedes implementation and empirical validation. Before a system can be implemented and tested, its formal objects, inference mechanisms, and computational properties must first be specified precisely. The present work establishes correctness, relative completeness, termination, and decidability under its stated assumptions.

Implementation and empirical evaluation remain important subsequent stages. They include, among other things, the construction of executable morphological and syntactic analyzers, the encoding of jurisprudential rule repositories, the testing of complete closed chapters, and the measurement of practical performance.

The absence of a complete implementation therefore does not invalidate the mathematical results established here. The paper's contribution is the formal foundation upon which such implementations can subsequently be constructed and evaluated.

\subsection*{On the Historical Significance of the Algebraic Representation of Inheritance}

The algebraic treatment of inheritance associated with Imam al-Khwārizmī provides an instructive historical example of the methodological principle underlying the present work.

Its significance is not merely that particular inheritance problems could be solved numerically. More importantly, algebra provided a general symbolic representation for a defined class of problems. Legal constraints concerning shares, liabilities, bequests, and remaining estate could thereby be expressed within a common mathematical structure and subjected to general procedures rather than being handled exclusively as isolated numerical cases.

In this sense, the transition to algebra represents an instance of a broader jurisprudential methodological objective: replacing a collection of separate solutions with a جامع مانع representation of a class of cases, and thereby enabling general procedures that reduce particulars to governing rules.

The historical lesson is not that every jurisprudential question should be reduced to algebra. Rather, it is that when a suitable mathematical or logical representation exists for a defined class of legal problems, adopting that representation can raise the level of abstraction, generality, verification, and algorithmic applicability of the legal method.

This principle is directly relevant to the computational formalization developed in the present paper. The objective is not to replace jurisprudential reasoning with arbitrary computation, but to identify the appropriate formal representation for a specified domain and then construct a general, terminating, decidable mechanism for applying the represented rules.

\subsection*{On the Possible Impact of the Current Formalization}

The formalization developed in this paper may have consequences extending beyond the construction of a computational engine for the derivation of Fiqh-rulings. Its potentially deeper significance lies in a structural characteristic that distinguishes Fiqh from most modern systems of positive or secular jurisprudence: the ultimate normative source of the Fiqh-system is not constructed solely from human legislation, social convention, judicial precedent, or empirically determined policy objectives, but is explicitly related to divine revelation. Consequently, concepts such as justice, obligation, prohibition, rights, responsibility, and the preservation of human interests are not introduced into the system merely as external philosophical criteria that must subsequently be reconciled with formally specified legal rules. Rather, they are incorporated into the normative structure of the law through the revealed sources and the juristic methodology used to interpret and apply them.

This distinction is important for computational jurisprudence. A purely formal legal system can, in principle, determine whether a conclusion follows from a specified set of legal premises. The more difficult question is what normative content those premises should have in the first place. In modern jurisprudential systems, questions concerning justice, morality, proportionality, human welfare, fairness, and the ethical acceptability of a legal outcome may therefore remain partly dependent upon interpretation, competing theories of value, legislative policy, or judicial discretion. The formal legal mechanism may be perfectly precise while the normative criteria governing the choice or interpretation of its rules remain external to that mechanism.

The Fiqh-system approaches this problem differently. Because its normative rules are ultimately connected to revelation, the semantic content of the legal system is not required to be reconstructed independently by the computational mechanism from unrestricted philosophical notions of justice or morality. The formal system receives its normative premises from a prior revealed and juristically interpreted source. The computational task is consequently shifted from inventing or independently deciding the ultimate values to representing, organizing, and correctly applying the normative rules that embody those values.

This does not mean that every semantic question in Fiqh is automatically trivial, nor that all questions of interpretation, conflicting evidence, empirical knowledge, or \textit{uṣūl al-fiqh} disappear once revelation is introduced. On the contrary, these remain genuine scholarly problems. The important point is more precise: once the normative premises of a particular Fiqh-domain have been established, the computational system need not independently solve the philosophical problem of what constitutes justice or moral goodness in the abstract. Those normative commitments are already incorporated into the jurisprudential knowledge base from which the formal derivation proceeds.

This observation places the present work within a much older philosophical aspiration: the possibility of making normative and philosophical reasoning sufficiently precise to admit systematic treatment. Kurt Gödel is particularly significant in this connection. Gödel's philosophical writings reveal a sustained commitment to rationalism and to the possibility of developing exact methods in philosophy. His aspiration was not merely to make philosophical argumentation more rigorous in an informal sense, but to approach a level of precision comparable, in appropriate respects, to that achieved by mathematical reasoning. This aspect of Gödel's thought is summarized by noting his aspiration to develop exact methods in philosophy and to transform philosophy into an exact science. Two statements were recorded from Gödel's list of fourteen philosophical convictions: `There are systematic methods for the solution of all problems'' and `There is a scientific (exact) philosophy and theology, which deals with concepts of the highest abstractness'' [26].

These statements do not constitute a proposal by Gödel for a formal system of ethics in the technical sense used in mathematical logic, and the present paper makes no such historical claim. Rather, they establish a broader philosophical aspiration: that problems traditionally regarded as belonging to philosophy, theology, and normative reasoning might admit sufficiently exact conceptual analysis and systematic methods of solution. Gödel's own philosophical position was substantially more nuanced than a simple reduction of philosophy to formal calculation; nevertheless, his writings clearly express the conviction that rigorously systematic treatment of highly abstract philosophical concepts is possible [26].

From this perspective, the formalization presented in the present paper suggests a potentially important possibility. Fiqh may constitute a historically developed normative system in which a substantial body of ethical and juridical content is already anchored in a transcendent normative source and organized through an elaborate tradition of definitions, distinctions, causes, conditions, exceptions, analogical reasoning, and rules of inference. If a closed Fiqh-domain can then be represented by a formally specified knowledge base and processed by a sound, complete, terminating, and decidable inference mechanism, the result is not merely an automated calculator for legal verdicts. It is an example of a normative system in which semantic commitments and formal inference can be connected within a single structured framework.

In this limited but important sense, the present formalization may be viewed as a possible practical realization of an aspiration that has repeatedly appeared in the history of rational philosophy: the construction of a sufficiently precise normative system in which questions of obligation and permitted action can be represented by formal structures and subjected to rigorous inference. The distinctive feature of Fiqh is that the normative content is not generated by the formal system itself. The formal system provides the mechanism of representation and deduction, while the normative content is grounded in revelation and in the established methods of juristic interpretation.

The possible impact of the present work should therefore be understood at three distinct levels. First, it provides a formal computational mechanism for the derivation of rulings within the adopted Fiqh fragment. Second, it demonstrates that normative legal knowledge can be connected to a mathematically specified inference architecture without reducing the normative content to arbitrary computational conventions. Third, it suggests a broader research program in which increasingly large portions of Fiqh may be represented as formally specified normative theories, thereby allowing questions of consistency, completeness relative to a domain, decidability, and computational complexity to become objects of rigorous mathematical investigation.

The claim is necessarily limited. This paper does not establish that all of Fiqh is decidable, that all ethical questions have a unique computational solution, or that formalization eliminates the need for human juristic interpretation. Rather, it establishes a formal framework in which such questions can be studied with considerably greater precision. If the program can subsequently be extended to larger and more sophisticated Fiqh-domains while preserving soundness, completeness relative to the relevant domain, and decidability, then the resulting body of work could provide a concrete example of how a revealed normative tradition can be transformed into an exact computational science of legal reasoning.

In that sense, the possible significance of computational Fiqh extends beyond automation. It concerns the possibility of joining three elements that have historically been treated as belonging to different domains: revealed normative knowledge, rigorous formal logic, and executable computation. The present formalization provides a first mathematical framework for investigating whether such a union can be achieved systematically.

\subsection*{Closing Remarks}

The principal contribution of this work is therefore not merely the construction of another symbolic representation of legal language. It is the establishment of a formally analyzable computational architecture in which linguistic grounding, jurisprudential representation, and deductive inference are connected by explicit mathematical interfaces.

Under the assumptions stated throughout the paper, the resulting system is correct, relatively complete, terminating, and decidable. The stratified architecture permits each layer to be analyzed independently, while the bridging methodology provides a principled connection between inductive legal-rule extraction and deductive legal application.

The foundation is consequently laid for a computational treatment of Islamic jurisprudence in which the traditional objective of deriving general rules capable of governing particular cases can be connected to modern formal methods. The next stages are the systematic formalization of additional closed Fiqh chapters, implementation of the corresponding linguistic and deductive components, and empirical validation of the resulting systems.

The ultimate claim of the paper is therefore deliberately precise: when a jurisprudential domain is explicitly bounded, its ontology and admissible questions are specified, its legal rules are formally represented, and the corresponding inference mechanism satisfies the proved conditions, that domain can be transformed into a computational system whose legal reasoning is mathematically analyzable, terminating, and decidable.

\newpage
\vspace*{\fill}
\begin{center}
    \section*{Dedication}
    \addcontentsline{toc}{section}{Dedication}
    \vspace{0.5cm}
    {\itshape
    To my first and foremost teacher\\[0.5cm]
    The woman who became the first certified female fiqh scholar in modern history\\
    Whose seminal work in women-related jurisprudence remains unparalleled in its rigor, objectivity and impact\\
    The talented poet, linguist and gifted arabic language teacher \\
    The fierce fighter for justice, equality and freedom \\
    The unprecedented rare individual who surpassed all her peers \\ ... \\
    My colleague and co-author of the first paper in fiqh-formalization, the one upon which this work is based\\
    ...\\
    To the one who gave me my name and introduced me to Islam, teaching me in every single detail of her life to love and cherish its unequaled principles\\
    ... \\ 
    My best friend, mentor, role model and beloved mother:\\[0.25cm]
    \textbf{Nadra Ahmad Shannan}\\[0.5cm]
    Hoping that from her current place in heaven, Inshallah, she is able to see the seeds of her own work grow to become what she intended them to be\\
    Amen...
    }
\end{center}
\vspace*{\fill}
\newpage



\begin{thebibliography}{99}

\item[]\textbf{Main References}

\bibitem{abdelwahab2016}
Abdelwahab, E. M., Daghbouche, K., \& Shannan, N. A. (2016).
The Algorithm of Islamic Jurisprudence (Fiqh) with Validation of an Entscheidungsproblem.
\textit{Journal Academica (New York)}, 4(2), 52–87.
\url{http://journalacademica.org}

\bibitem{abdelwahab2026sabr}
Abdelwahab, E. (2026).
Computational Extraction of Legal Causes via al-Sabr wa al-Taqsim: A Set-Theoretic Formalization for Closed Fiqh Chapters.
\textit{Journal Academica}, 15(1), 3–18.
\url{https://arxiv.org/abs/2607.25605}

\bibitem{abdelwahab2026meaning}
Abdelwahab, E. M. (2026).
The no-meaning falsity: The structural impossibility of the arbitrary sign in classical Arabic.
\textit{arXiv preprint} arXiv:2608.07737.
\url{https://arxiv.org/abs/2608.07737}

\bibitem{aho1986}
Aho, A. V., Sethi, R., \& Ullman, J. D. (1986).
\textit{Compilers: Principles, Techniques, and Tools}. Reading, MA: Addison-Wesley.

\bibitem{berggren1986}
Berggren, J. L. (1986).
\textit{Episodes in the Mathematics of Medieval Islam}. New York: Springer.

\bibitem{black2010}
Black, D. (2010).
\textit{Logic and Aristotle’s Rhetoric and Poetics in Medieval Arabic Philosophy}. Leiden: Brill.

\bibitem{bohas1984}
Bohas, G., \& Guillaume, J. P. (1984).
\textit{Studies in the Theory of the Arabic Lexicon}. Wiesbaden: Otto Harrassowitz.

\bibitem{ceri1990}
Ceri, S., Gottlob, G., \& Tanca, L. (1990).
\textit{Logic Programming and Databases}. Berlin: Springer.

\bibitem{habash2010}
Habash, N. (2010).
\textit{Introduction to Arabic Natural Language Processing}. San Rafael, CA: Morgan \& Claypool.

\bibitem{hallaq1997}
Hallaq, W. B. (1997).
\textit{A History of Islamic Legal Theories: An Introduction to Sunnī Uṣūl al-Fiqh}. Cambridge: Cambridge University Press.

\bibitem{hodges1993}
Hodges, W. (1993).
\textit{Model Theory}. Cambridge: Cambridge University Press.

\bibitem{kamali2003}
Kamali, M. H. (2003).
\textit{Principles of Islamic Jurisprudence}. Cambridge: Islamic Texts Society.

\bibitem{khwarizmi}
Al-Khwarizmi, M. ibn M. (c. 830).
\textit{Kitab Al-Jabr wa-l-Muqabala} (The Compendious Book on Calculation by Completion and Balancing).
(F. Rosen, Trans., 1831). Hildesheim: Georg Olms Verlag.

\bibitem{knuth1968}
Knuth, D. E. (1968).
Semantics of Context-Free Languages.
\textit{Mathematical Systems Theory}, 2(2), 127–145.

\bibitem{lloyd1987}
Lloyd, J. W. (1987).
\textit{Foundations of Logic Programming}. Berlin: Springer.

\bibitem{owens1988}
Owens, J. (1988).
\textit{The Foundations of Grammar: An Introduction to Medieval Arabic Grammatical Theory}. Amsterdam: John Benjamins.

\bibitem{rashed1994}
Rashed, R. (1994).
\textit{The Development of Arabic Mathematics: Between Arithmetic and Algebra}. Dordrecht: Springer.

\bibitem{ryding2005}
Ryding, K. C. (2005).
\textit{A Reference Grammar of Modern Standard Arabic}. Cambridge: Cambridge University Press.

\bibitem{shieber1986}
Shieber, S. M. (1986).
\textit{An Introduction to Unification-Based Approaches to Grammar}. Stanford, CA: Center for the Study of Language and Information (CSLI).

\bibitem{smrz2008}
Smrz, O. (2008).
Arabic Computational Morphology: A System for Processing Arabic Text. In A. Soudi, A. van den Bosch, \& G. Neumann (Eds.), \textit{Arabic Computational Linguistics} (pp. 21–46). Stanford, CA: CSLI.

\bibitem{street2004}
Street, T. (2004).
Arabic Logic. In D. M. Gabbay \& J. Woods (Eds.), \textit{Handbook of the History of Logic} (Vol. 1, pp. 523–585). Amsterdam: Elsevier.

\bibitem{turing1936}
Turing, A. M. (1936).
On Computable Numbers, with an Application to the Entscheidungsproblem.
\textit{Proceedings of the London Mathematical Society}, 2(1), 230–265.

\item[]\textbf{Special References}

\bibitem{axler1997linear}
Axler, S. (1997). \textit{Linear Algebra Done Right}. New York: Springer.

\bibitem{chang1990model}
Chang, C. C., \& Keisler, H. J. (1990). \textit{Model Theory} (3rd ed.). Amsterdam: Elsevier.

\bibitem{schrijver1998theory}
Schrijver, A. (1998). \textit{Theory of Linear and Integer Programming}. Chichester: John Wiley \& Sons.

\bibitem{wang1996}
Wang, H.
\textit{A Logical Journey: From G\"odel to Philosophy},
The MIT Press, Cambridge, Massachusetts, 1996.


\end{thebibliography}
\end{document}